\RequirePackage{fix-cm}
\documentclass[smallcondensed]{svjour3}     % onecolumn (ditto)
\smartqed  % flush right qed marks, e.g. at end of proof
\usepackage{graphicx}
\usepackage{epstopdf}
\usepackage{graphicx}
\usepackage{hyperref} % Automatically links the label references
\usepackage{algorithm,algorithmic}
\usepackage{colortbl}
\usepackage{color}
\usepackage[numbers]{natbib}
\usepackage{chngcntr}
\usepackage{longtable}
\usepackage{multirow}
\usepackage{rotating}
\usepackage{subfig}
\usepackage[title,toc,titletoc,page]{appendix}

\usepackage{amsmath,amsfonts,amssymb,amsthm}

\theoremstyle{plain}
\newtheorem{assumption1}{Assumption}
\newtheorem{theorem1}{Theorem}
\newtheorem{lemma1}{Lemma}

 \journalname{Applied Intelligence}
\begin{document}

\title{SAAGs: Biased Stochastic Variance Reduction Methods for Large-scale Learning\thanks{This is a post-peer-review, pre-copyedit version of an article published in Applied Intelligence. The final authenticated version is available online at: \url{https://doi.org/10.1007/s10489-019-01450-3}}
}


\author{Vinod~Kumar~Chauhan$\color{red}^*$   \thanks{$\color{red}^*$corresponding author}         \and
	Anuj~Sharma \and
	Kalpana~Dahiya
}


\institute{Vinod Kumar Chauhan \at
	Computer Science \& Applications\\
	Panjab University Chandigarh, INDIA\\
	%              Tel.: +123-45-678910\\
	%              Fax: +123-45-678910\\
	\email{jmdvinodjmd@gmail.com, vkumar@pu.ac.in}   \\
	Homepage: \url{https://sites.google.com/site/jmdvinodjmd}
	\and
	Anuj Sharma \at
	Computer Science \& Applications\\
	Panjab University Chandigarh, INDIA\\
	\email{anujs@pu.ac.in}\\
	Homepage: \url{https://sites.google.com/view/anujsharma}
	\and
	Kalpana Dahiya \at
	University Institute of Engineering and Technology\\
	Panjab University Chandigarh, INDIA\\
	\email{kalpanas@pu.ac.in}	
}

\date{Received: date / Accepted: date}

\maketitle

\begin{abstract}
Stochastic approximation is one of the effective approach to deal with the large-scale machine learning problems and the recent research has focused on reduction of variance, caused by the noisy approximations of the gradients. In this paper, we have proposed novel variants of SAAG-I and II (Stochastic Average Adjusted Gradient) Chauhan et el. (2017) \cite{Chauhan2017Saag}, called SAAG-III and IV, respectively. Unlike SAAG-I, starting point is set to average of previous epoch in SAAG-III, and unlike SAAG-II, the snap point and starting point are set to average and last iterate of previous epoch in SAAG-IV, respectively. To determine the step size, we have used Stochastic Backtracking-Armijo line Search (SBAS) which performs line search only on selected mini-batch of data points. Since backtracking line search is not suitable for large-scale problems and the constants used to find the step size, like Lipschitz constant, are not always available so SBAS could be very effective in such cases. We have extended SAAGs (I, II, III and IV) to solve non-smooth problems and designed two update rules for smooth and non-smooth problems. Moreover, our theoretical results have proved linear convergence of SAAG-IV for all the four combinations of smoothness and strong-convexity, in expectation. Finally, our experimental studies have proved the efficacy of proposed methods against the state-of-art techniques.
\end{abstract}

\section{Introduction}
\label{sec_intro}
Large-scale machine learning problems have large number of data points or large number of features in each data point, or both are large. This leads to high per-iteration complexity of the iterative learning algorithms, which results in slow training of models. Thus, large-scale learning or learning on the big data is one of the major challenge today in machine learning \cite{Chauhan2017Saag,Zhou2017}. To tackle this large-scale learning challenge, recently research has focused on stochastic optimization approach \cite{Chauhan2018STRON}, coordinate descent approach \cite{Wright2015}, proximal algorithms \cite{Parikh2014}, parallel and distributed algorithms \cite{Yang2016} (as discussed in \cite{Chauhan2018Review}), and momentum acceleration algorithms \cite{Allen2017}. Stochastic approximation leads to variance in the values of deterministic gradient and noisy gradients which are calculated using stochastic approximation, and affects the convergence of learning algorithm. There are several approaches to deal with stochastic noise but most important of these (as discussed in \cite{Csiba2016}) are: (a) using mini-batching \cite{Chauhan2018SS_AI}, (b) decreasing learning rates \cite{Shalev-Shwartz2007}, (c) variance reduction \cite{Roux2012}, and (d) importance sampling \cite{Csiba2016}. To deal with the large-scale learning problems, we use mini-batching and variance reduction in this paper.

\subsection{Optimization Problem}
\label{subsec_op}
In this paper, we consider composite convex optimization problem, as given below:
\begin{equation}
\label{eq_op}
\underset{w\in \mathbb{R}^d}{\min} \;\; \left\lbrace F(w) = f(w) + g(w) = \dfrac{1}{n} \sum_{i=1}^{n} f_i (w) + g(w),\right\rbrace
\end{equation}
where $f(w)$ is a finite average of component functions $ f_i(w): \mathbb{R}^d \rightarrow \mathbb{R}, \; i=1,2,...,n$, are convex and smooth, $g(w): \mathbb{R}^d \rightarrow \mathbb{R}$ is a relatively simple convex but possibly non-differentiable function (also referred to as regularizer and sometimes as proximal function). 
%In this paper, we mainly focus on strongly convex problem with smooth and non-smooth regularizer. 
This kind of optimization problems can be found in operation research, data science, signal processing, statistics and machine learning etc. For example,  regularized Empirical Risk Minimization (ERM) problem is a common problem in machine learning, which is average of losses on the training dataset. In ERM, component function $f_i(w)$ denotes value of loss function at one data point, e.g., in binary classification, it can be logistic loss, i.e., $f_i(w) = \log \left( 1 + \exp \left( -y_i w^T x_i \right) \right)$, where $\{(x_1, y_1), (x_2, y_2),...,(x_n, y_n)\}$ is collection of training data points, and hinge-loss, i.e., $f_i(w) = \max \left( 0, 1-y_i w^T x_i \right)$; for regression problem, it can be least squares, i.e., $f_i(w) = 1/2(w^Tx_i - y_i)^2$. The regularizer can be $\lambda_1 \|w\|_1$ ($l_1$-regularizer), $\lambda_2/2 \|w\|^2$ ($l_2$-regularizer) and $\lambda_1 \|w\|_1 + \lambda_2/2 \|w\|^2$ (elastic net regularizer), where $\lambda_1$ and $\lambda_2$ are regularization coefficients. Thus, problems like logistic regression, SVM, ridge regression and lasso etc. fall under ERM.

\subsection{Solution Techniques for Optimization Problem}
\label{subsec_solve_op}
The simple first order method to solve problem (\ref{eq_op}) is given by Cauchy in his seminal work in 1847, known as Gradient Descent (GD) method \cite{Cauchy1847}, is given below for $(k+1)^{th}$ iteration as:
\begin{equation}
\label{eq_gd}
w_{k+1} = w_{k} - \eta_k \left[ \dfrac{1}{n}\sum_{i=1}^{n} \nabla f_i(w_k) + \nabla g(w_k) \right],
\end{equation}
where $\eta_k$ is the learning rate (also known as step size in optimization). For non-smooth regularizer, i.e., when $g(w)$ is non-smooth then typically, proximal step is calculated after the gradient step, and method is called Proximal Gradient Descent (PGD), as given below:
\begin{equation}
\label{eq_pgd}
w_{k+1} = Prox_{\eta_k}^{g} \left( z_k \right) = \underset{w\in \mathbb{R}^d}{\arg\min} \left\lbrace \dfrac{1}{2\eta_k} \| w-z_k \|^2 + g(w) \right\rbrace,
\end{equation}
where $z_k = w_{k} -   \eta_k/n\sum_{i=1}^{n} \nabla f_i(w_k)$. GD converges linearly for strongly-convex and smooth problem, and (\ref{eq_pgd}) converges at a rate of $O(1/k)$ for non-strongly convex differentiable problems, where $k$ is the number of iterations. The per-iteration complexity of GD and PGD methods is $O(nd)$. Since for large-scale learning problems, the values of $n$ (number of data points) and/or $d$ (number of features in each data point) is very large, so the per-iteration complexity of these methods is very large. Each iteration becomes computationally expensive and might even be infeasible to process for a limited capacity machine, which leads to slow training of models in machine learning. Thus, the challenge is to develop efficient algorithms to deal with the large-scale learning problems \cite{Chauhan2017Saag,Zhou2017}.\\
To tackle this challenge, stochastic approximation is one of the popular approach, first introduced by Robbins and Monro, in their seminal work back in 1951, which makes each iteration independent of number of data points \cite{Kiefer1952,Robbins1951}. Based on this approach, we have Stochastic Gradient Descent (SGD) method \cite{Bottou2010}, as given below, to solve problem (\ref{eq_op}) for the smooth case:
\begin{equation}
\label{eq_sgd}
w_{k+1} = w_{k} - \eta_k \left[ \nabla f_{i_k}(w_k) + \nabla g(w_k) \right],
\end{equation}
where $i_k$ is selected uniformly randomly from \{1,2,...,n\} and $\eta_k \propto 1/\sqrt{k}$. The per-iteration complexity of SGD is $O(d)$ and it is very effective to deal with problems with large number of data points. Since $ E\left[ \nabla f_{i_k} (w_k) \right] = \nabla f(w_k)$, $\nabla f_{i_k} (w_k)$ is an unbiased estimator of $\nabla f(w_k)$, but the variance in these two values need decreasing learning rates. This leads to slow convergence of learning algorithms. Recently, a lot of research is going on to reduce the variance between $\nabla f_{i_k} (w_k)$ and $\nabla f(w_k)$. First variance reduction method introduced by \cite{Roux2012}, called SAG (Stochastic Average Gradient), some other common and latest methods are SVRG (Stochastic Variance Reduced Gradient) \cite{Johnson2013}, Prox-SVRG \cite{Xiao2014}, S2GD (Semi-Stochastic Gradient Descent) \cite{Konecny2013,Yang2018}, SAGA \cite{Defazio2014}, Katyusha \cite{Allen2017}, VR-SGD (Variance Reduced Gradient Descent) \cite{Shang2018}, SAAG-I and II \cite{Chauhan2017Saag} etc. Like, GD, these methods utilize full gradient and like, SGD, these methods calculate gradient for one or few data points during each iteration. Thus, just like GD, these methods converge linearly for strongly convex and smooth problems and like, SGD, they calculate gradient using one or few data points and have low per-iteration complexity. Thus, the variance reduction methods enjoy best of GD and SGD. Please refer to, \cite{Bottou2016} for a review on optimization methods for solving large-scale machine learning problems. In this paper, we have proposed new variants of SAAG-I and II, called SAAG-III and IV, as variance reduction techniques.

\subsection{Research Contributions}
\label{subsec_contri}
The research contributions of this paper are summarized below:
\begin{enumerate}
	\item[(a)] Novel variants of SAAG-I and II are proposed, called SAAG-III and SAAG-IV, respectively. Unlike SAAG-I, for SAAG-III the starting point is set to the average of iterates of previous epoch except for the first one, $ w^{s+1}_0 = 1/m \sum_{i=1}^{m} w_i^{s}$, where $m$ is number of inner iterations. Unlike SAAG-II, for SAAG-IV, the starting point and snap point are set to the last iterate and average of previous epoch except for the first one, $w^{s+1}_0 = w^{s}_m$ and $ \tilde{w}^{s}_0 = 1/m \sum_{i=1}^{m} w_i^{s}$.
	
	\item[(b)] SAAG-I and II, including SAAG-III and IV, are extended to solve problems with non-smooth regularizer by introducing two different update rules for smooth and non-smooth cases (see Section \ref{sec_saagi_ii} and \ref{sec_saagiii_iv}, for details).
	
	\item[(c)] Theoretical results prove linear convergence of SAAG-IV for all the four combinations of smoothness and strong-convexity in expectation.
	
	\item[(d)] Finally, empirical results prove the efficacy of proposed methods against state-of-art methods in terms of convergence and accuracy against training time, number of epochs and number of gradient evaluations.
\end{enumerate}

\section{Notations and Related Work}
\label{sec_related_work}
This section discusses notations used in the paper and related work.

\subsection{Notations}
\label{subsec_notations}
The training dataset is represented as $\left\lbrace (x_1,y_1), (x_2,y_2),...,(x_n,y_n) \right\rbrace$, where $n$ is the number of data points and $x_i \in \mathbb{R}^d, \; d$ is the number of features. $w$ denotes the parameter vector and $\lambda_i, \; i=1,2$ denotes the regularization parameter. $\|.\|$ denotes Euclidean norm, also called $l_2$-norm, and $\|.\|_1$ denotes $l_1$-norm. $L$ and $\mu$ are used to denote $L$-smoothness and $\mu$-strong convexity of problem, respectively. $\eta_k^s$ denotes the learning rate, $s\in  \left\lbrace 1,2,...,S \right\rbrace$ denotes epoch number and $S$ is the total number of epochs. $b$ denotes the mini-batch size, $k\in  \left\lbrace 1,2,...,m \right\rbrace$ denotes the inner iterations, s.t., $n = mb$. The value of loss function at $(x_i,y_i)$ is denoted by component function $f_i$. $w^{*} = \arg\min_{w} F(w)$ and $F(w^{*})$ is the optimal objective function value, sometimes denoted as $F^{*}$.

\subsection{Related Work}
\label{subsec_literature}
The emerging technologies and the availability of different data sources have lead to rapid expansion of data in all science and engineering domains. On one side, this massive data has potential to uncover more fine-grained patterns, take timely and accurate decisions, and on other side it creates a lot of challenges to make sense of it, like, slow training and scalability of models, because of inability of traditional technologies to process this huge data. The term ``Big Data" was coined to highlight the data explosion and need of new technologies to process this massive data. Big data is a vast subject in itself. Big data can be characterized, mainly using three Vs: Volume, Velocity and Variety but recently a lot of other Vs have been used. When one deals with `volume' aspect of big data in machine learning, it is called the large-scale machine learning problem or big data problem. The large-scale learning problems have large number of data points or large number of features in each data point, or both, which lead to large per-iteration complexity of learning algorithms, and ultimately lead to slow training of machine learning models. Thus, one of the major challenge before machine learning is to develop efficient and scalable learning algorithms \cite{Chauhan2017Saag,Chauhan2018Review,Zhou2017}.\\
To solve problem (\ref{eq_op}) for smooth regularizer, a simple method is GD, given by \cite{Cauchy1847}, and it converges linearly for strongly-convex and smooth problems. For non-smooth regularizer, typically, proximal step is applied to GD step, called PGD method which converges at a rate of $O(1/k^2)$ for non-strongly convex problems. The per-iteration complexity of GD and PGD is $O(nd)$ which is very large for large-scale learning problems and results in slow training of models. Stochastic approximation is one of the approach to tackle this challenge. It was first introduced by Robbins and Monro \cite{Robbins1951} and is very effective to deal with problems with large number of data points because each iteration uses one (or few) data points, like in SGD \cite{Bottou2010,Zhang2004}. In SGD, each iteration is $n$ times faster than GD, as their per-iteration complexities are $O(d)$ and $O(nd)$, respectively. SGD need decreasing learning rates, i.e., $\eta_k \propto 1/\sqrt{k}$ for $k^{th}$ iteration, because of variance in gradients, so it converges slower than GD, with sub-linear convergence rate $O(1/T)$ even for strongly convex problem \cite{Rakhlin2012}. There are several approaches to deal with stochastic noise but most important of these (as discussed in \cite{Csiba2016}) are: (a) using mini-batching \cite{Yang2018}, (b) decreasing learning rates\cite{Shalev-Shwartz2007}, (c) variance reduction \cite{Roux2012}, and (d) importance sampling \cite{Csiba2016}.\\
Variance reduction techniques, first introduced by \cite{Roux2012}, called SAG, converges linearly, like, GD for strongly convex and smooth problems, and uses one randomly selected data point, like SGD during each iteration. SAG enjoys benefits of both GD and SGD, as it converges linearly for strongly convex and smooth case like, GD but it has per-iteration complexity of SGD. Later, a lot of variance reduction methods were proposed, like, SVRG \cite{Johnson2013}, SAGA \cite{Defazio2014}, S2GD \cite{Konecny2013}, SDCA \cite{Shalev2013a}, SPDC \cite{Zhang2015}, Katyusha \cite{Allen2017}, Catalyst \cite{Lin2015}, SAAG-I, II \cite{Chauhan2017Saag} and VR-SGD \cite{Shang2018} etc. These variance reduction methods can use constant learning rate and can be divided into three categories (as discussed in \cite{Shang2018}): (a) primal methods which can be applied to primal optimization problem, like, SAG, SAGA, SVRG etc., (b) dual methods which can be applied to dual problems, like, SDCA, and (c) primal-dual methods which involve primal and dual variable both, like, SPDC etc.\\
In this paper, we have proposed novel variants of SAAG-I and II, named as SAAG-III and SAAG-IV, respectively. Unlike SAAG-I, for SAAG-III the starting point is set to the average of iterates of previous epoch except for the first one, $ w^{s+1}_0 = 1/m \sum_{i=1}^{m} w_i^{s}$, where $m$ is number of inner iterations. Unlike SAAG-II, for SAAG-IV, the starting point and snap point are set to the last iterate and average of previous epoch except for the first one, $w^{s+1}_0 = w^{s}_m$ and $ \tilde{w}^{s}_0 = 1/m \sum_{i=1}^{m} w_i^{s}$. \cite{Chauhan2017Saag} proposed Batch Block Optimization Framework (BBOF) to tackle the big data (large-scale learning) challenge in machine learning, along with two variance reduction methods SAAG-I and II. BBOF is based on best of stochastic approximation (SA) and best of coordinate descent (CD) (another approach which is very effective to deal with large-scale learning problems especially problems with high dimensions). Techniques based on best features of SA and CD approaches are also used  in \cite{Wang2014,Xu2015,ZebangShen2017,Zhao2014a}, and \cite{ZebangShen2017} calls it doubly stochastic since both data points and coordinate are sampled during each iteration. It is observed that for ERM, it is difficult to get the advantage of BBOF in practice because results with CD or SGD are faster than BBOF setting as BBOF needs extra computations while sampling and updating the block of coordinates. When one block of coordinates is updated and as we move to another block, the partial gradients need dot product of parameter vector ($w$) and data points ( like, in logistic regression). Since each update changes $w$ so for each block update needs to calculate the dot product. On other hand, if all coordinates are updated at a time, like in SGD, that would need to calculate dot product only once. Although, Gauss-Seidel update of parameters helps in faster convergence but the overall gain is less because of extra computational load. Moreover, SAAG-I and II have been proposed to work in BBOF (mini-batch and block-coordinate) setting as well as mini-batch (and considering all coordinates). Since BBOF is not very helpful for ERM so the SAAG-III and IV are proposed for mini-batch setting only. SAAGs (I, II, III and IV) can be extended to stochastic setting (consider one data point during each iteration) but SAAG-I and II are unstable for stochastic setting, and SAAG-III and IV, could not beat the existing methods in stochastic setting. SAAGs has been extended to deal with smooth and non-smooth regularizers, as we have used two different update rules, like \cite{Shang2018} (see Section \ref{sec_saagi_ii} for details).

\section{SAAG-I, II and Proximal Extensions}
\label{sec_saagi_ii}
Originally, \cite{Chauhan2017Saag} proposed SAAG-I and II for smooth problems, we have extended SAAG-I and II to non-smooth problems. Unlike proximal methods which use single update rule for both smooth and non-smooth problems, we have used two different update rules and introduced proximal step for non-smooth problem. For mini-batch $B_k$ of size $b$, epoch $s$ and inner iteration $k$, SAAG-I and II are given below:\\

SAAG-I:
\begin{equation}
\label{eq_saag_1}
\begin{array}{ll}
\text{smooth} & w_{k+1}^s = w_{k}^s - \eta_k^s \left[ \sum_{i=1}^{n} \tau^{s,k}_i + \nabla g(w^s_k) \right],\\
\text{non-smooth } & w_{k+1}^s = Prox_{\eta_k^s}^g \left( w_{k}^s - \eta_k^s \sum_{i=1}^{n} \tau^{s,k}_i   \right),	
\end{array}
\end{equation}
where $\tau^{s,k}_i =
\begin{cases}
\dfrac{1}{b} \nabla f_i (w^s_k) \quad \text{if}\; i \in B_k\\
\dfrac{1}{l} \tilde{\tau}^{s,k}_i \quad \text{otherwise}
\end{cases}$ and
$\tilde{\tau}^{s,k}_i =
\begin{cases}
\nabla f_i (w^s_k) \quad \text{if}\; i \in B_k\\
\tilde{\tau}^{s-1,k}_i  \quad \text{otherwise}.
\end{cases}$\\

SAAG-II:
\begin{equation}
\label{eq_saag_2}
\begin{array}{ll}
\text{smooth} & w_{k+1}^s = w_{k}^s - \eta_k^s \left[ \tilde{\nabla} f_{B_k}(w^s_k) + \nabla g(w^s_k) \right],\\
\text{non-smooth } & w_{k+1}^s = Prox_{\eta_k^s}^g \left( w_{k}^s - \eta_k^s \tilde{\nabla} f_{B_k}(w^s_k)   \right),
\end{array}
\end{equation}
where $\tilde{\nabla} f_{B_k}(w^s_k) = 1/b \sum_{i\in B_k} \nabla f_i (w^s_k) - 1/n\sum_{i\in B_k} \nabla f_i (\tilde{w}^s) + \tilde{\mu}^s$ and $\tilde{\mu}^s = 1/n\sum_{i=1}^{n} \nabla f_i (\tilde{w}^s)$. Unlike SVRG and VRSGD, and like SAG, SAAGs are biased gradient estimators because the expectation of gradient estimator is not equal to full gradient, i.e., $\mathbb{E}[\tilde{\nabla} f_{B_k}(w^s_k)  ] = \nabla f(w^s_k) + (m-1)/m \nabla f(\tilde{w}^{s-1})$, as detailed in Lemma~\ref{lemma_variance_bound_smooth}.\\
%SAAG-III extends SAAG-I and use average of previous iterates as the starting point for each epoch, except the first one. SAAG-IV extends SAAG-II and use previous iterate and average of previous iterates as the starting point and snap point, respectively, in all epochs, except first one. SVRG, which is the most popular reduction technique, use iterate average as the starting point and snap-point. Recently, VR-SGD extended SVRG and used previous iterate and iterate average as the starting point and snap-point, respectively.\\
SAAG-I algorithm, represented by Algorithm \ref{algo_saagi}, divides the dataset into $m$ mini-batches of equal size (say) $b$ and takes input $S$, number of epochs. During each inner iteration, it randomly selects one mini-batch of data points from $[n]$, calculates gradient over mini-batch, updates the total gradient value and performs stochastic backtracking-Armijo line search (SBAS) over $B_k$. Then parameters are updated using Option I for smooth regularizer and using Option II for non-smooth regularizer. Inner iterations are run $m$ times where $ n = mb$ and then last iterate is used as the starting point for next epoch.\\
\begin{algorithm}[htb]
	\caption{SAAG-I}
	\label{algo_saagi}
	\textbf{Inputs:} $m=\#$mini-batches and $S=$max $\# epochs$.\\
	\textbf{Initialize:} $w^1_0$
	\begin{algorithmic}[1]
		\FOR{$s=1,2,...,S$}
		\FOR{$k=0,1,...,(m-1)$}
		\STATE Randomly select one mini-batch $B_k$ from [n].
		\STATE Update the gradient values, $\tau^{s,k}_i$ and $\tilde{\tau}^{s,k}_i$.
		\STATE Calculate $\eta_k^s$ using stochastic backtracking line search on $B_k$.
		\STATE Option I (smooth): $w_{k+1}^s = w_{k}^s - \eta_k^s \left[ \sum_{i=1}^{n} \tau^{s,k}_i + \nabla g(w^s_k) \right],$
		\STATE Option II (non-smooth): $w_{k+1}^s = Prox_{\eta_k^s}^g \left( w_{k}^s - \eta_k^s \sum_{i=1}^{n} \tau^{s,k}_i   \right),	$\\
		where $\tau^{s,k}_i =
		\begin{cases}
		\dfrac{1}{b} \nabla f_i (w^s_k) \quad \text{if}\; i \in B_k\\
		\dfrac{1}{l} \tilde{\tau}^{s,k}_i \quad \text{otherwise}
		\end{cases}$ and
		$\tilde{\tau}^{s,k}_i =
		\begin{cases}
		\nabla f_i (w^s_k) \quad \text{if}\; i \in B_k\\
		\tilde{\tau}^{s-1,k}_i  \quad \text{otherwise}.
		\end{cases}$
		\ENDFOR
		\ENDFOR
		\STATE \textbf{Output:} $w^S_m$
	\end{algorithmic}
\end{algorithm}
\begin{algorithm}[htb]
	\caption{SAAG-II}
	\label{algo_saagii}
	\textbf{Inputs:} $m=\#$mini-batches and $S=$max $\# epochs$.\\
	\textbf{Initialize:} $w^1_0$
	\begin{algorithmic}[1]
		\FOR{$s=1,2,...,S$}
		\STATE $\tilde{w}^s = w^s_0$
		\STATE $\tilde{\mu}^s = \dfrac{1}{n} \sum_{i=1}^{n} \nabla f_i (\tilde{w}^{s})$ \hspace*{\fill} // calculate full gradient
		\FOR{$k=0,1,...,(m-1)$}
		\STATE Randomly select one mini-batch $B_k$ from [n].
		\STATE Calculate $\tilde{\nabla} f_{B_k} (w^s_k)= \dfrac{1}{b} \sum_{i \in B_k} \nabla f_i (w^s_k) - \dfrac{1}{n}\sum_{i \in B_k} \nabla f_i (\tilde{w}^{s}) + \tilde{\mu}^s$.
		\STATE Calculate $\eta_k^s$ using stochastic backtracking-Armijo line search on $B_k$.		
		\STATE Option I (smooth): $ w^s_{k+1} = w^s_k - \eta_k^s \left[ \tilde{\nabla} f_{B_k} (w^s_k) + \nabla g(w^s_k) \right]$
		\STATE Option II (non-smooth ): $w^s_{k+1} = Prox^g_{\eta_k^s} \left( w^s_k - \eta_k^s  \tilde{\nabla} f_{B_k} (w^s_k) \right)$
		\ENDFOR
		\STATE $w^{s+1}_0 = w^s_m$
		\ENDFOR
		\STATE \textbf{Output:} $w^S_m$
	\end{algorithmic}
\end{algorithm}
SAAG-II algorithm, represented by Algorithm \ref{algo_saagii}, takes input as number of epochs ($S$) and number of mini-batches ($m$) of equal size (say) $b$. It initializes $w^1_0 = \tilde{w}^0$. During each inner iteration, it randomly selects one mini-batch $B_k$, calculates two gradients over $B_k$ at last iterate and snap-point, updates $\tilde{\nabla} f_{B_k} (w^s_k)$ and performs stochastic backtracking-Armijo line search (SBAS) over $B_k$. Then parameters are updated using Option I for smooth regularizer and using Option II for non-smooth regularizer. After inner iterations, it uses the last iterate to set the snap point and the starting point for the next epoch.

%\section{SAAG-III and IV}
%\label{sec_saagiii_iv}
%In this section, SAAG-III and IV are discussed and compared with SAAG-I and II.

\section{SAAG-III and IV Algorithms}
\label{sec_saagiii_iv}
\begin{figure}[htb]
	\subfloat{\includegraphics[width=.332\linewidth]{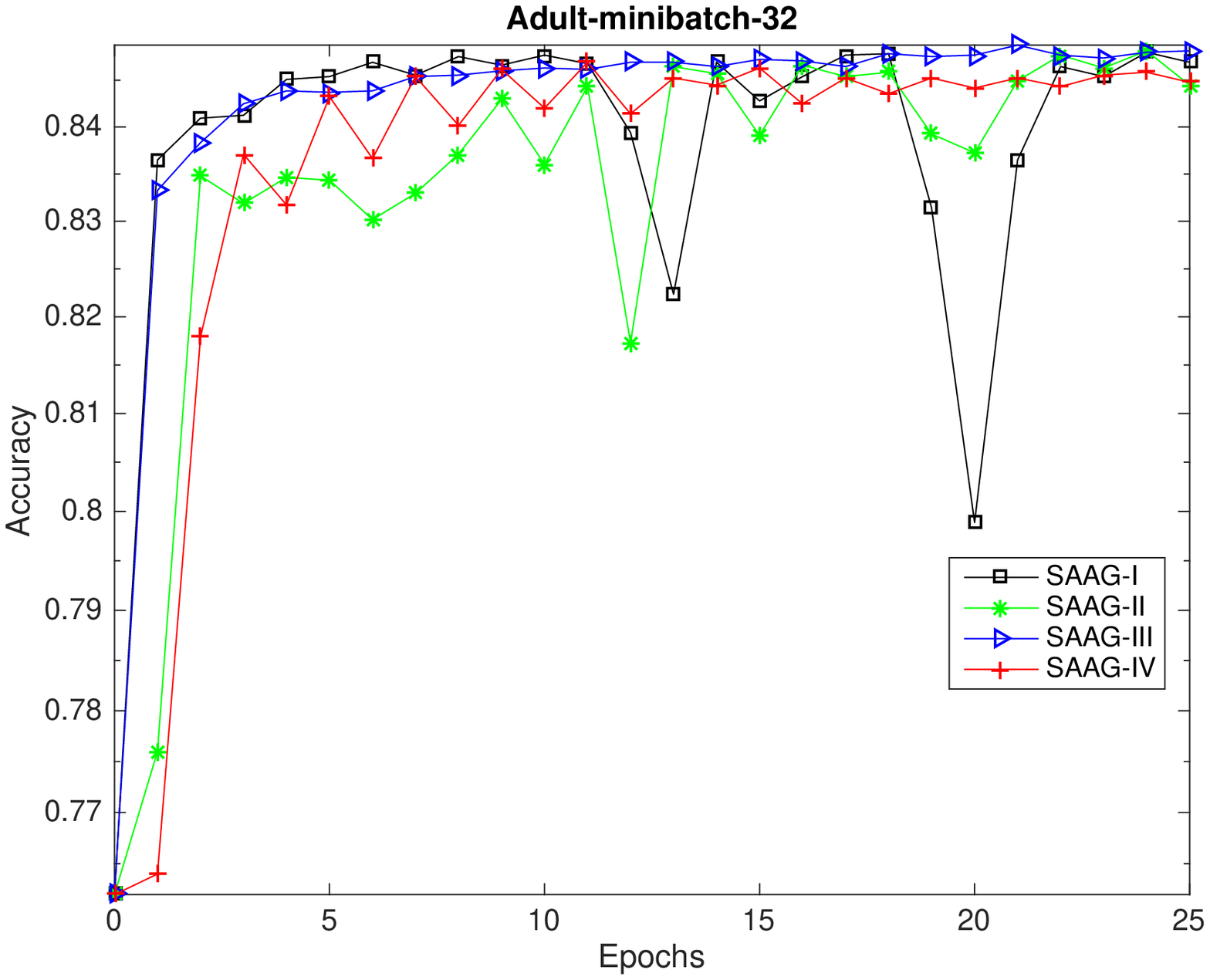}}
	\subfloat{\includegraphics[width=.332\linewidth]{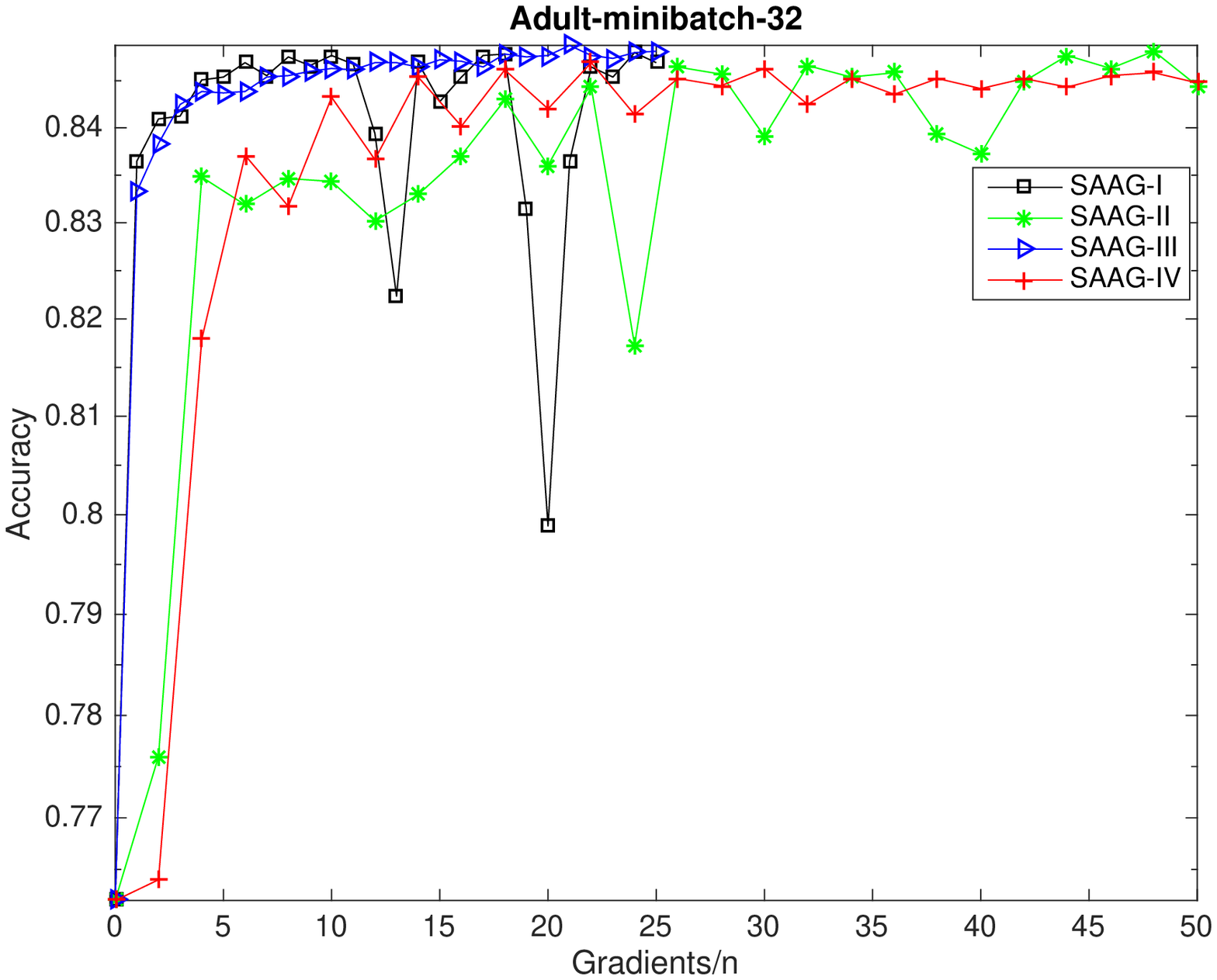}}
	\subfloat{\includegraphics[width=.332\linewidth]{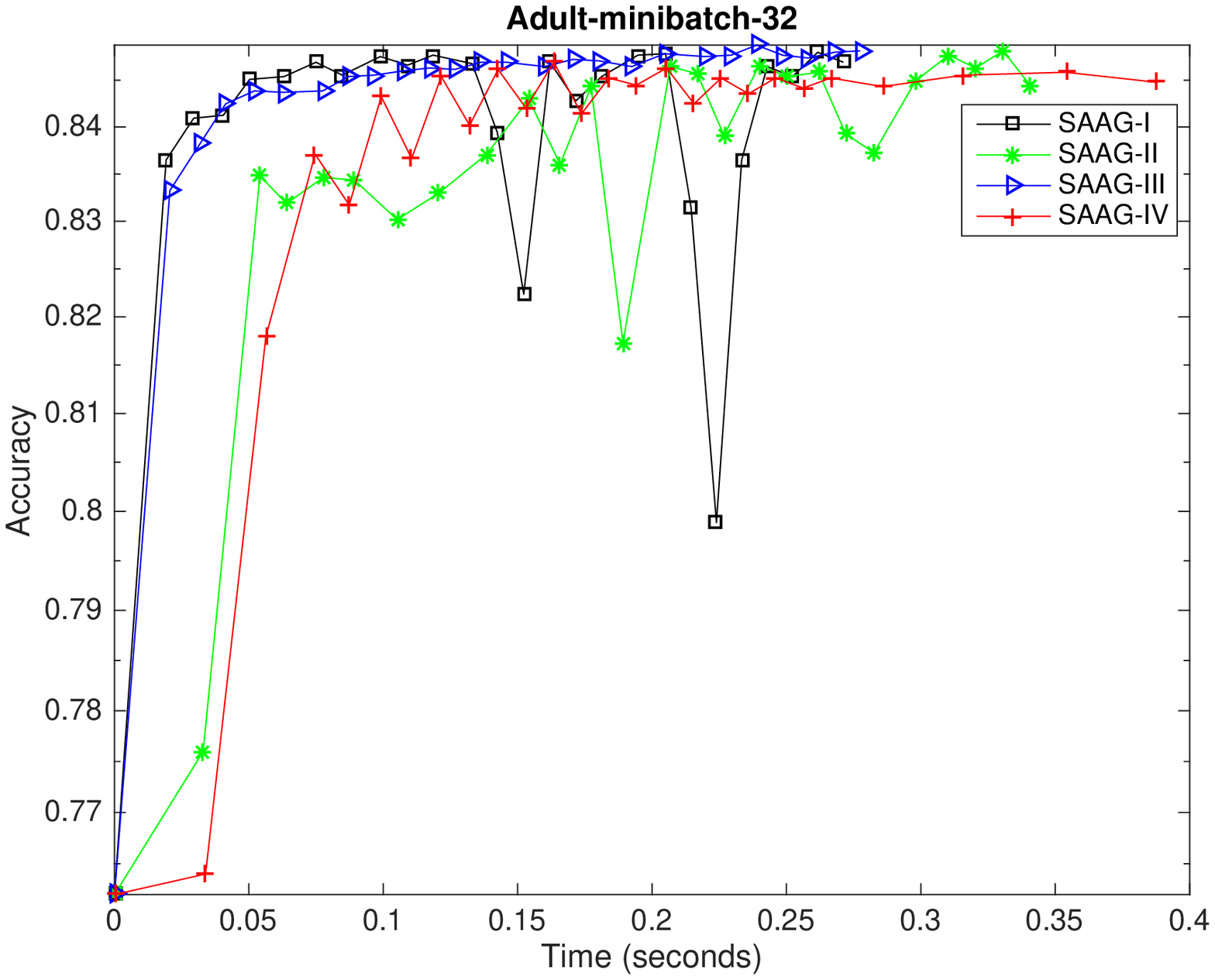}}
	
	\subfloat{\includegraphics[width=.332\linewidth]{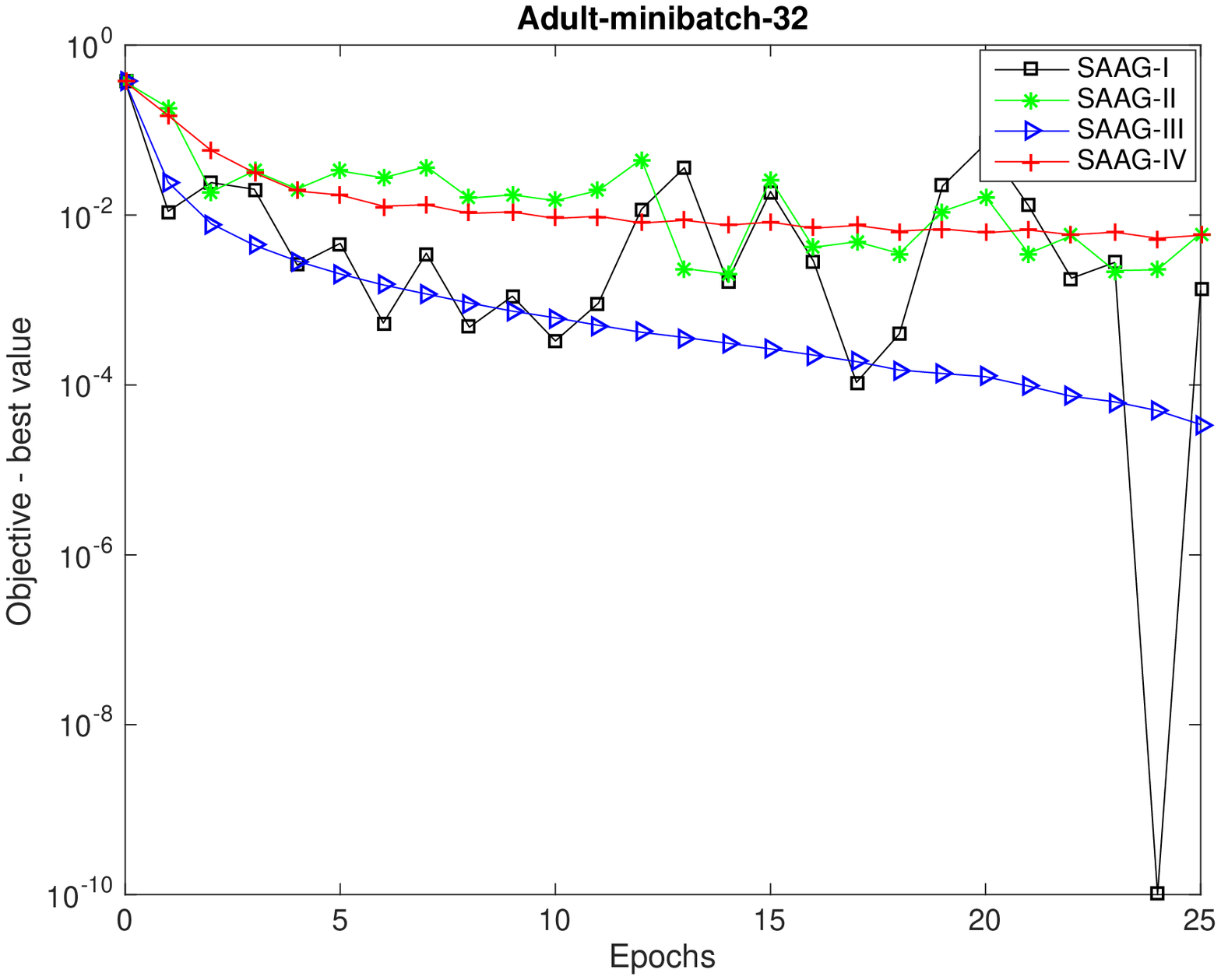}}
	\subfloat{\includegraphics[width=.332\linewidth]{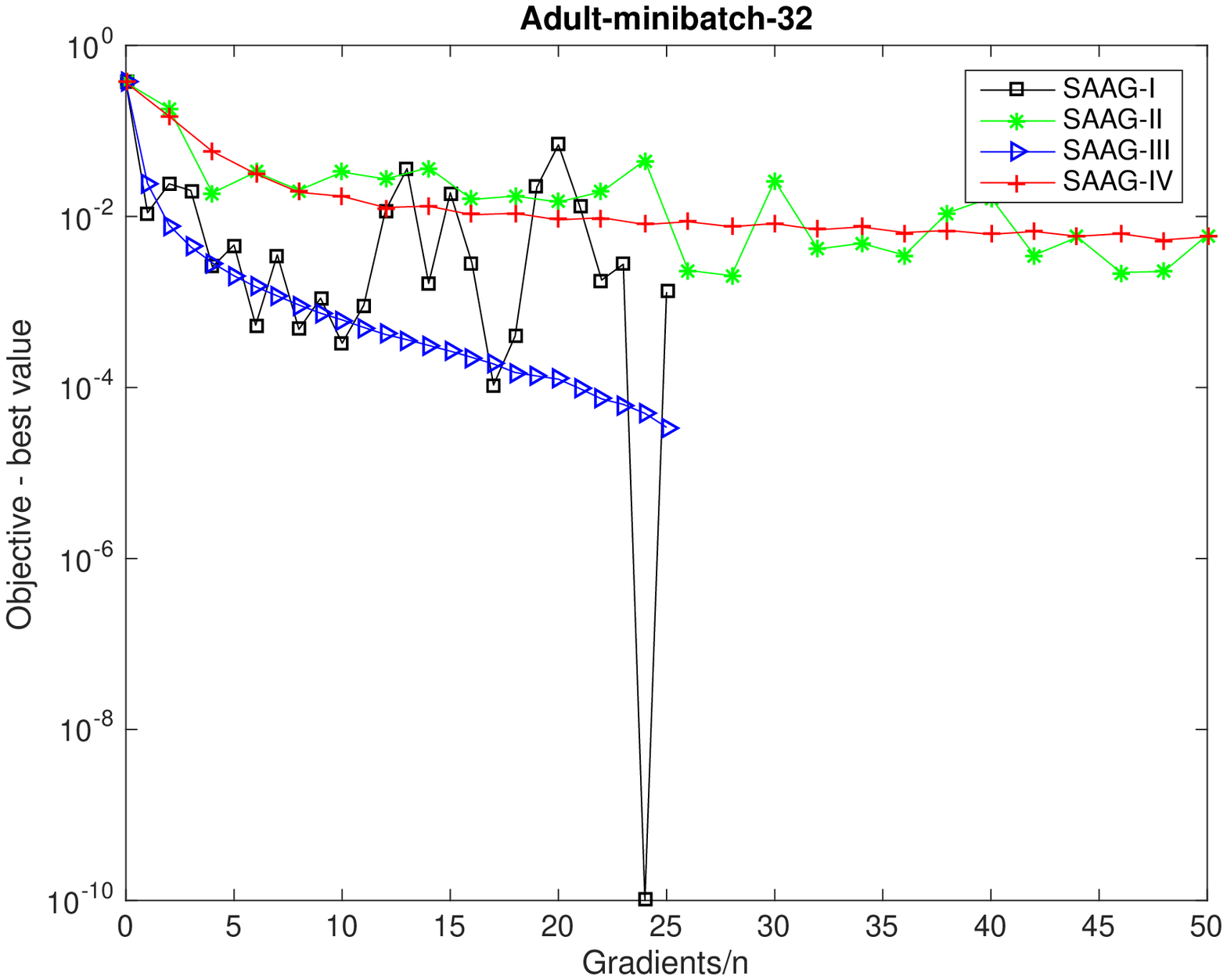}}
	\subfloat{\includegraphics[width=.332\linewidth]{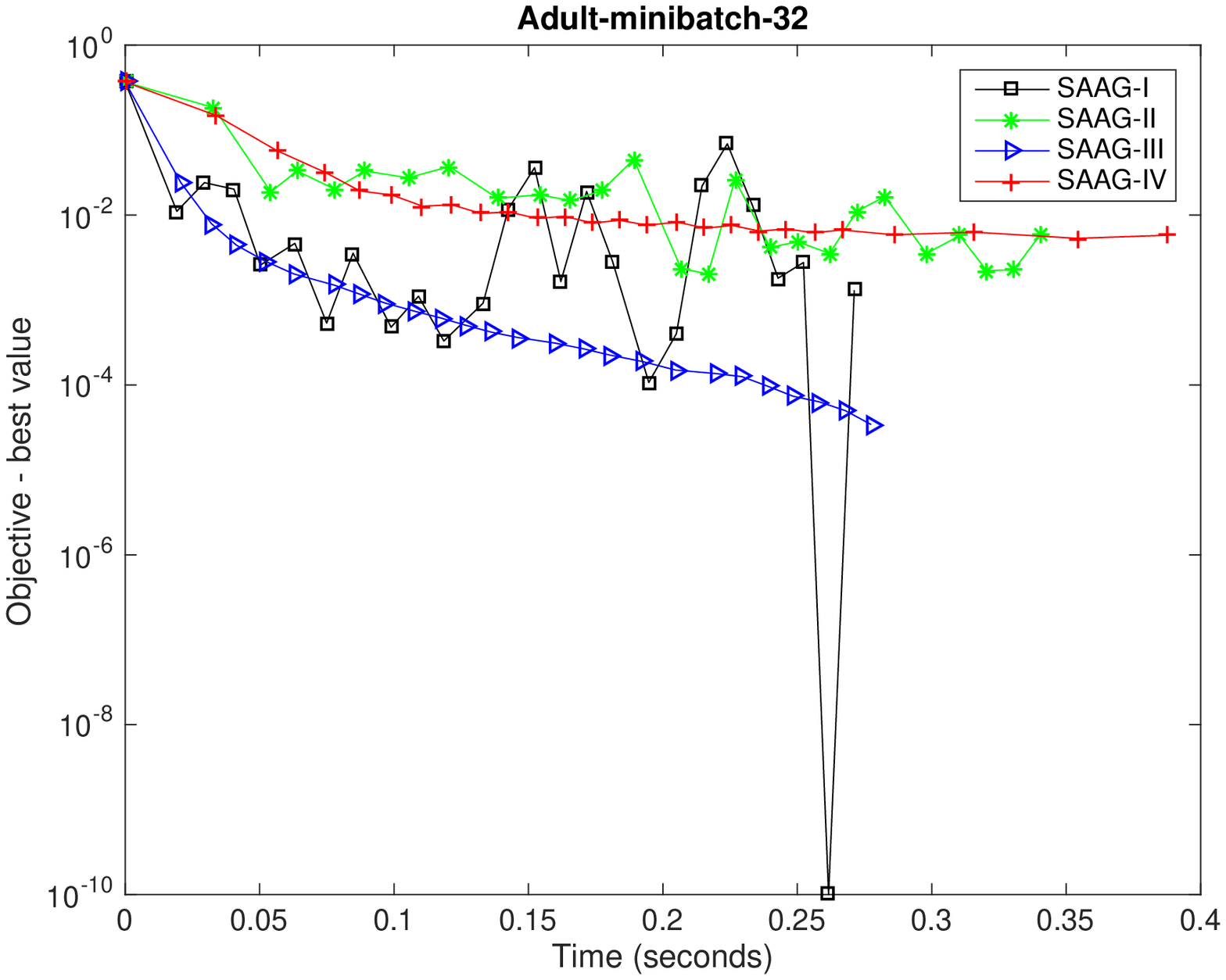}}
	
	\caption{Comparison of SAAG-I, II, III and IV on smooth problem using Adult dataset with mini-batch of 32 data points. First row compares accuracy against epochs, gradients/n and time, and second row compares suboptimality against epochs, gradients/n and time.}
	\label{fig_saags_smooth}
\end{figure}

%\label{subsec_saagiii_iv_algos}
SAAG-III algorithm, represented by Algorithm \ref{algo_saagiii}, divides the dataset into $m$ mini-batches of equal size (say) $b$ and takes input $S$, number of epochs. During each inner iteration, it randomly selects one mini-batch of data points $B_k$ from $[n]$, calculates gradient over mini-batch, updates the total gradient value and performs stochastic backtracking-Armijo line search (SBAS) over $B_k$. Then parameters are updated using Option I for smooth regularizer and using Option II for non-smooth regularizer. Inner iterations are run $m$ times where $ n = mb$ and then iterate average is calculated  and used as the starting point for next epoch, $w^{s+1}_0 = 1/m \sum_{i=1}^{m} w_i^{s}$.\\
\begin{algorithm}[htb]
	\caption{SAAG-III}
	\label{algo_saagiii}
	\textbf{Inputs:} $m=\#$mini-batches and $S=$max $\# epochs$.\\
	\textbf{Initialize:} $w^1_0$
	\begin{algorithmic}[1]
		\FOR{$s=1,2,...,S$}
		\FOR{$k=0,1,...,(m-1)$}
		\STATE Randomly select one mini-batch $B_k$ from [n].
		\STATE Update the gradient values, $\tau^{s,k}_i$ and $\tilde{\tau}^{s,k}_i$.
		\STATE Calculate $\eta_k^s$ using stochastic backtracking line search on $B_k$.
		\STATE Option I (smooth): $w_{k+1}^s = w_{k}^s - \eta_k^s \left[ \sum_{i=1}^{n} \tau^{s,k}_i + \nabla g(w^s_k) \right],$
		\STATE Option II (non-smooth): $w_{k+1}^s = Prox_{\eta_k^s}^g \left( w_{k}^s - \eta_k^s \sum_{i=1}^{n} \tau^{s,k}_i   \right),	$\\
		where $\tau^{s,k}_i =
		\begin{cases}
		\dfrac{1}{b} \nabla f_i (w^s_k) \quad \text{if}\; i \in B_k\\
		\dfrac{1}{l} \tilde{\tau}^{s,k}_i \quad \text{otherwise}
		\end{cases}$ and
		$\tilde{\tau}^{s,k}_i =
		\begin{cases}
		\nabla f_i (w^s_k) \quad \text{if}\; i \in B_k\\
		\tilde{\tau}^{s-1,k}_i  \quad \text{otherwise}.
		\end{cases}$
		\ENDFOR
		\STATE $w^{s+1}_0 = 1/m \sum_{i=1}^{m} w_i^{s}$ \hspace*{\fill} // iterate averaging
		\ENDFOR
		\STATE \textbf{Output:} $w^S_m$
	\end{algorithmic}
\end{algorithm}
\begin{algorithm}[htb]
	\caption{SAAG-IV}
	\label{algo_saagiv}
	\textbf{Inputs:} $m=\#$mini-batches and $S=$max $\# epochs$.\\
	\textbf{Initialize:} $w^1_0 = \tilde{w}^0$
	\begin{algorithmic}[1]
		\FOR{$s=1,2,...,S$}
		\STATE $\tilde{\mu}^s = \dfrac{1}{n} \sum_{i=1}^{n} \nabla f_i (\tilde{w}^{s-1})$ \hspace*{\fill} // calculate full gradient
		\FOR{$k=0,1,...,(m-1)$}
		\STATE Randomly select one mini-batch $B_k$ from [n].
		\STATE Calculate $\tilde{\nabla} f_{B_k} (w^s_k)= \dfrac{1}{b} \sum_{i \in B_k} \nabla f_i (w^s_k) - \dfrac{1}{n}\sum_{i \in B_k} \nabla f_i (\tilde{w}^{s-1}) + \tilde{\mu}^s$.
		\STATE Calculate $\eta_k^s$ using stochastic backtracking-Armijo line search on $B_k$.		
		\STATE Option I (smooth): $ w^s_{k+1} = w^s_k - \eta_k^s \left[ \tilde{\nabla} f_{B_k} (w^s_k) + \nabla g(w^s_k) \right]$
		\STATE Option II (non-smooth): $w^s_{k+1} = Prox^g_{\eta_k^s} \left( w^s_k - \eta_k^s  \tilde{\nabla} f_{B_k} (w^s_k) \right)$
		\ENDFOR
		\STATE $\tilde{w}^{s}= 1/m \sum_{i=1}^{m} w_i^{s}$ \hspace*{\fill} // iterate averaging
		\STATE $w^{s+1}_0 = w^s_m$ \hspace*{\fill} //initialize starting point
		\ENDFOR
		\STATE \textbf{Output:} $w^S_m$
	\end{algorithmic}
\end{algorithm}
SAAG-IV algorithm, represented by Algorithm \ref{algo_saagiv}, takes input as number of epochs ($S$) and number of mini-batches ($m$) of equal size (say) $b$. It initializes $w^1_0 = \tilde{w}^0$. During each inner iteration, it randomly selects one mini-batch $B_k$, calculates two gradients over $B_k$ at last iterate and snap-point, updates $\tilde{\nabla} f_{B_k} (w^s_k)$ and performs stochastic backtracking-Armijo line search (SBAS) over $B_k$. Then parameters are updated using Option I for smooth regularizer and using Option II for non-smooth regularizer. After inner iterations, it calculates average to set the snap point, and uses last iterate as the starting point for the new epoch, as $\tilde{w}^{s}= 1/m \sum_{i=1}^{m} w_i^{s}$ and $w^{s+1}_0 = w^s_m$, respectively.\\
The comparative study of SAAGs is represented by Figure \ref{fig_saags_smooth} for smooth problem ($l_2$-regularized logistic regression), which compares accuracy and suboptimality against training time (in seconds), gradients$/n$ and epochs. The results are reported on Adult dataset with mini-batch of 32 data points. It is clear from all the six criteria plots that results for SAAG-III and IV are very stable than SAAG-I and II, respectively, because of averaging of iterates. SAAG-IV performs better than SAAG-II and SAAG-III performs closely but stably than SAAG-I. Moreover, SAAG-I and SAAG-II stabilize with increase  in mini-batch size but the performance of methods decreases with mini-batch size (see, Appendix for effect of mini-batch sizes on SAAGs). %For larger mini-batches (including large datasets) SAAG-IV performs better than SAAG-II but SAAG-I and SAAG-III perform closely.
% Comparison of SAAG-I and SAAG-II with SAAG-III and IV for smooth problem

%\subsection{Extension to Momentum Acceleration}
%\label{subsec_acceleration}

\section{Analysis}
\label{sec_analysis}
In general, SAAG-IV gives better results for large-scale learning problems as compared to SAAG-III as shown by empirical results presented in Fig. 3, 4 with news20, rcv1 datasets and results in Appendix~\ref{appsub_minibatch_effect_nonsmooth} ‘Effect of mini-batch size’. So, in this section, we have provided convergence rates of SAAG-IV considering all cases of smoothness with strong convexity. Moreover, analysis of SAAG-III represents a typical case due to the biased nature of gradient estimator and the fact that the full gradient is incrementally maintained rather than being calculated at a fix point, like in SAAG-IV. So, analysis of SAAG-III is left open. The convergence rates for all the different combinations of smoothness and strong convexity are given below:
\begin{theorem1}
	\label{theorem_nsc_s}
	Under the assumptions of Lipschitz continuity with smooth regularizer, the convergence of SAAG-IV is given below:
	\begin{equation}
	\label{eq_saag4_conv_nsc_s}
	\mathbb{E}\left[ f(\tilde{w}^s) - f^{*} \right] \le C^s \left[ f(\tilde{w}^0) - f^{*}\right] + V,
	\end{equation}
	where, $C = \left[ \dfrac{4\alpha(b)}{(\beta-1-4\alpha(b))}\dfrac{c}{m} +\dfrac{4\left(\alpha(b)m^2+(m-1)^2\right)}{m^2(\beta-1-4\alpha(b))} \right]<1$ and $V $ is constant.
\end{theorem1}
\begin{theorem1}
	\label{theorem_sc_s}
	Under  the assumptions of Lipschitz continuity and strong convexity with smooth regularizer, the convergence of SAAG-IV method is given below:
	\begin{equation}
	\label{eq_saag4_conv_sc_s}
	\mathbb{E}\left[ f(\tilde{w}^s) - f^{*} \right] \le C^s \left[ f(\tilde{w}^0) - f^{*}\right] + V,
	\end{equation}
	where,\\ $C = \left[\dfrac{cL\beta}{m\mu} + \dfrac{4\left(\alpha(b)m^2+(m-1)^2\right)}{m^2(\beta-1)}   -\dfrac{c(m-1)}{m^2} + \dfrac{4\alpha(b)}{(\beta-1)}\right]\\ \left(1-\dfrac{4\alpha(b)}{(\beta-1)} - c\left( \dfrac{m-1}{m^2} - \dfrac{4\alpha(b)}{(\beta-1)} \dfrac{1}{m}\right)\right)^{-1}<1$ and $V$ is constant.
\end{theorem1}
\begin{theorem1}
	\label{theorem_nsc_ns}
	Under  the assumptions of Lipschitz continuity with non-smooth regularizer, the convergence of SAAG-IV is given below:
	\begin{equation}
	\label{eq_saag4_conv_nsc_ns}
	\mathbb{E} \left[ F(\tilde{w}^{s}) - F(w^{*})\right] \le  C^s \left[ F(\tilde{w}^{0}) - F(w^{*})  \right]  + V,
	\end{equation}
	where,\\ $C = \left( \dfrac{4\alpha(b)}{(\beta-1-4\alpha(b))} \dfrac{c}{m}+ \dfrac{4\left(\alpha(b)m^2+(m-1)^2\right)}{m^2(\beta-1-4\alpha(b))}\right)< 1$ and $V$ is constant.
\end{theorem1}
\begin{theorem1}
	\label{theorem_sc_ns}
	Under  the assumptions of Lipschitz continuity and strong convexity with non-smooth regularizer, the convergence of SAAG-IV is given below:
	\begin{equation}
	\label{eq_saag4_conv_sc_ns}
	\mathbb{E}\left[ F(\tilde{w}^s) - F^{*} \right] \le C^s \left[ F(\tilde{w}^0) - F^{*}\right] + V,
	\end{equation}
	where,\\ $C = \left(\dfrac{Lc\beta}{m\mu} + \dfrac{4c\alpha(b)}{m(\beta-1)} - \dfrac{c(m-1)}{m^2}  + \dfrac{4\left(\alpha(b)m^2+(m-1)^2\right)}{m^2(\beta-1)}\right)\\ \left(1-\dfrac{4\alpha(b)}{(\beta-1)} - \dfrac{c(m-1)}{m^2} + \dfrac{4c\alpha(b)}{m(\beta-1)}\right)^{-1} <1$ and $V$ is constant.
\end{theorem1}
All the proofs are given in the Appendix~\ref{app_proofs} and all these results prove linear convergence (as per definition of convergence) of SAAG-IV for all the four combinations of smoothness and strong-convexity with some initial errors due to the constant terms in the results. SAAGs are based on intuitions from practice \cite{Chauhan2017Saag} and they try to give more importance to the latest gradient values than the older gradient values, which make them biased techniques and results into this extra constant term. This constant term signifies that SAAGs converge to a region close to the solution, which is very practical because all the machine learning algorithms are used to solve the problems approximately and we never find an exact solution for the problem \cite{Bottou2007Tradeoffs}, because of computational difficulty. Moreover, the constant term pops up due to the mini-batched gradient value at optimal point, i.e., $\dfrac{1}{|B_k|} \sum_{i \in B_k} \nabla f_i(w^{*})$. If the size of the mini-batch increases and eventually becomes equal to the dataset then this constant becomes equal to full gradient and vanishes, i.e., $\dfrac{1}{n} \sum_{i=1}^{n} \nabla f_i(w^{*}) = 0.$ \\
\indent The linear convergence for all combinations of strong convexity and smoothness of the regularizer, is the maximum rate exhibited by the first order methods without curvature information. SAG, SVRG, SAGA and VR-SGD also exhibit linear convergence for the strong convexity and smooth problem but except VR-SGD, they don't cover all the cases, e.g., SVRG does not cover the non-strongly convex cases. However, the theoretical results provided by VR-SGD, prove linear convergence for strongly convex cases, like our results, but VR-SGD provides only $O(1/T)$ convergence for non-strongly convex cases, unlike our linear convergence results.

\section{Experimental Results}
\label{sec_experiments}
In this section, we have presented the experimental results\footnote{experimental results can be reproduced using the code available at link: \url{https://drive.google.com/open?id=1Rdp_pmHLQAA9OBxBtHzz6FCduCypAzhd}}. SAAG-III and IV are compared against the most widely used variance reduction method, SVRG and one of the latest method VR-SGD which has been proved to outperform existing techniques. The results have been reported in terms of suboptimality and accuracy against time, epochs and gradients$/n$. The SAAGs can be applied to strongly and non-strongly convex problems with smooth or non-smooth regularizers. But the results have been reported with strongly convex problems with and without smoothness because problems can be easily converted to strongly convex problems by adding $l_2$-regularization.

\subsection{Experimental Setup}
\label{subsec_setup}
The experiments are reported using six different criteria which plot suboptimality (objective minus best value) versus epochs (where one epoch refers to one pass through the dataset), suboptimality versus gradients$/n$, suboptimality versus time, accuracy versus time, accuracy versus epochs and accuracy versus gradients$/n$. The x-axis and y-axis data are represented in linear and log scale, respectively. The experiments use the following binary datasets: rcv1 ($\#$data - 20, 242, $\#$features - 47, 236), news20 ($\#$data - 19, 996, $\#$features - 1, 355, 191), real-sim ($\#$data - 72,309, $\#$features - 20, 958) and Adult (also called as a9a, $\#$data - 32,561 and $\#$features - 123), which are available from the LibSVM website\footnote{\url{https://www.csie.ntu.edu.tw/\~cjlin/libsvmtools/datasets/}}. All the datasets are divided into 80\% and 20\% as training and test dataset, respectively. The value of regularization parameter is set as $\lambda = 1*10^{-5} (\text{including}\; \lambda_1, \lambda_2)$ for all the algorithms. The parameters for, Stochastic Backtracking-Armijo line Search (SBAS), are set as: $\alpha = 0.1, \beta = 0.5$ and learning rate is initialized as, $\eta=1.0$. The inner iterations are set as, $m= n/b$ (as used in \cite{Shang2018}). Moreover, in SBAS, algorithms looks for maximum 10 iterations and after that it returns the current value of learning rate if it reduces the objective value otherwise it returns 0.0. This is done to avoid sticking in the algorithm because of stochastic line search. All the experiments have been conducted on MacBook Air (8 GB 1600 MHz DDR3, 1.6 GHz Intel Core i5 and 256GB SSD) using MEX files.

\subsection{Results with Smooth Problem}
\label{subsec_l2lr}
\begin{figure}[htb]
	\subfloat{\includegraphics[width=.332\linewidth]{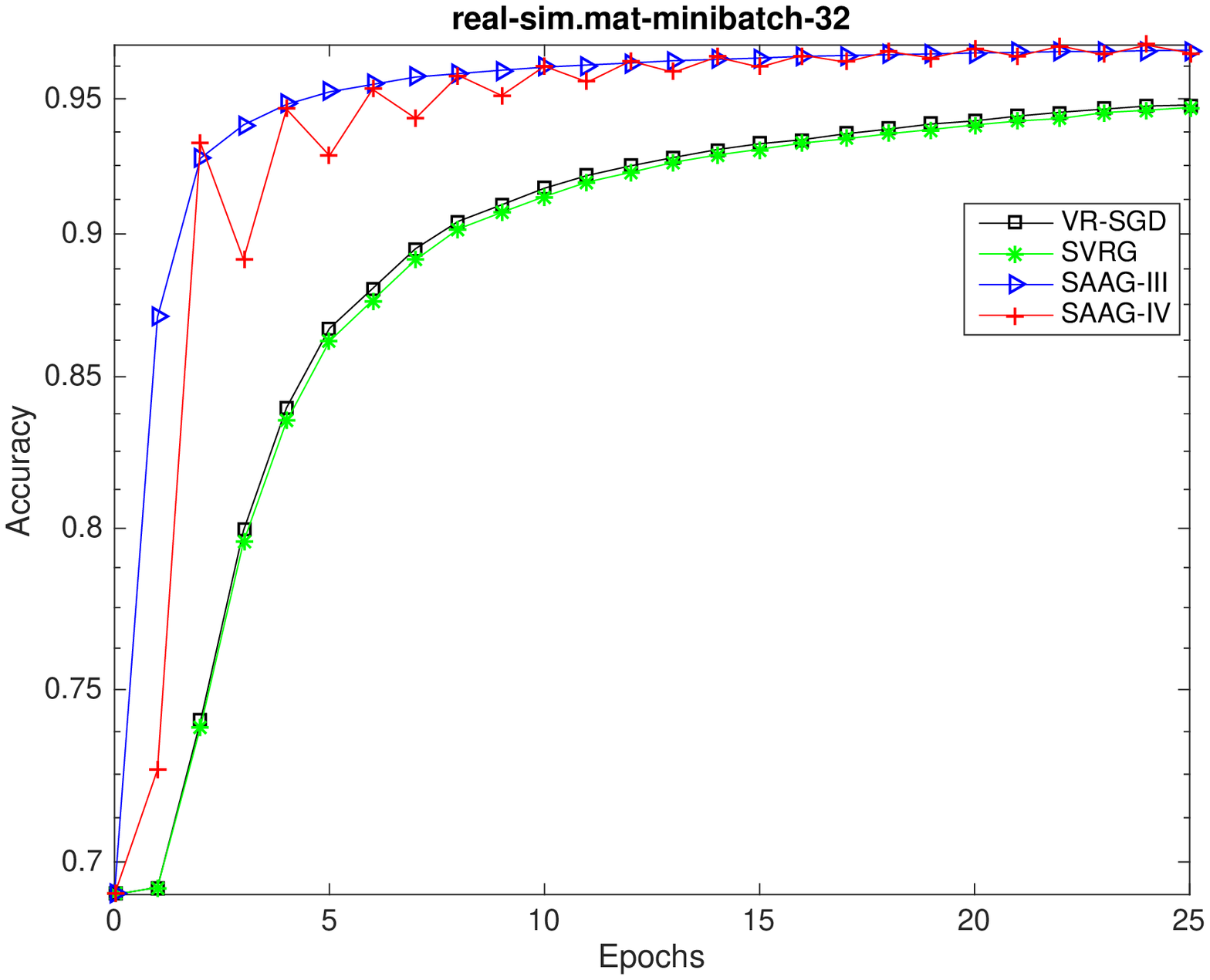}}
	\subfloat{\includegraphics[width=.332\linewidth]{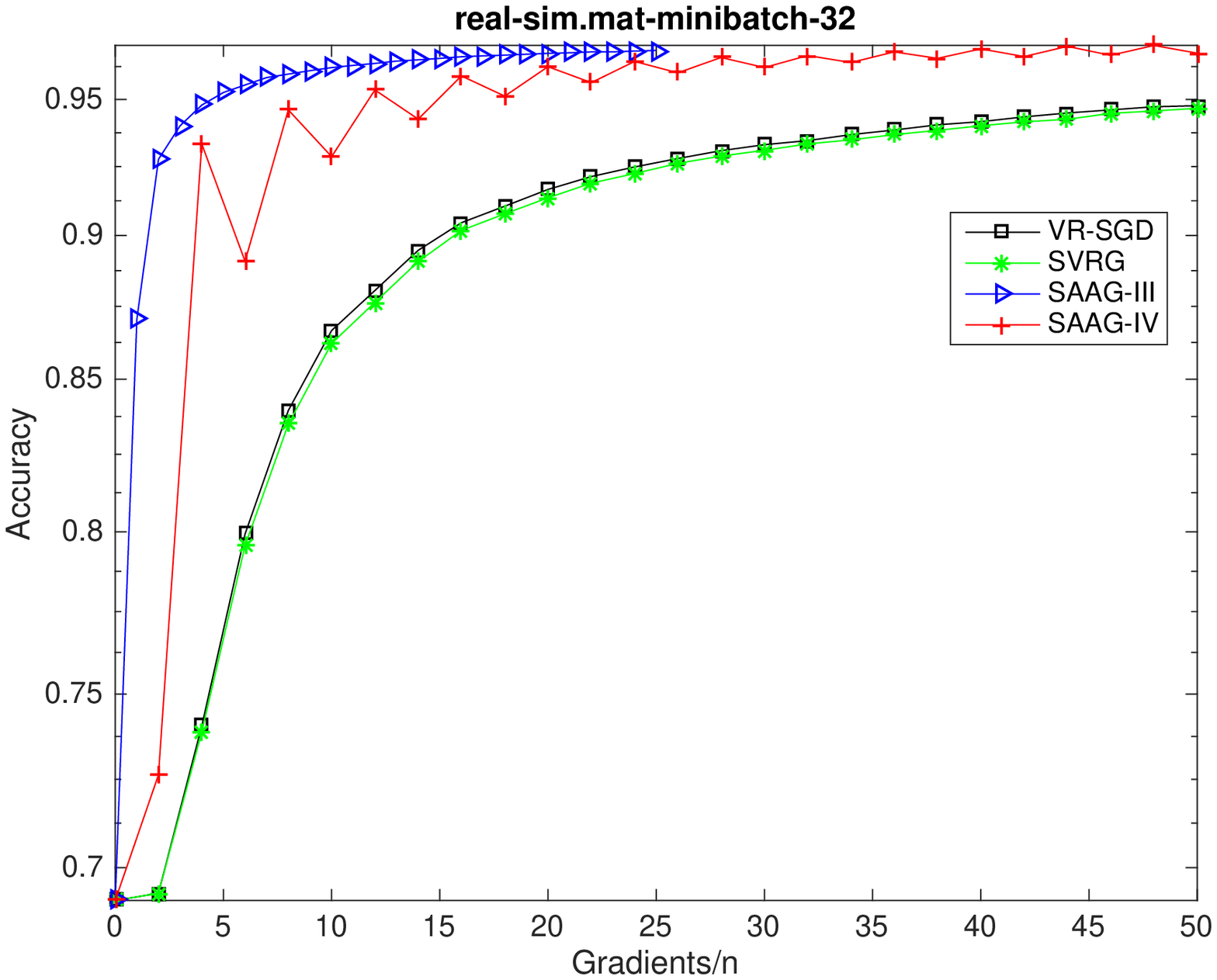}}
	\subfloat{\includegraphics[width=.332\linewidth]{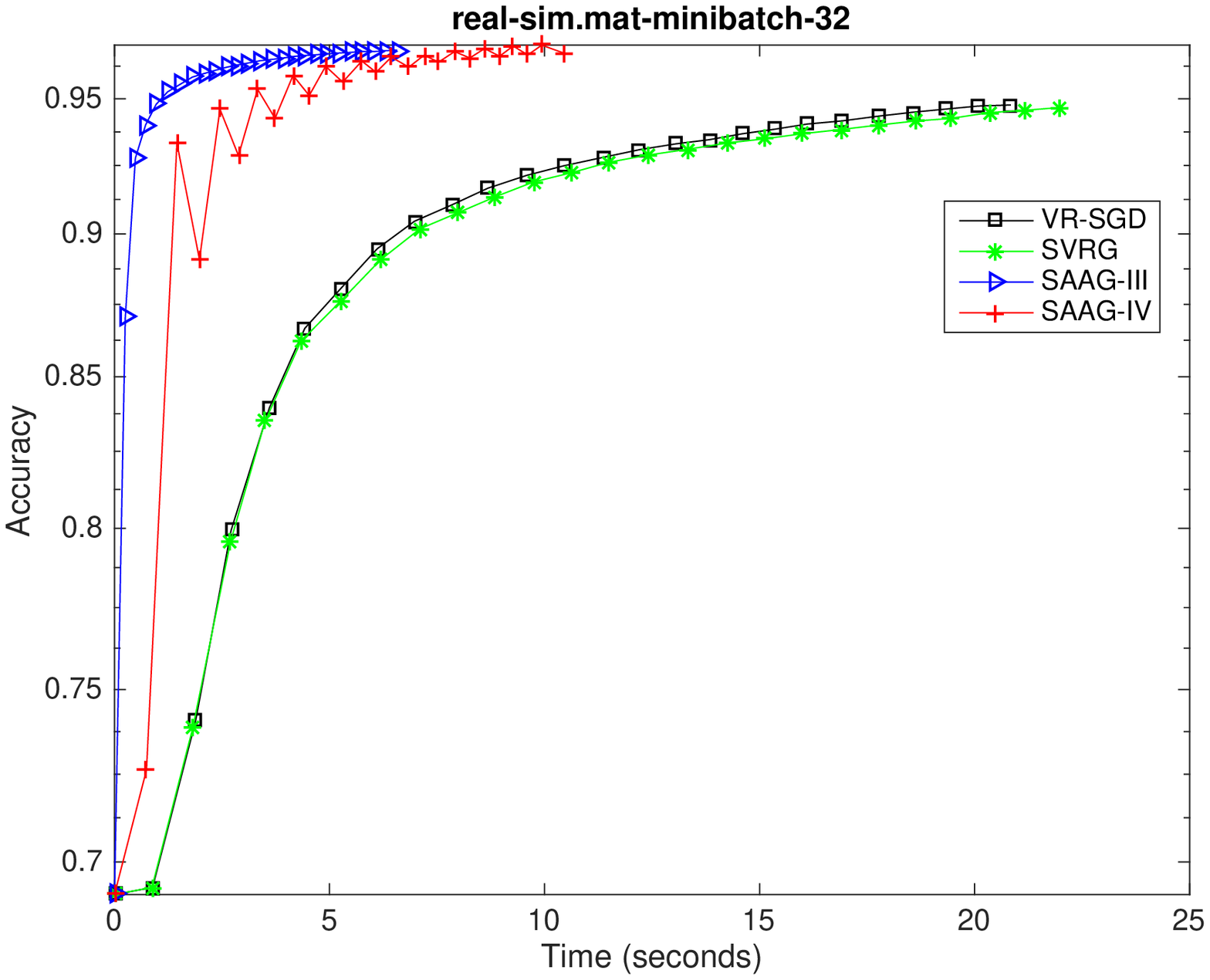}}
	
	\subfloat{\includegraphics[width=.332\linewidth]{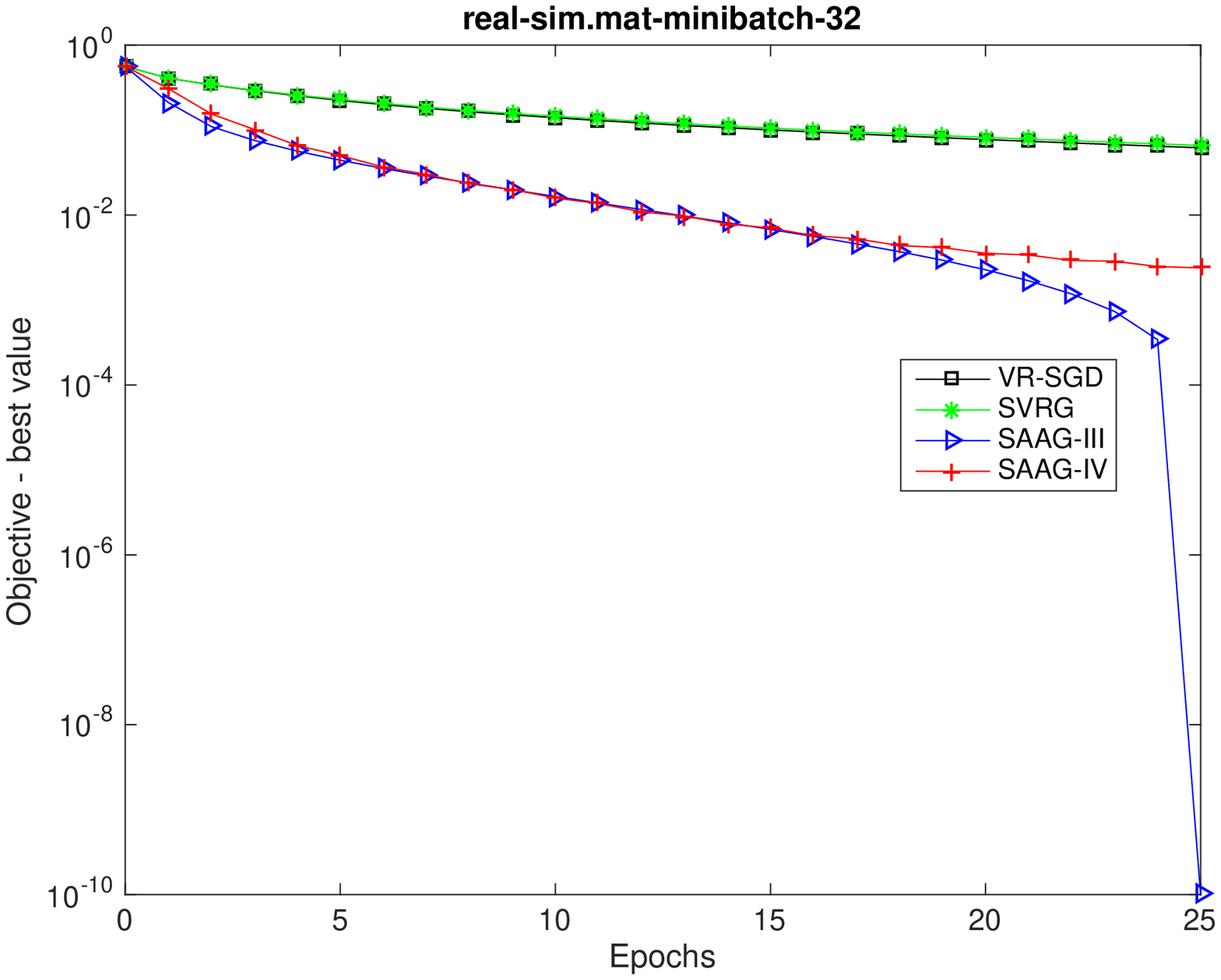}}
	\subfloat{\includegraphics[width=.332\linewidth]{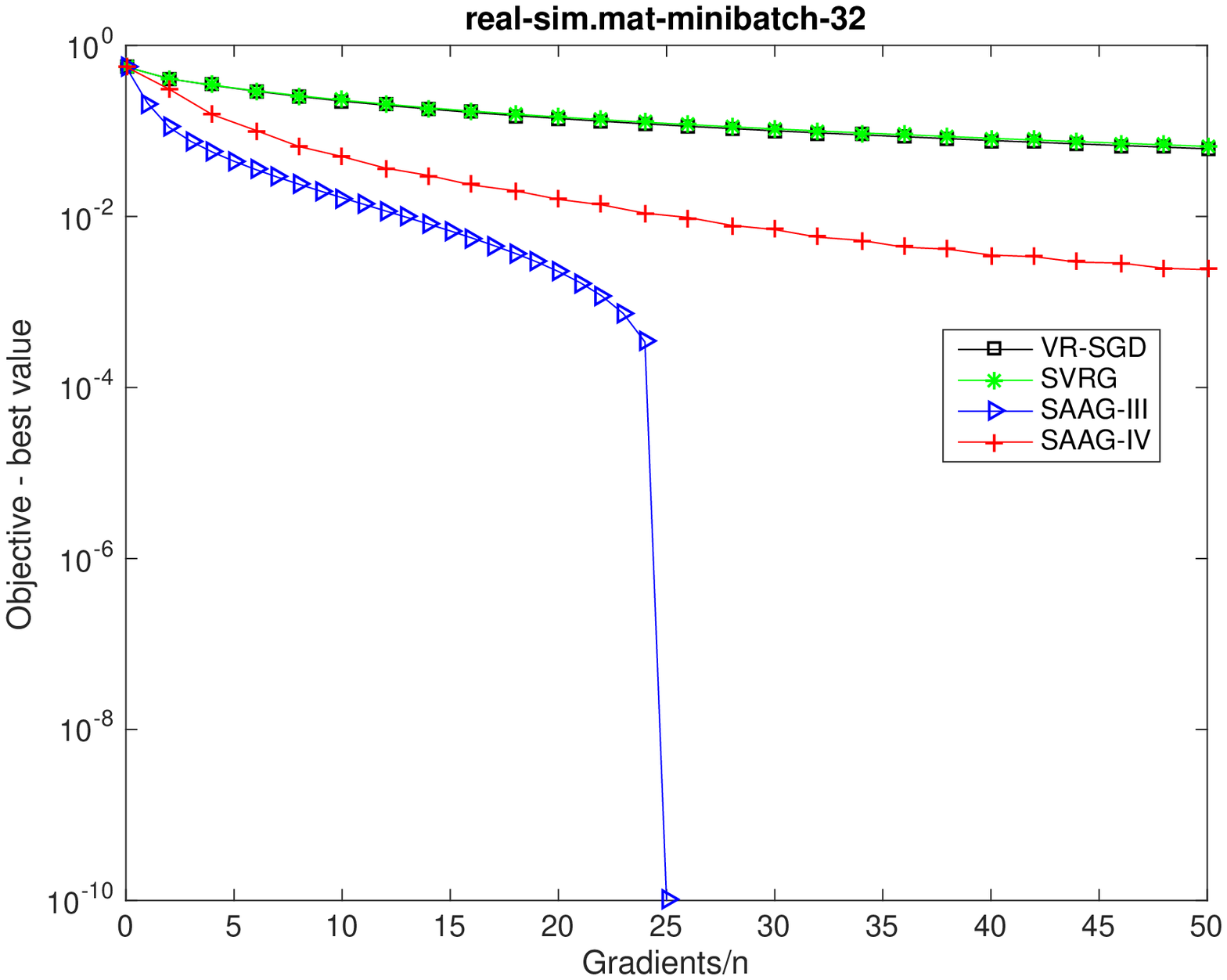}}
	\subfloat{\includegraphics[width=.332\linewidth]{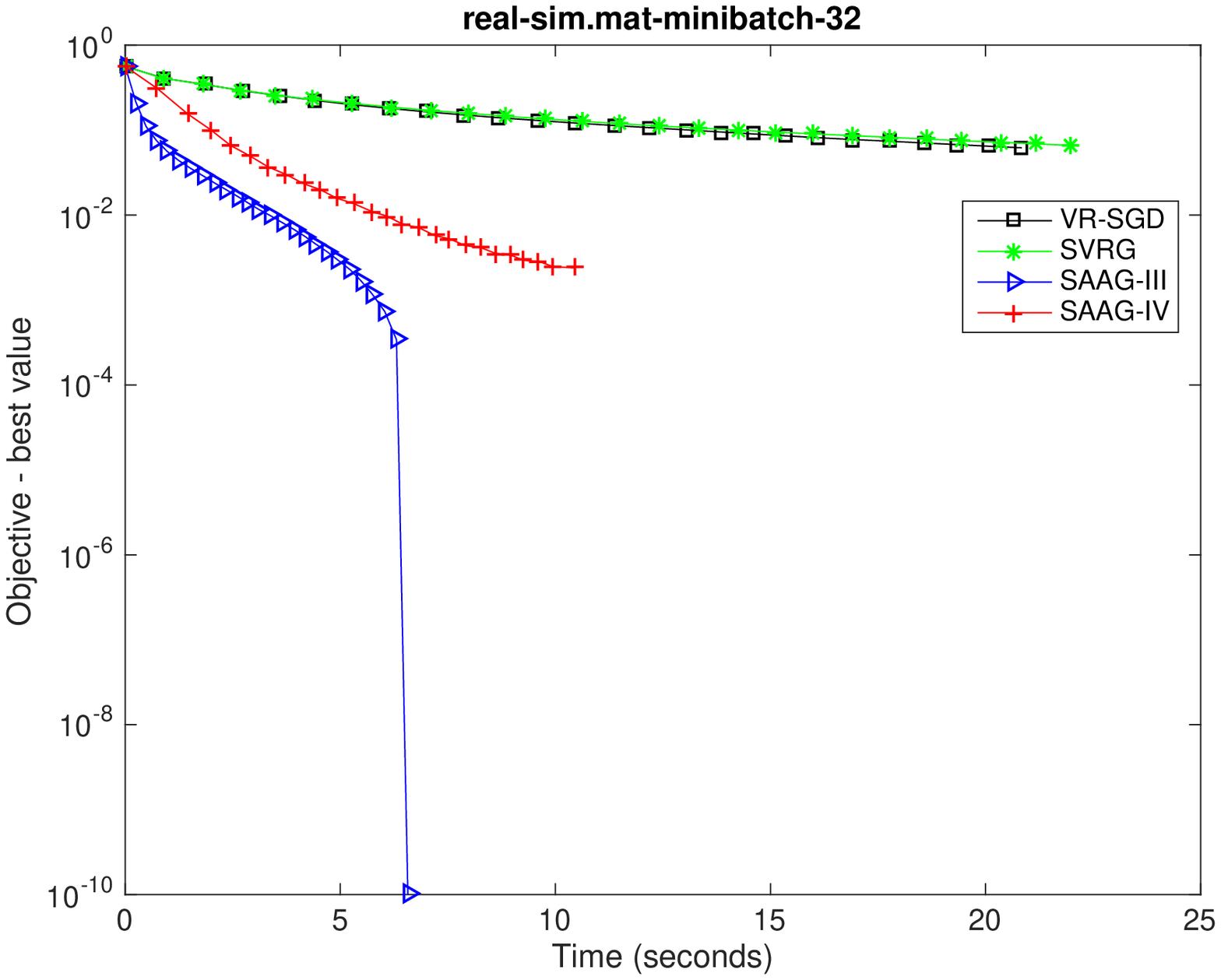}}
	
	\caption{Comparison of SAAG-III, IV, SVRG and VR-SGD on smooth problem using real-sim dataset with mini-batch of 32 data points. First row compares accuracy against epochs, gradients/n and time, and second row compares suboptimality against epochs, gradients/n and time.}
	\label{fig_1}
\end{figure}

\begin{figure}[htb]
	
	\subfloat{\includegraphics[width=.332\linewidth]{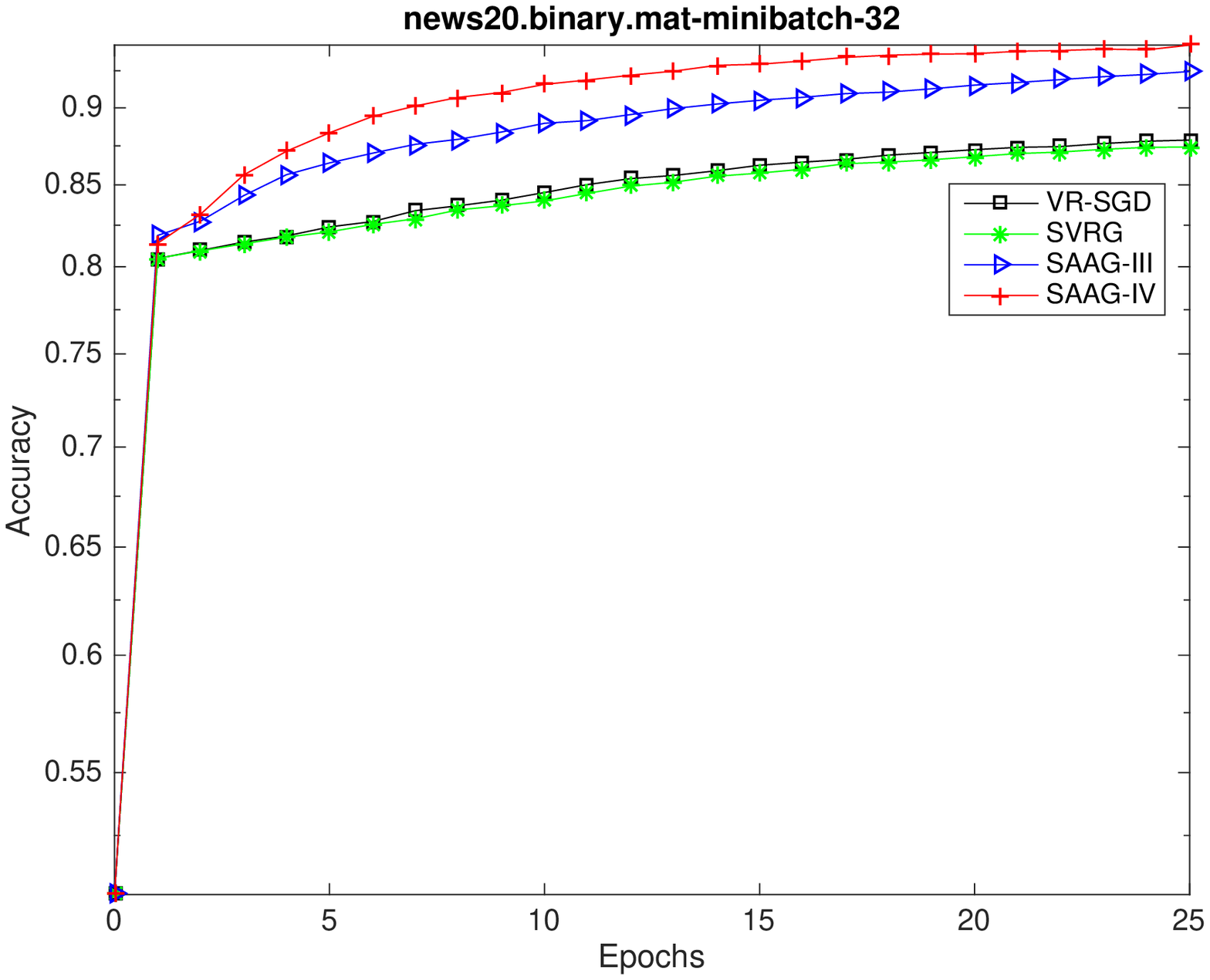}}
	\subfloat{\includegraphics[width=.332\linewidth]{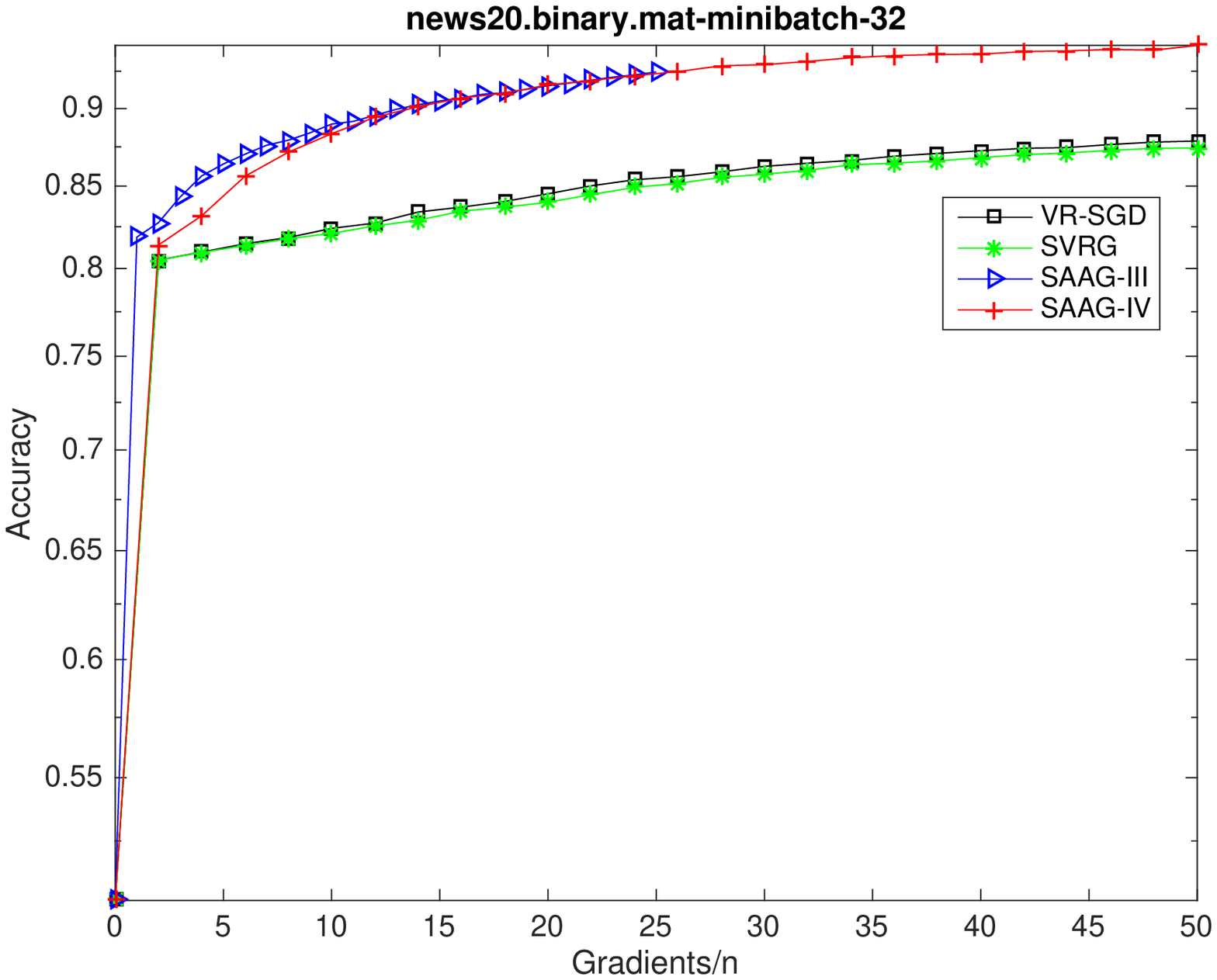}}
	\subfloat{\includegraphics[width=.332\linewidth]{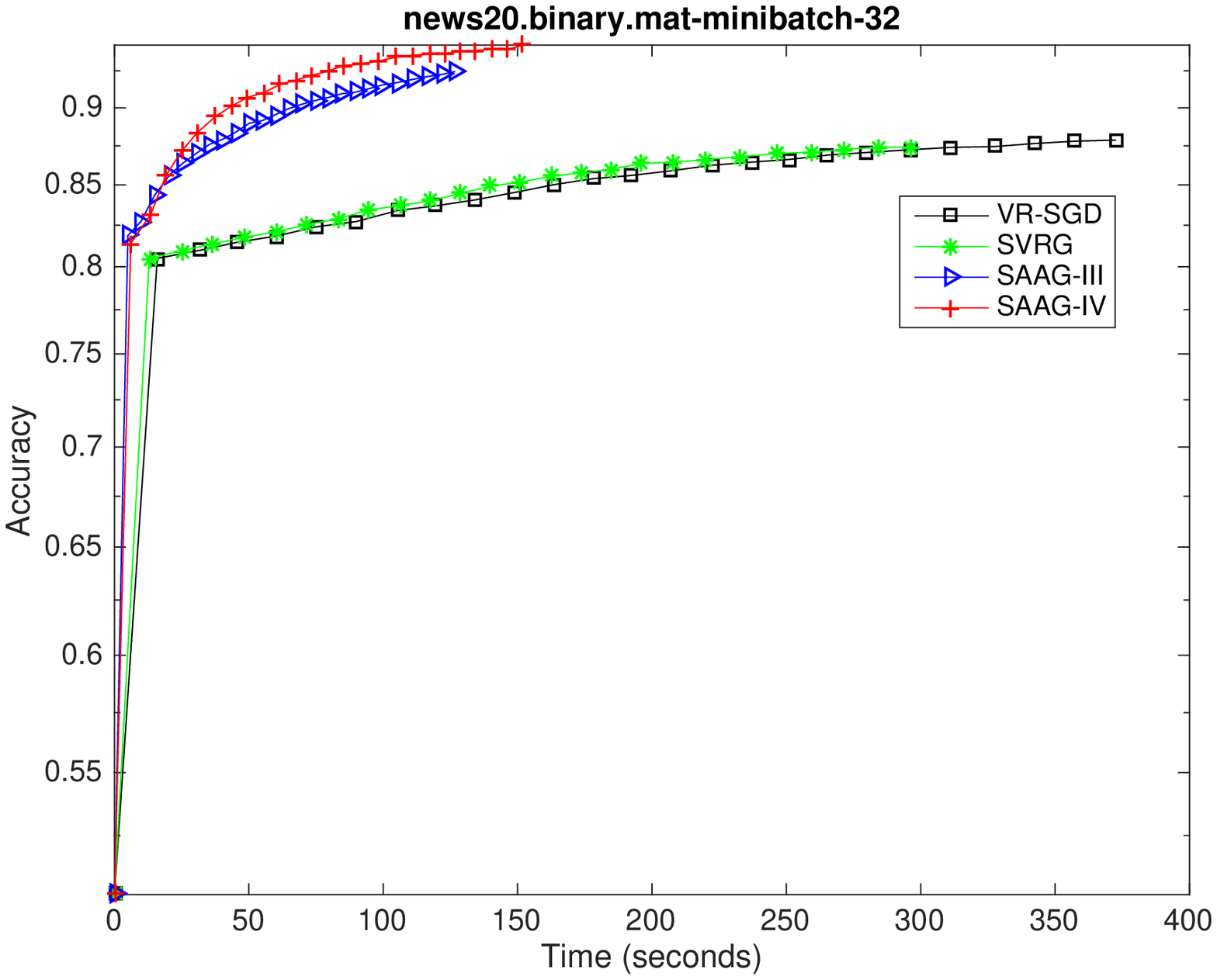}}
	
	\subfloat{\includegraphics[width=.332\linewidth]{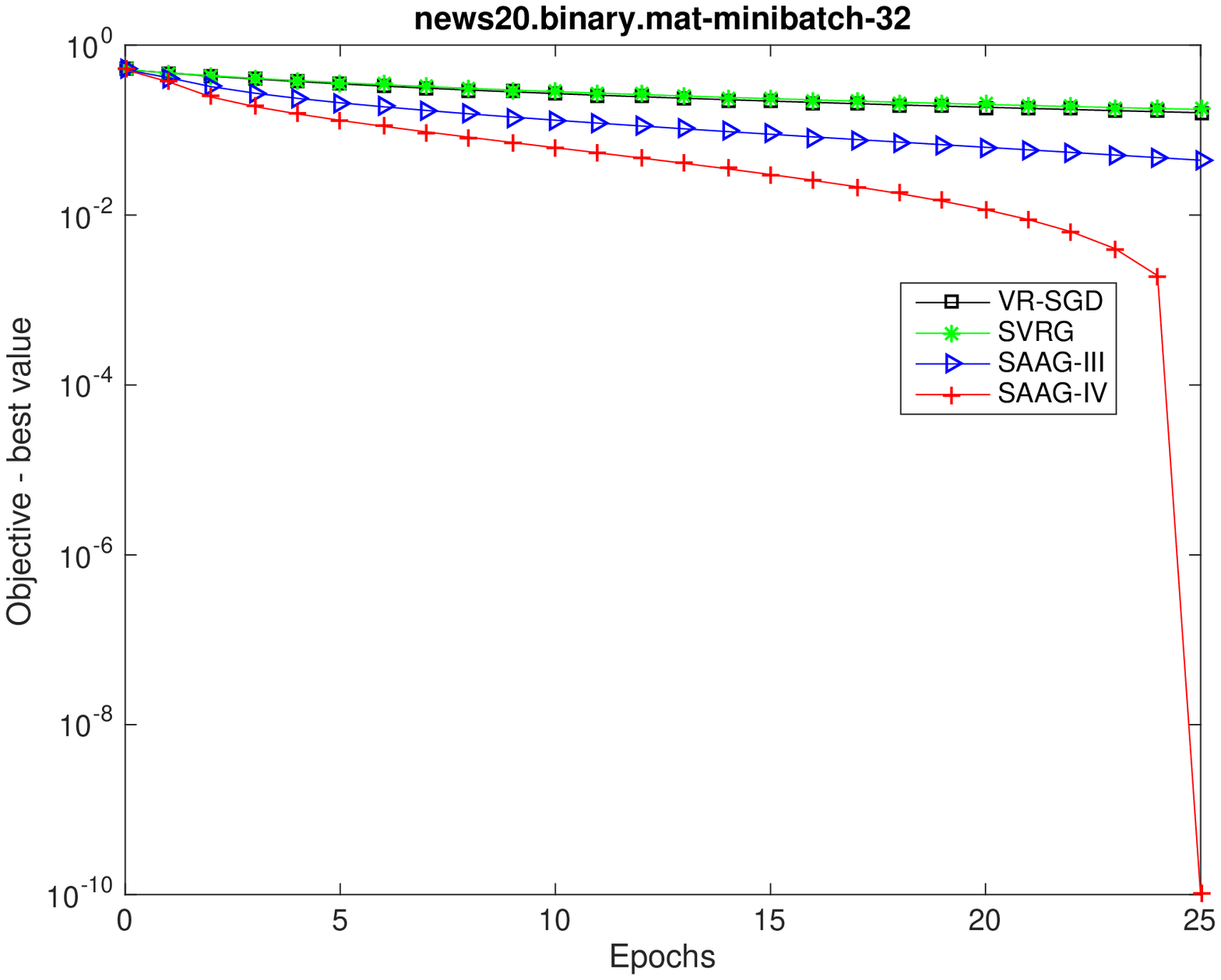}}
	\subfloat{\includegraphics[width=.332\linewidth]{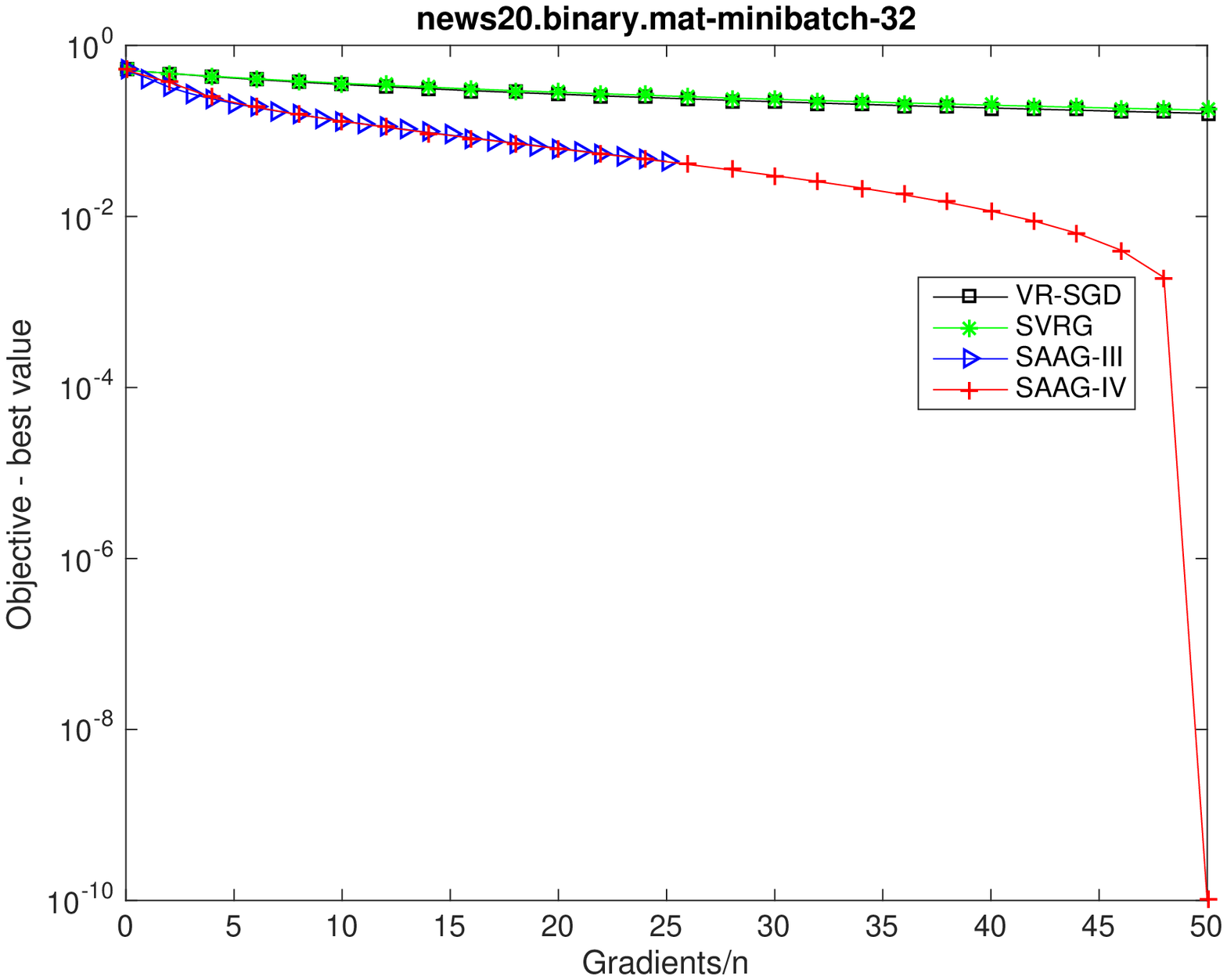}}
	\subfloat{\includegraphics[width=.332\linewidth]{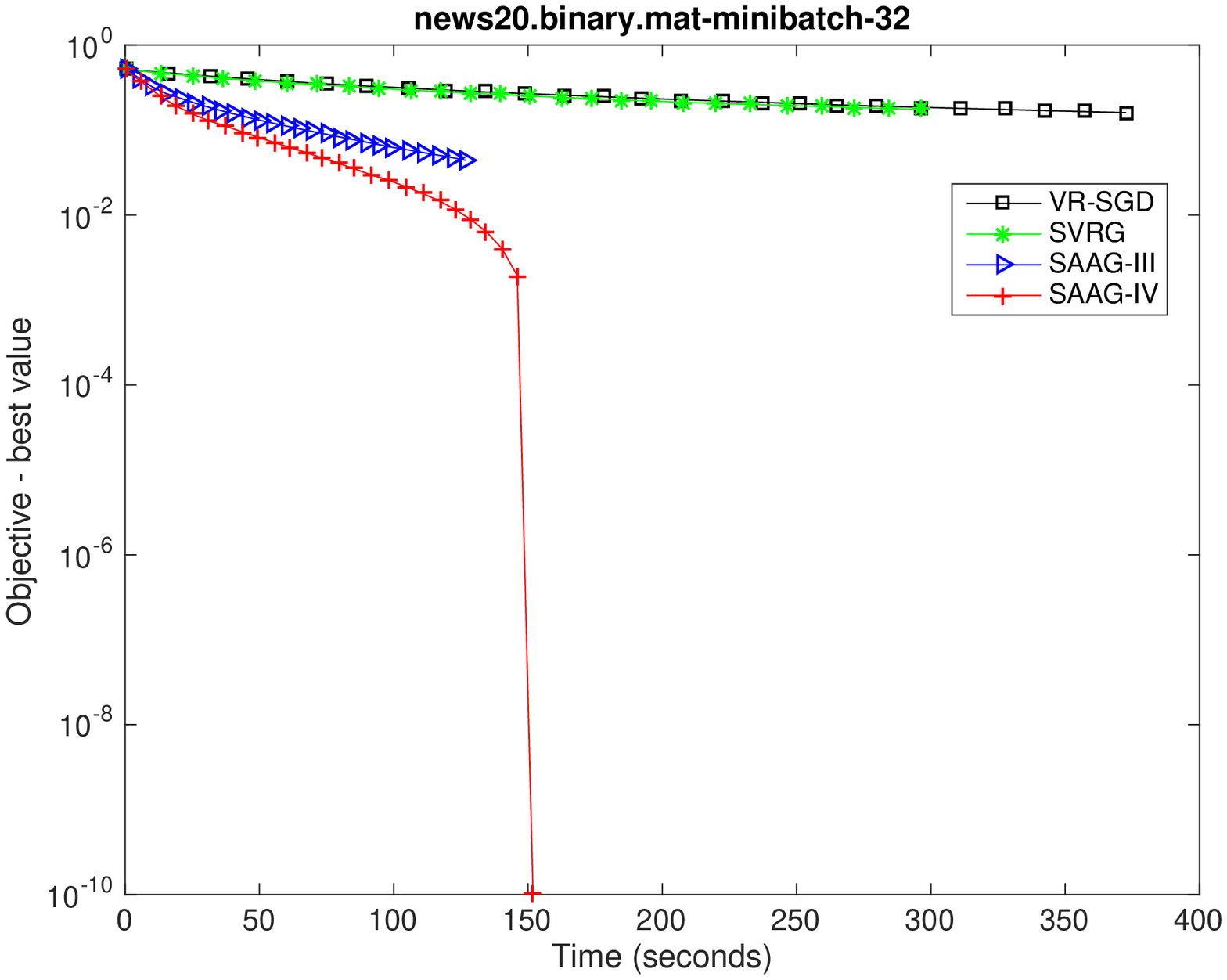}}
	
	\caption{Comparison of SAAG-III, IV, SVRG and VR-SGD on smooth problem using news20 dataset with mini-batch of 32 data points. First row compares accuracy against epochs, gradients/n and time, and second row compares suboptimality against epochs, gradients/n and time.}
	\label{fig_2}
\end{figure}

The results are reported with $l_2$-regularized logistic regression problem as given below:
\begin{equation}
\label{eq_l2lr}
\underset{w}{\min} \; F(w) = \dfrac{1}{n} \sum_{i=1}^{n} \log\left( 1 + \exp\left( - y_i w^T x_i \right) \right) + \dfrac{\lambda}{2} \|w\|^2.
\end{equation}
Figure \ref{fig_1} represents the comparative study of SAAG-III, IV, SVRG and VR-SGD on real-sim dataset. As it is clear from the first row of the figure, SAAG-III and IV give better accuracy and attain the results faster than other the other methods. From the second row of the figure, it is clear that SAAGs converges faster than SVRG and VR-SGD. Moreover, SAAG-III performs better than SAAG-IV and VR-SGD performs slightly better than SVRG as established in \cite{Shang2018}. Figure \ref{fig_2} reports results with news20 dataset and as depicted in the figure the results are similar to real-sim dataset Fig.~(\ref{fig_1}). SAAGs give better accuracy and converge faster than SVRG and VR-SGD methods, but SAAG-IV gives best results. This is because as the mini-batch size or the dataset size increases, SAAG-II and SAAG-IV perform better (as reported in \cite{Chauhan2017Saag}).
% Comparison of SAAG-III and IV with SVRG and VR-SGD

% Comparison of SAAG-III and IV with SVRG and VR-SGD

\subsection{Results with non-smooth Problem}
\label{subsec_nonsmooth}
\begin{figure}[htb]
	\subfloat{\includegraphics[width=.332\linewidth]{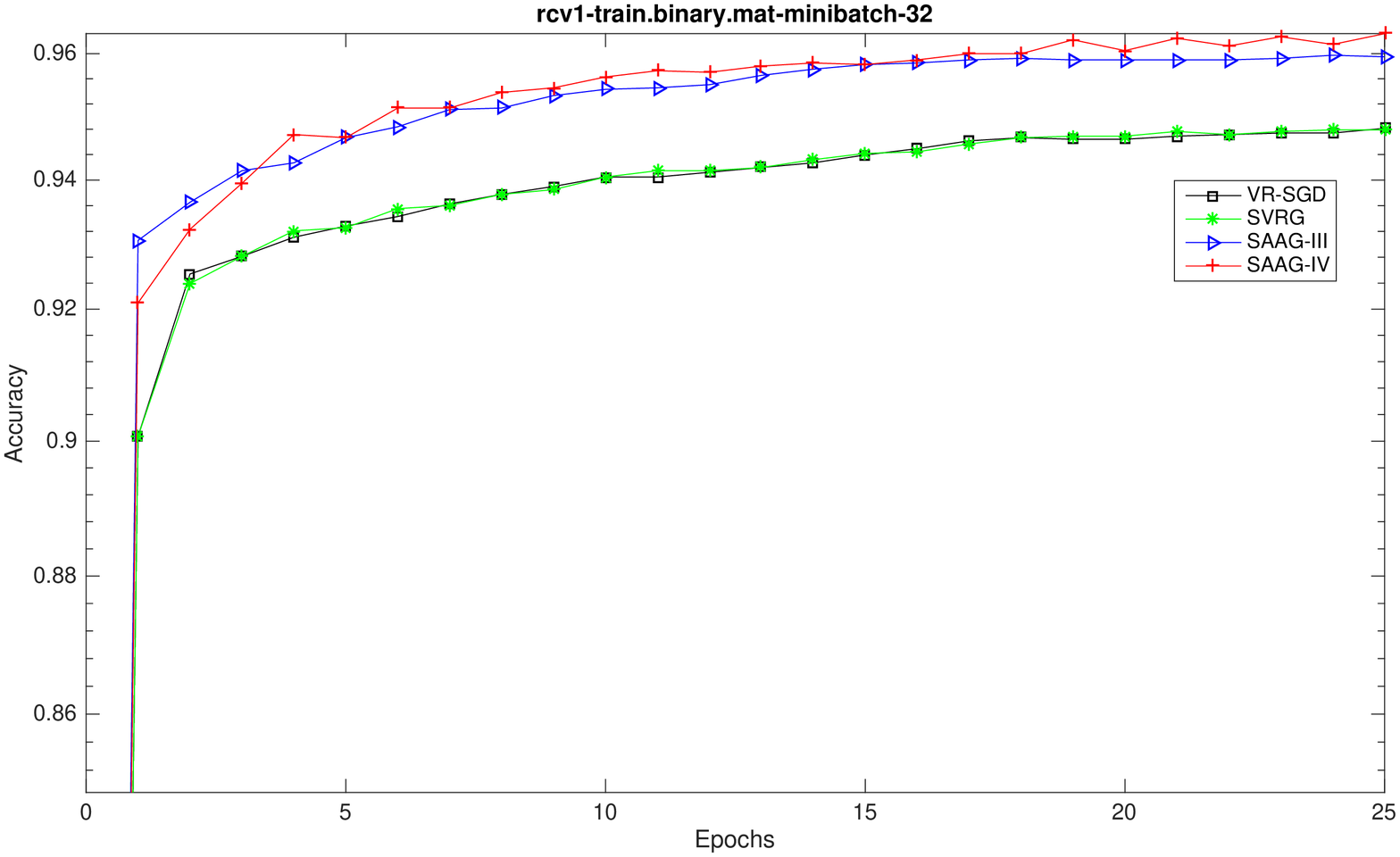}}
	\subfloat{\includegraphics[width=.332\linewidth]{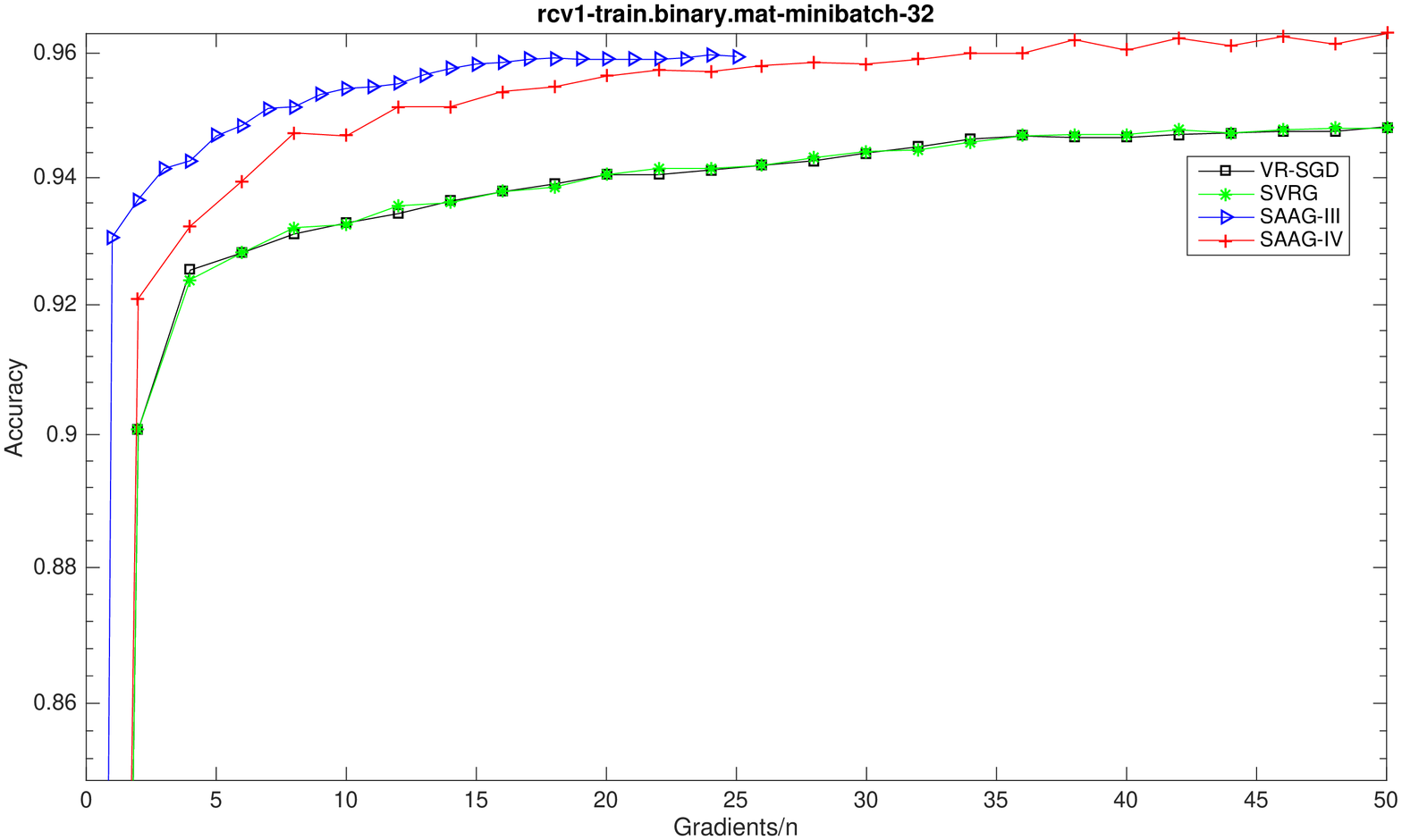}}
	\subfloat{\includegraphics[width=.332\linewidth]{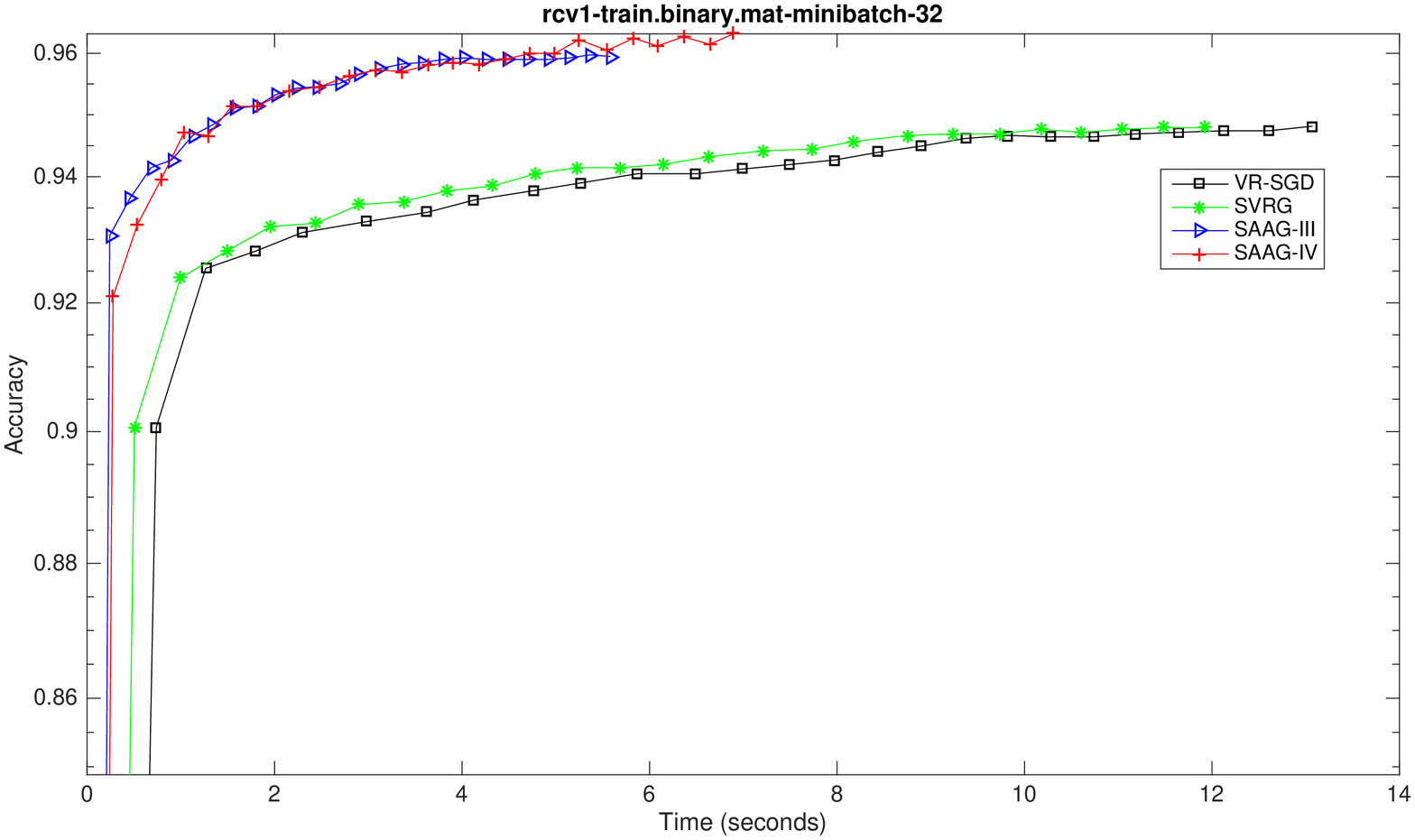}}
	
	\subfloat{\includegraphics[width=.332\linewidth]{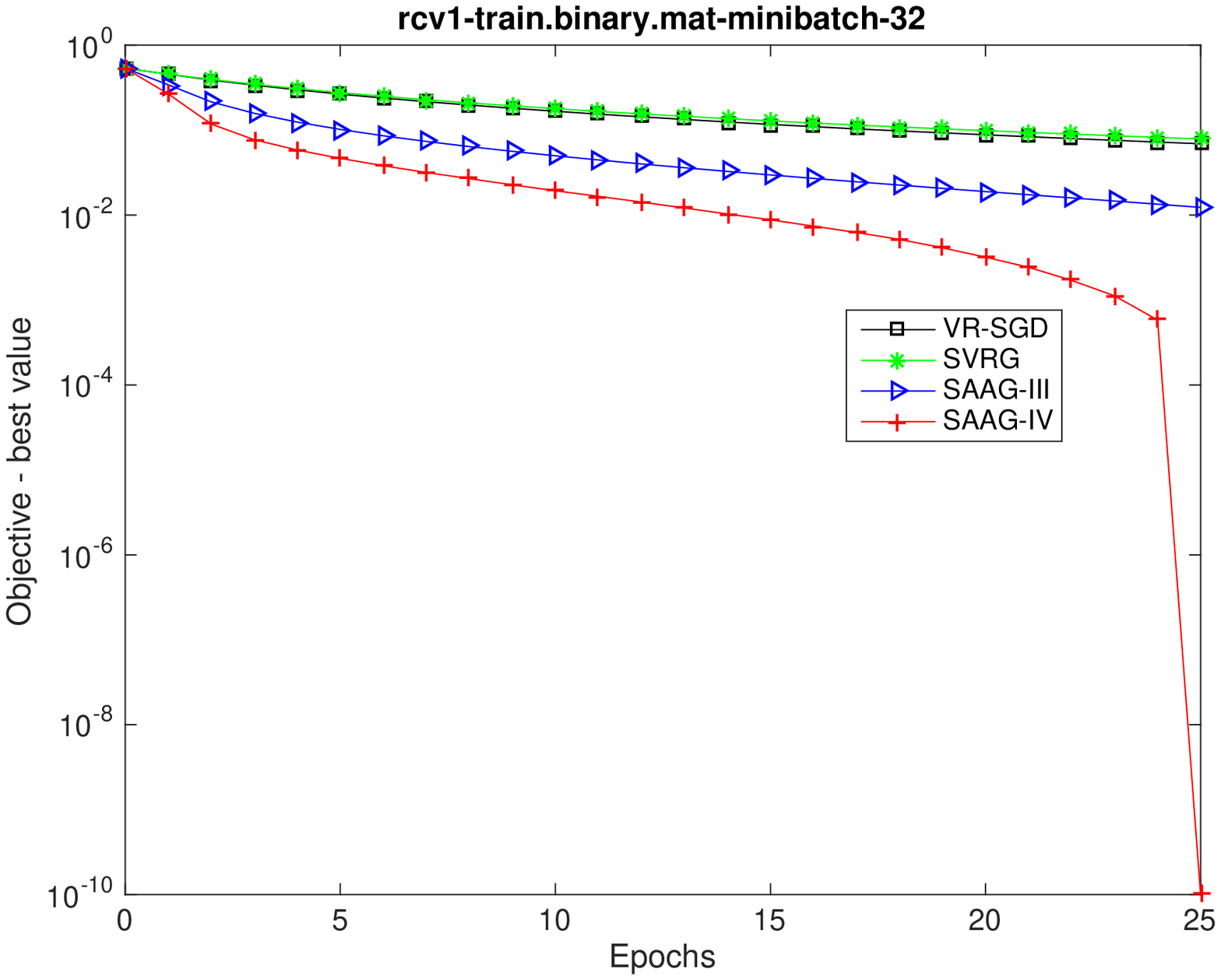}}
	\subfloat{\includegraphics[width=.332\linewidth]{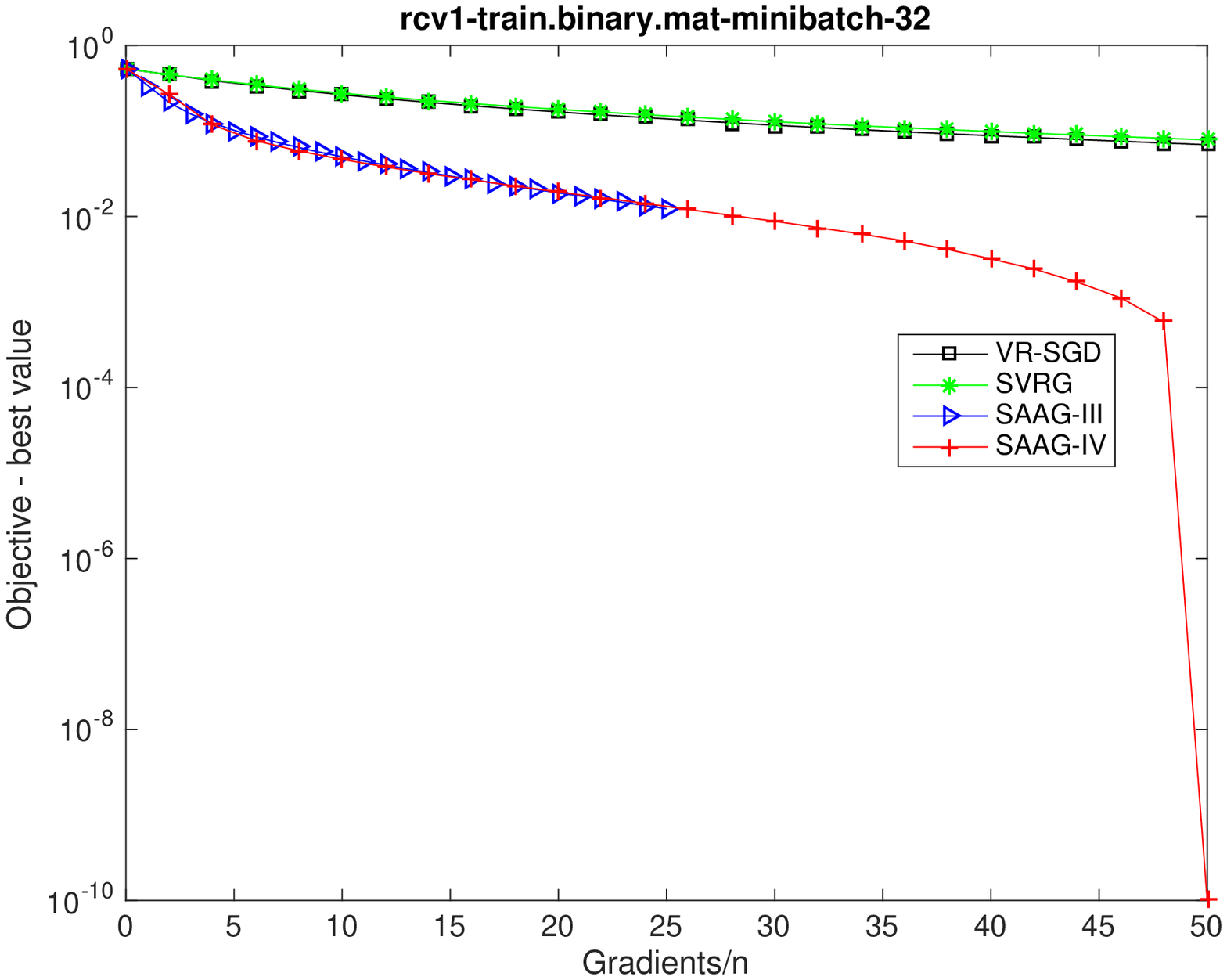}}
	\subfloat{\includegraphics[width=.332\linewidth]{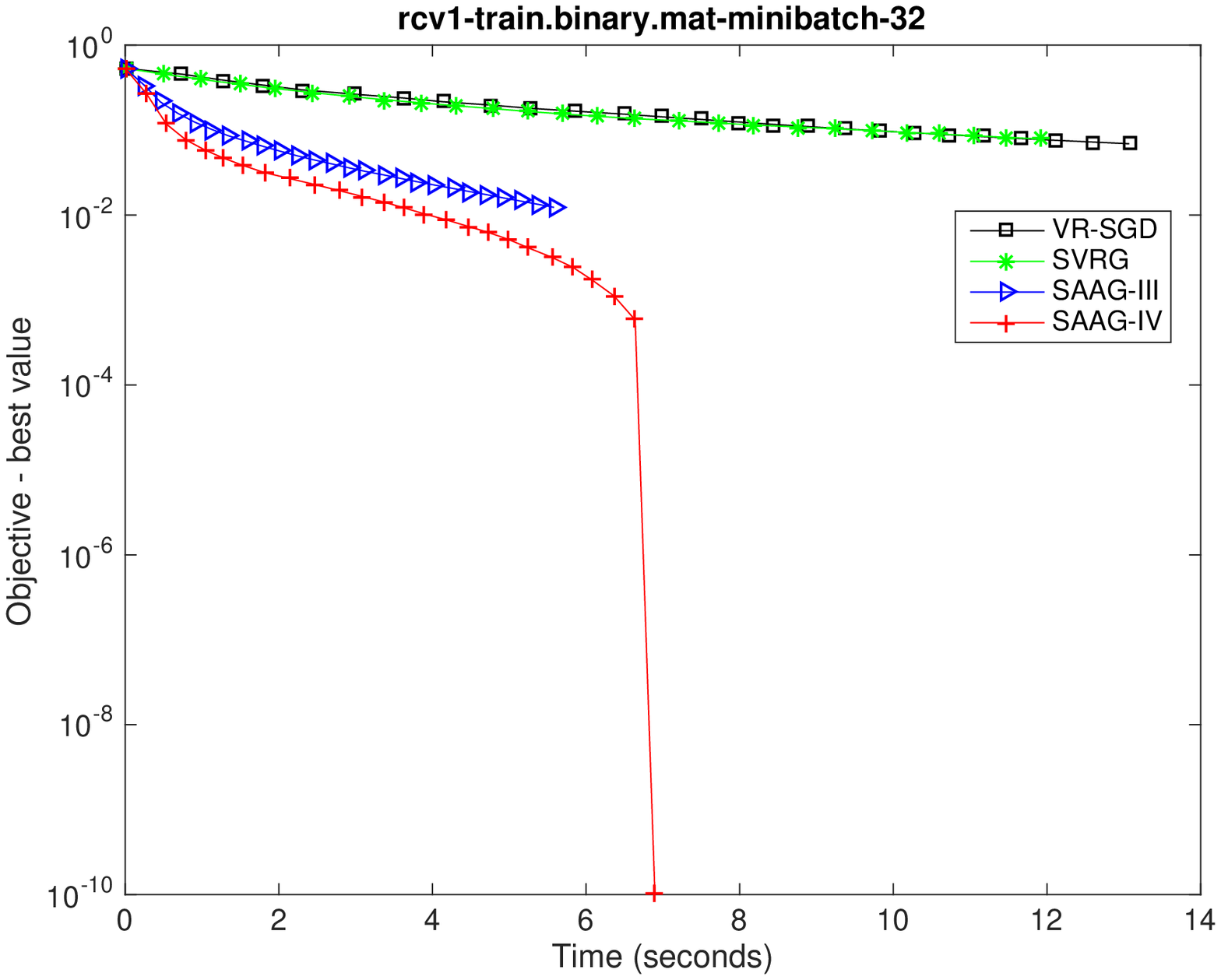}}
	
	\caption{Comparison of SAAG-III, IV, SVRG and VR-SGD on non-smooth problem using rcv1 dataset with mini-batch of 32 data points. First row compares accuracy against epochs, gradients/n and time, and second row compares suboptimality against epochs, gradients/n and time.}
	\label{fig_3}
\end{figure}

\begin{figure}[htb]
	\subfloat{\includegraphics[width=.332\linewidth]{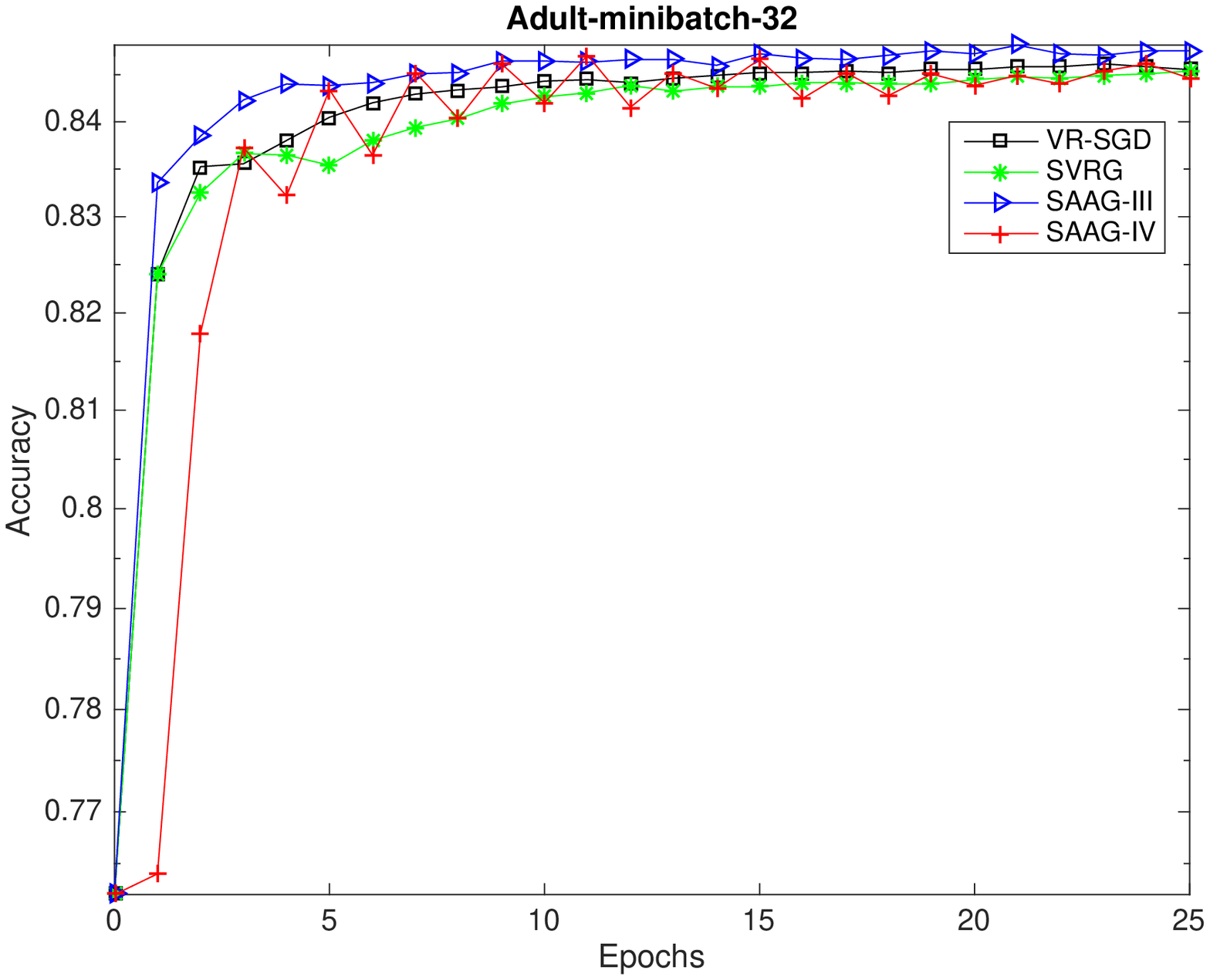}}
	\subfloat{\includegraphics[width=.332\linewidth]{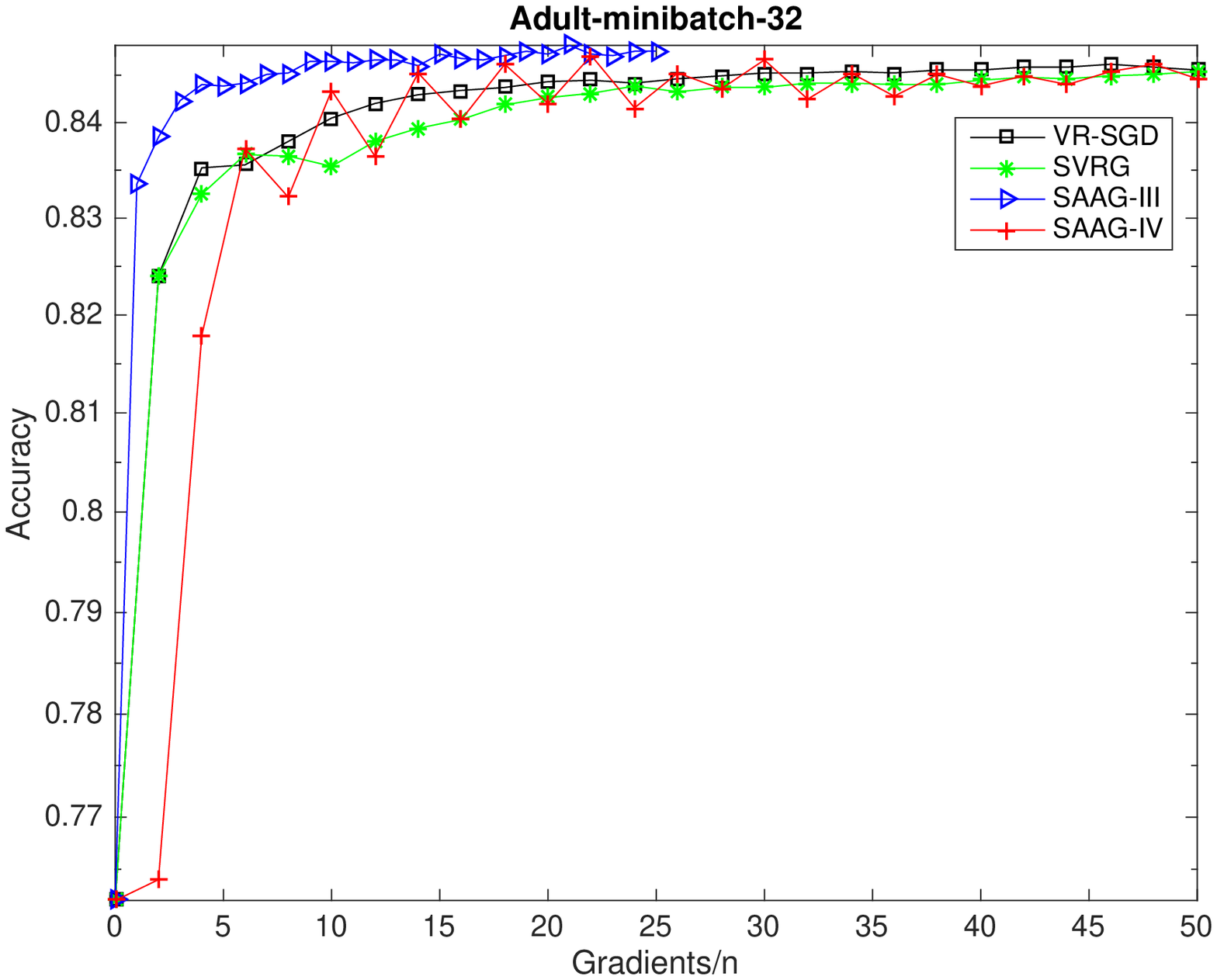}}
	\subfloat{\includegraphics[width=.332\linewidth]{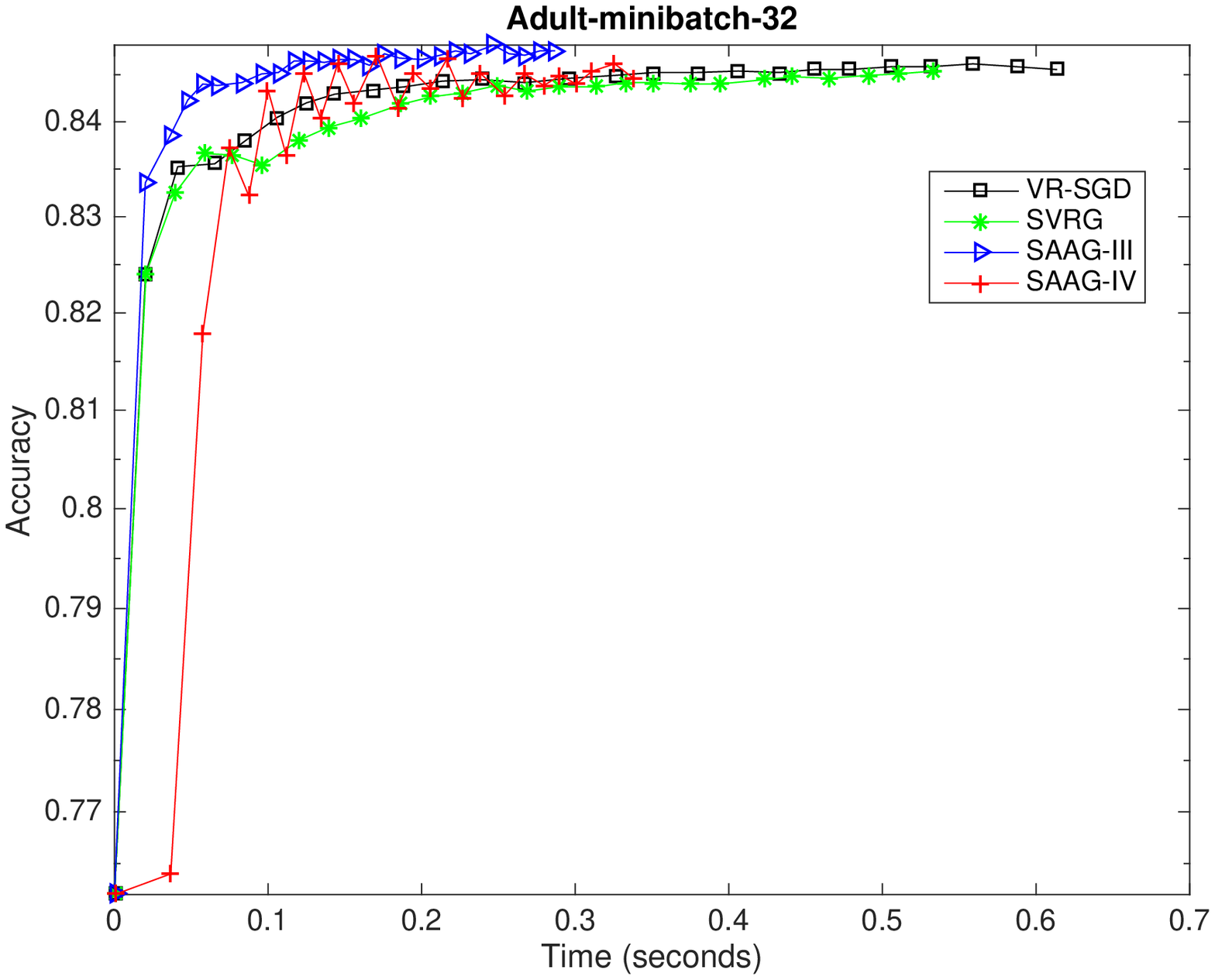}}
	
	\subfloat{\includegraphics[width=.332\linewidth]{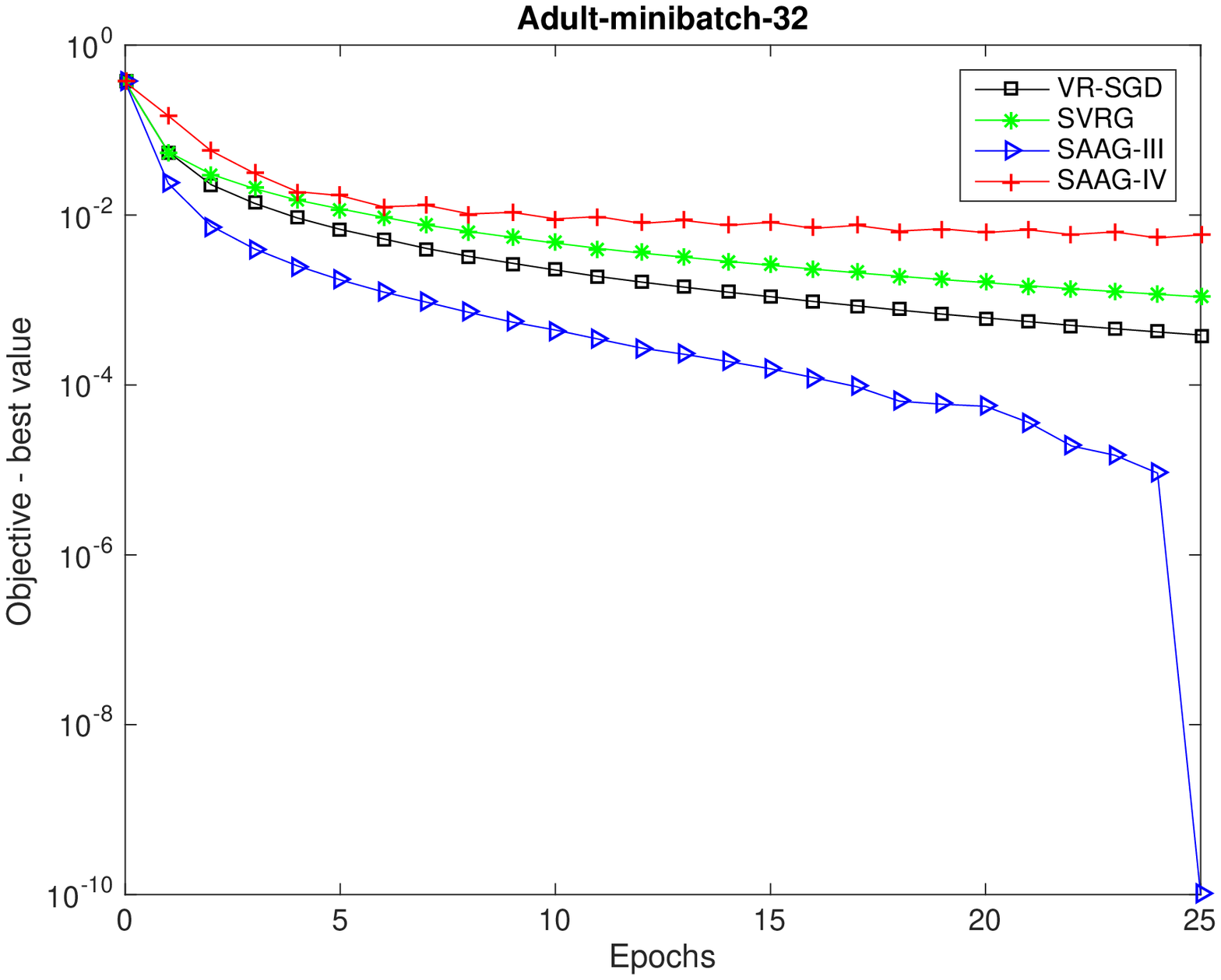}}
	\subfloat{\includegraphics[width=.332\linewidth]{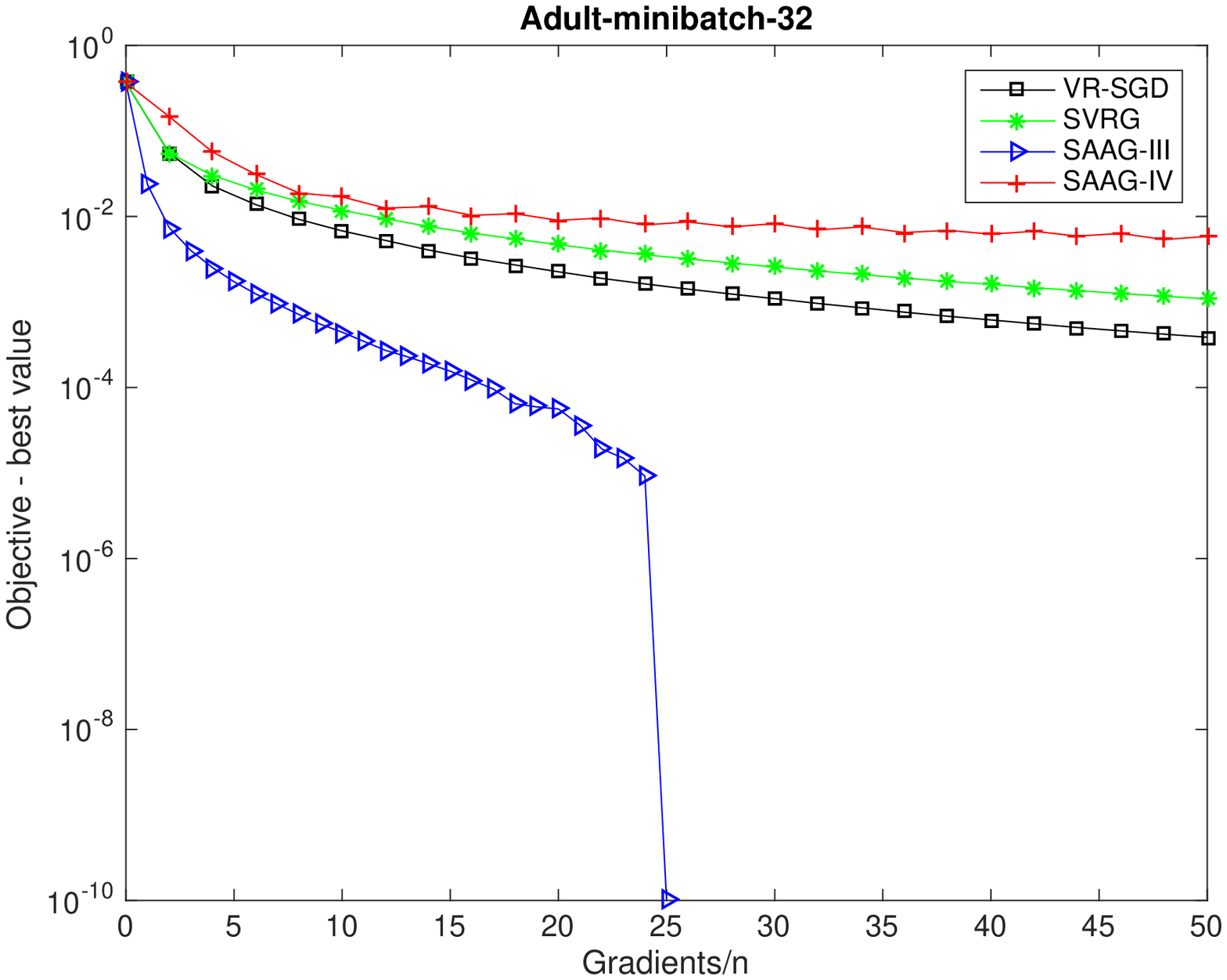}}
	\subfloat{\includegraphics[width=.332\linewidth]{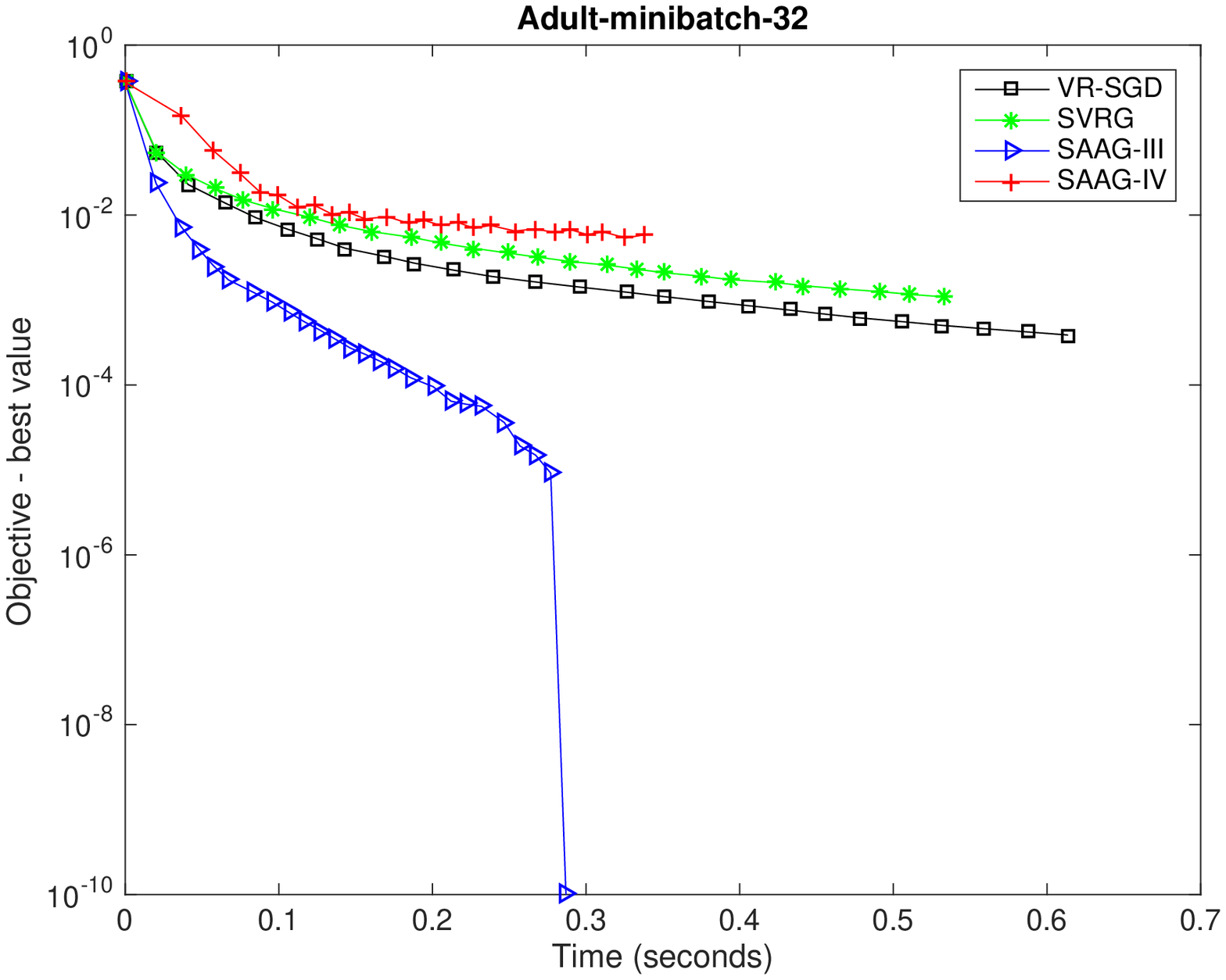}}
	
	\caption{Comparison of SAAG-III, IV, SVRG and VR-SGD on non-smooth problem using Adult dataset with mini-batch of 32 data points. First row compares accuracy against epochs, gradients/n and time, and second row compares suboptimality against epochs, gradients/n and time.}
	\label{fig_4}
\end{figure}

The results are reported with elastic-net-regularized logistic regression problem (non-smooth regularizer) as given below:
\begin{equation}
\label{eq_elastic_lr}
\underset{w}{\min} \; F(w) = \dfrac{1}{n} \sum_{i=1}^{n} \log\left( 1 + \exp\left( - y_i w^T x_i \right) \right) + \dfrac{\lambda_1}{2} \|w\|^2 + \lambda_2 \|w\|_1,
\end{equation}
where $ f(w) = \dfrac{1}{n} \sum_{i=1}^{n} \log\left( 1 + \exp\left( - y_i w^T x_i \right) \right) + \dfrac{\lambda_1}{2} \|w\|^2$ and $g(w) = \lambda_2\|w\|_1$.\\
Figure \ref{fig_3} represents the comparative study of SAAG-III, IV, SVRG and VR-SGD on rcv1 dataset. As it is clear from the figure, for all the six criteria plots, SAAG-III and IV outperform SVRG and VR-SGD, and provide better accuracy and faster convergence. SAAG-IV gives best results in terms of suboptimality but in terms of accuracy, SAAG-III and IV have close performance except for accuracy versus gradients$/n$, where SAAG-III gives better results because SAAG-III calculates gradients at last iterate only unlike SAAG-IV which calculates gradients at snap point and last iterate. Figure \ref{fig_4} reports results with Adult dataset, and as it is clear from plots, SAAG-III outperforms all the methods in all the six criteria plots. Moreover, SAAG-IV lags behind because the dataset and/or mini-batch size is small. Some results are also given with SVM in Appendix~\ref{appsub_svm_chapter_SAAGs}

% Comparison of SAAG-III and IV with SVRG and VR-SGD

% Comparison of SAAG-III and IV with SVRG and VR-SGD

\subsection{Effect of Regularization Constant}
\label{subsec_effect_of_regularization}
% Comparison of effect of regularization coefficient
\begin{figure}[htb]
	\subfloat{\includegraphics[width=.332\linewidth]{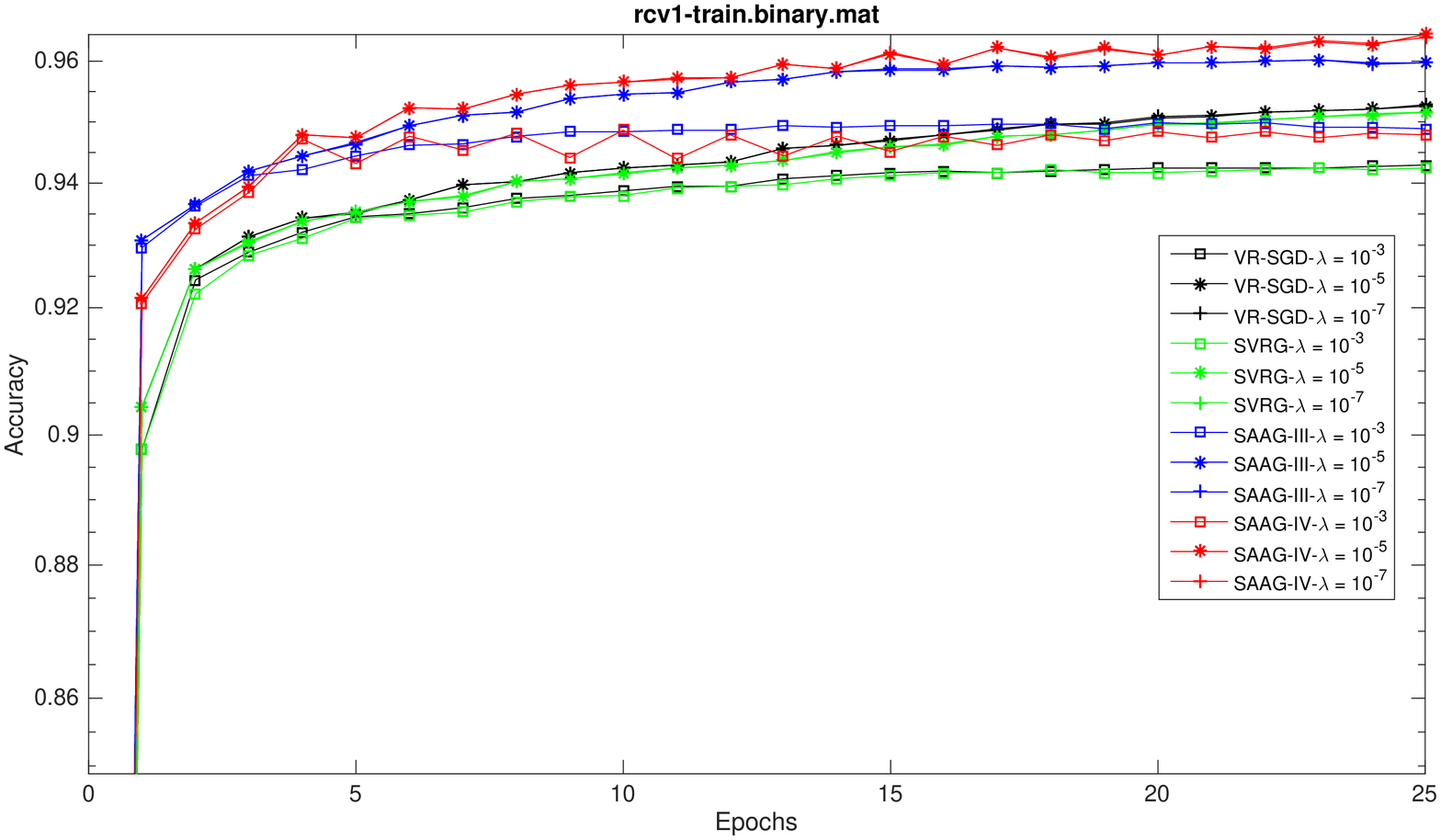}}
	\subfloat{\includegraphics[width=.332\linewidth]{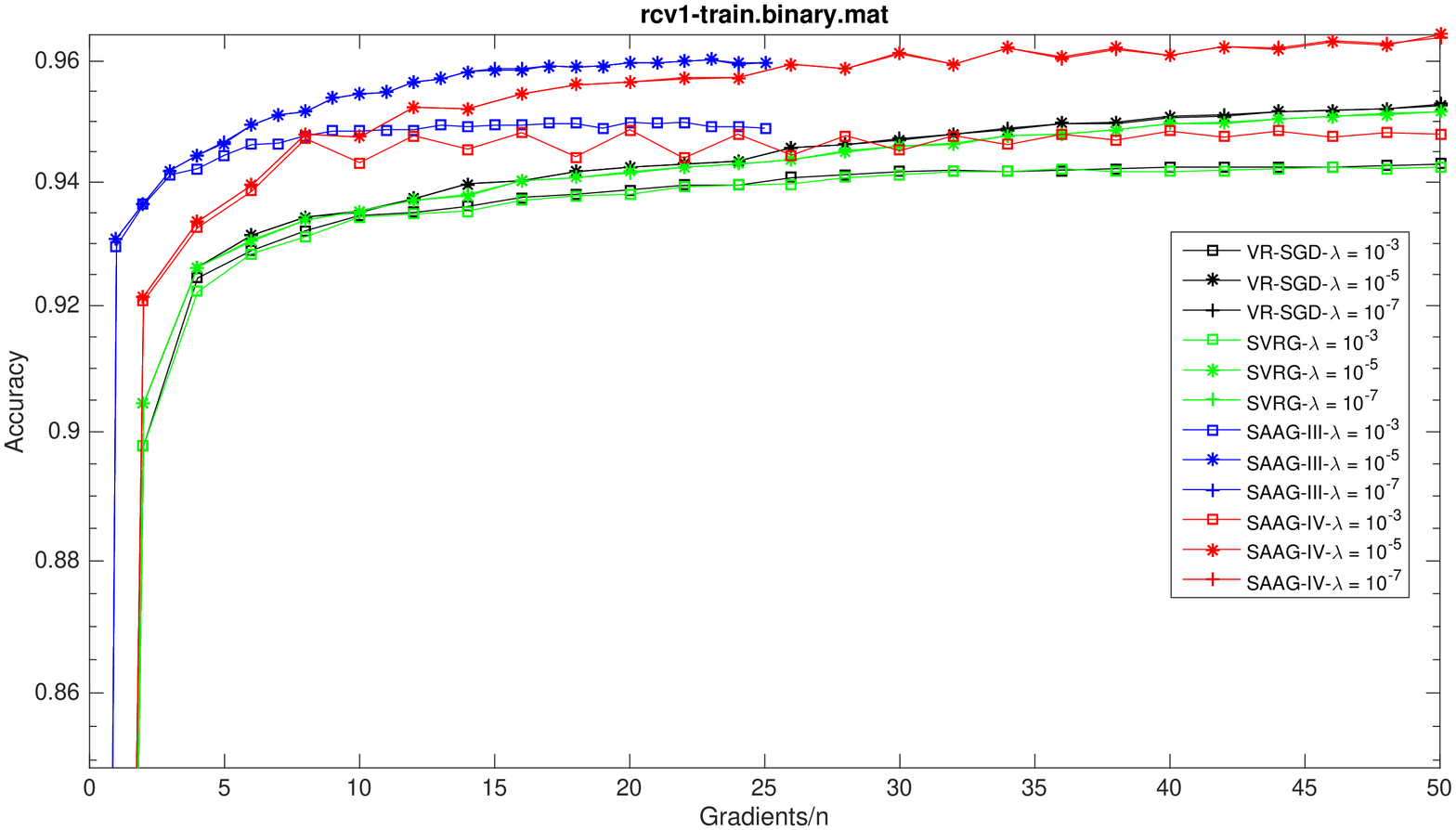}}
	\subfloat{\includegraphics[width=.332\linewidth]{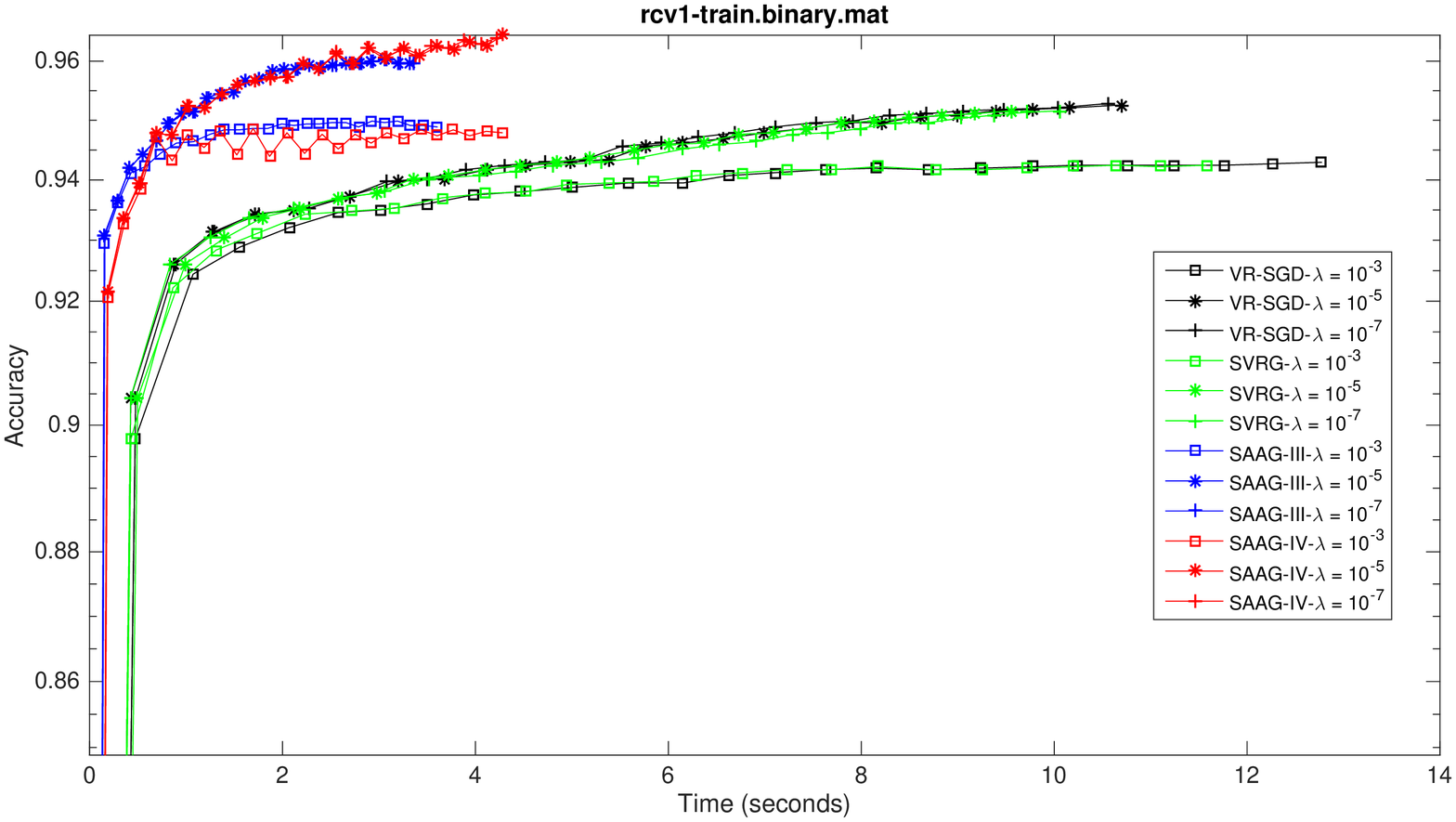}}
	
	\subfloat{\includegraphics[width=.332\linewidth]{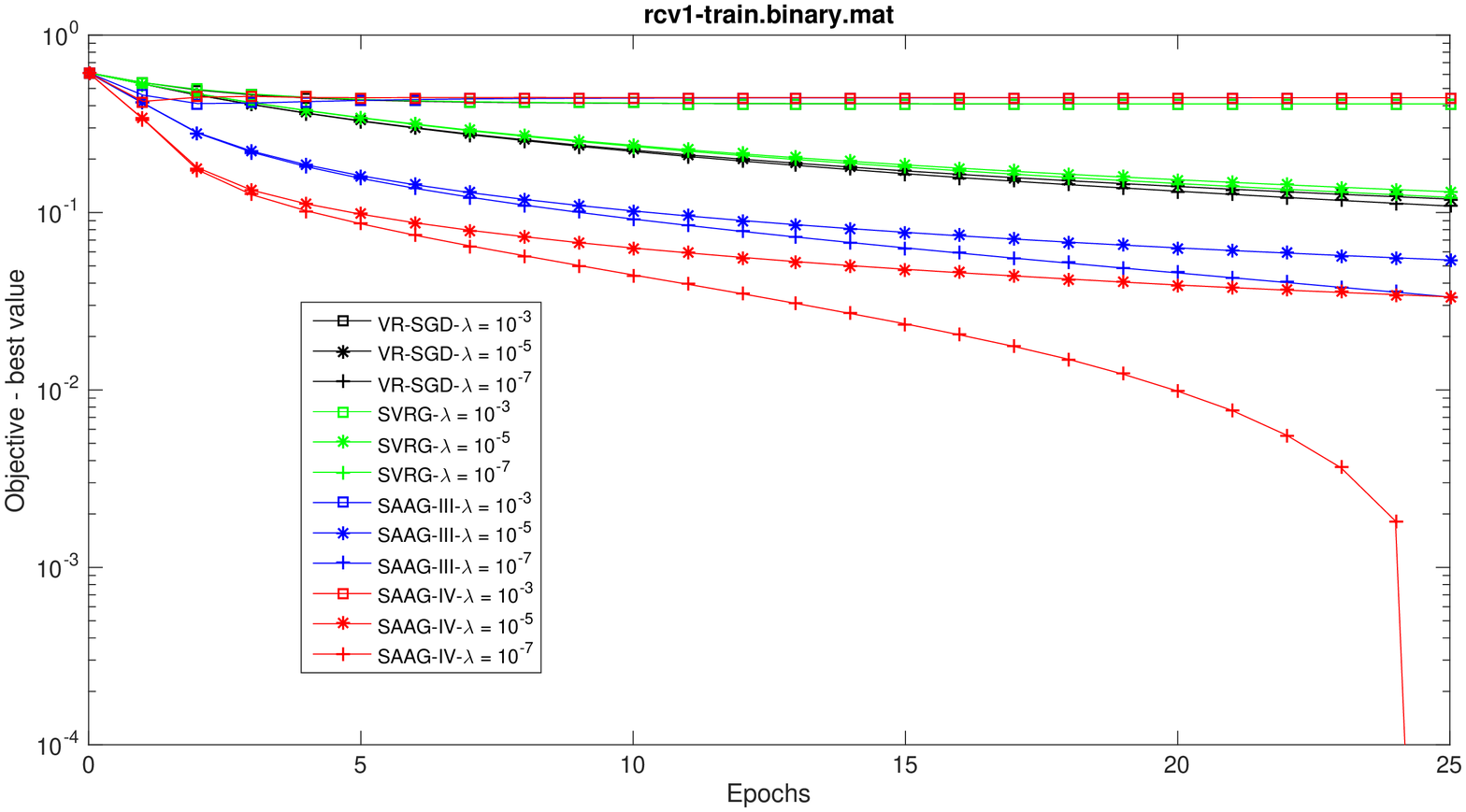}}
	\subfloat{\includegraphics[width=.332\linewidth]{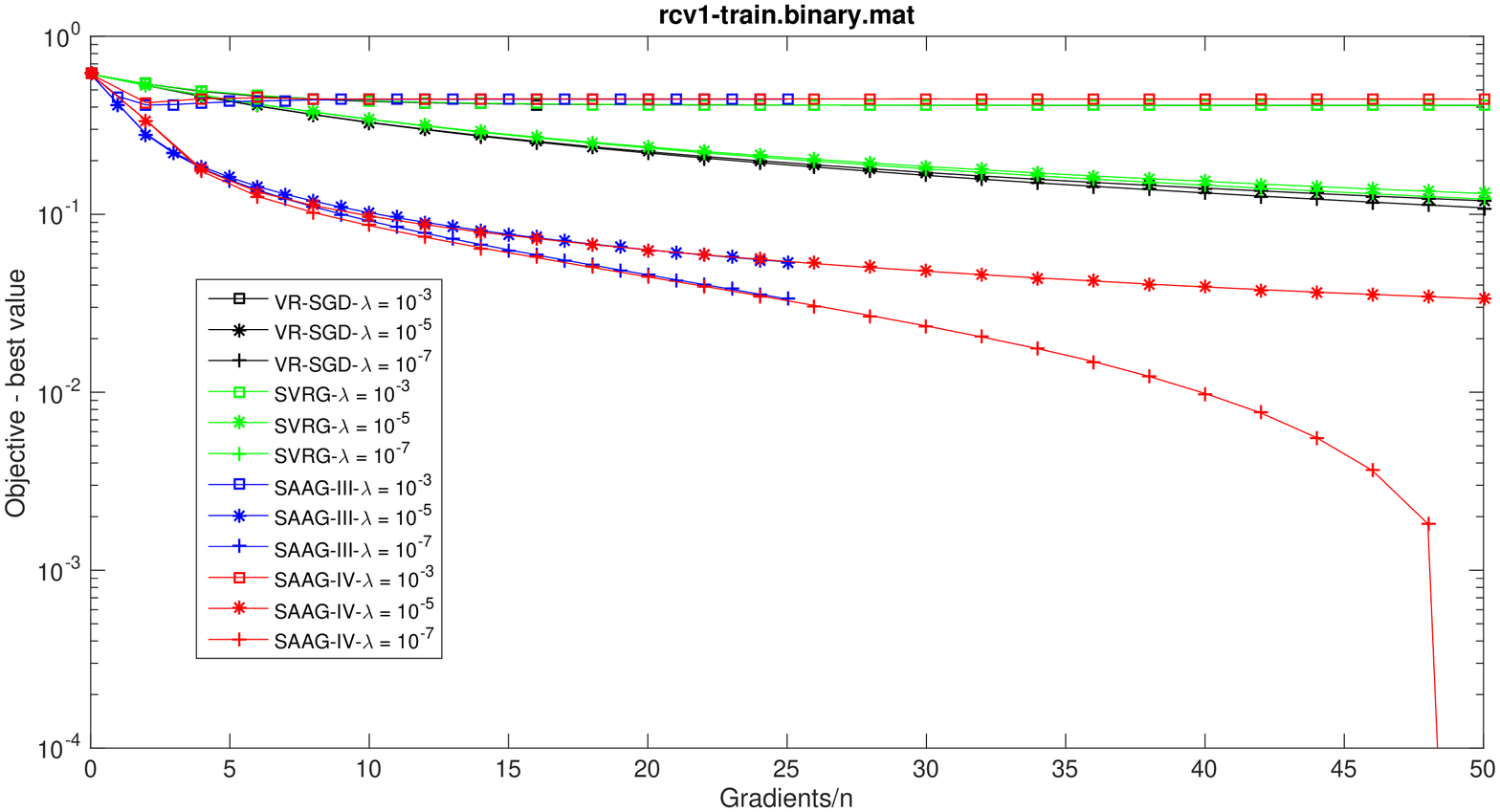}}
	\subfloat{\includegraphics[width=.332\linewidth]{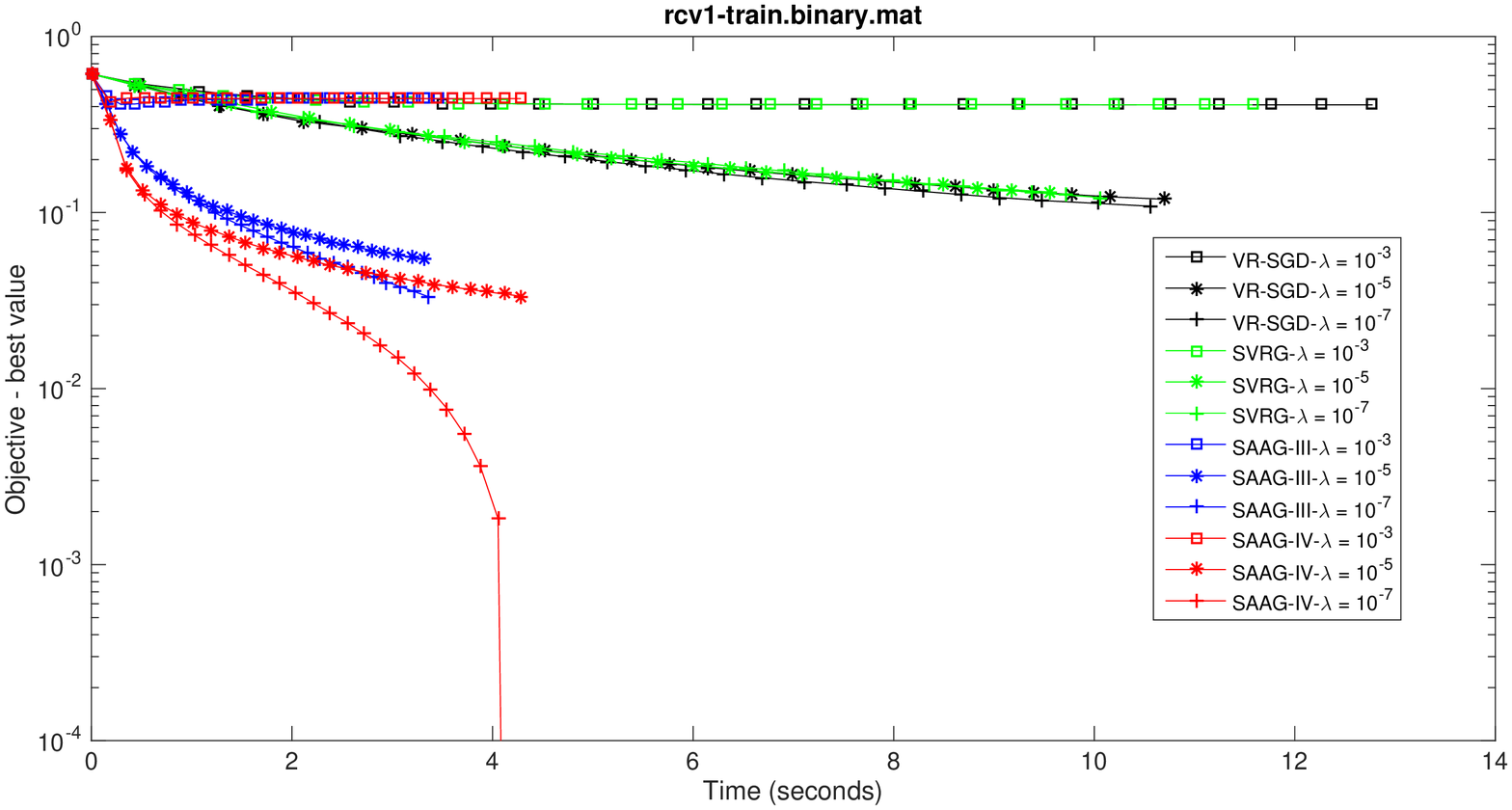}}
	
	\caption{Study of effect of regularization coefficient on SAAG-III, IV, SVRG and VR-SGD for smooth problem using rcv1 dataset and considering regularization coefficient values in \{$10^{-3}$, $10^{-5}$, $10^{-7}$\}. First row compares accuracy against epochs, gradients/n and time, and second row compares suboptimality against epochs, gradients/n and time.}
	\label{fig_regularization_smooth}
\end{figure}
Figure \ref{fig_regularization_smooth}, studies the effect of regularization coefficient on SAAG-III, IV, SVRG and VR-SGD for smooth problem ($l_2$-regularized logistic regression) using rcv1 dataset and considering regularization coefficient values in \{$10^{-3}$, $10^{-5}$, $10^{-7}$\}. As it is clear from the plots, all the methods are affected by the large ($10^{-3}$) regularization coefficient value and have low accuracy but for sufficiently small values methods don't have much effect. For suboptimality plots, all have slower convergence for large regularization coefficient but then convergence improves with the decrease in regularization, because decreasing the regularization, increases the over-fitting. The results for non-smooth problem are similar, so they are given in the Appendix \ref{appsub_regularization_nonsmooth}.

\subsection{Effect of mini-batch size}
% Comparison of effect of mini-batch size.
\begin{figure}[htb]
	\subfloat{\includegraphics[width=.332\linewidth]{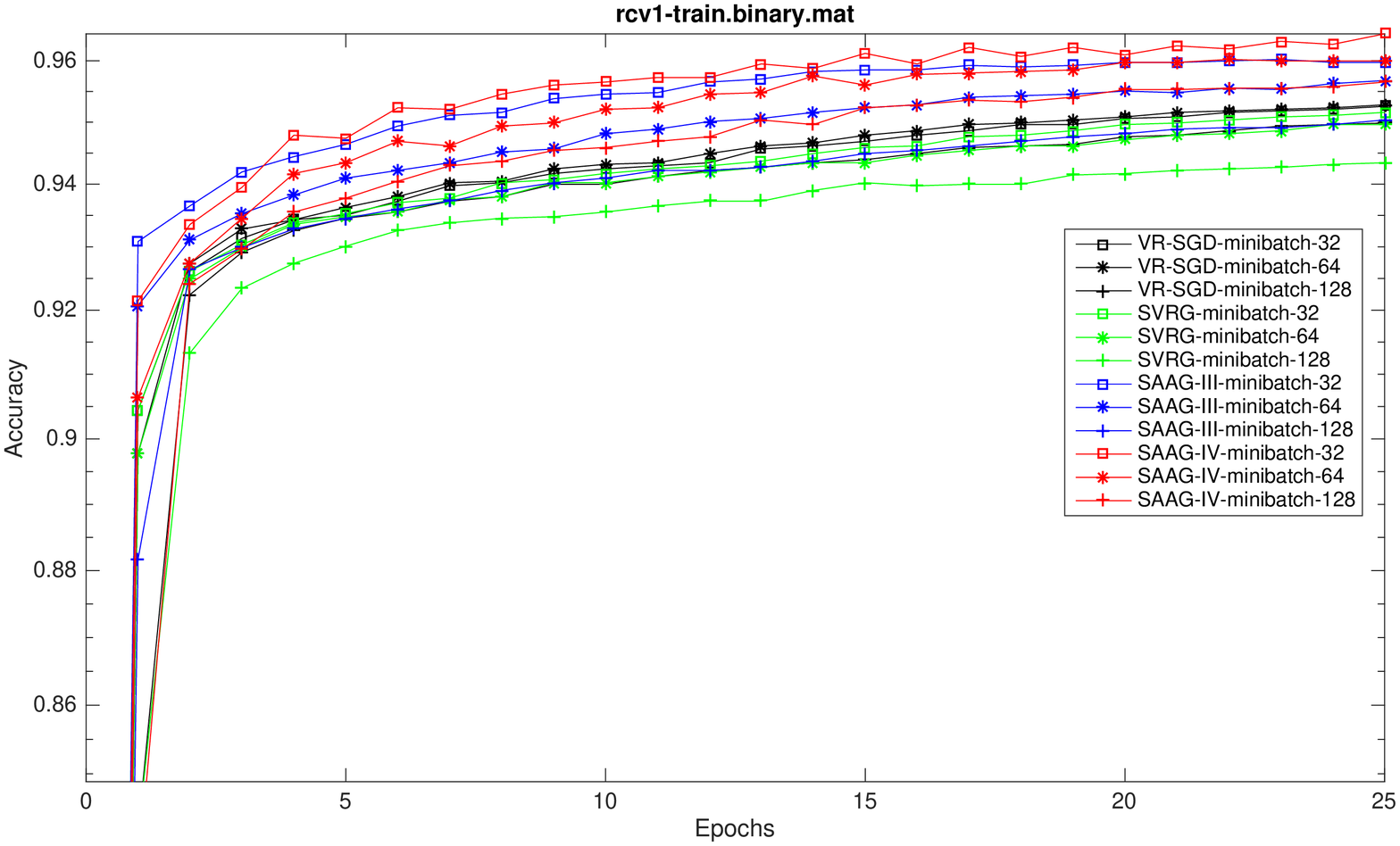}}
	\subfloat{\includegraphics[width=.332\linewidth]{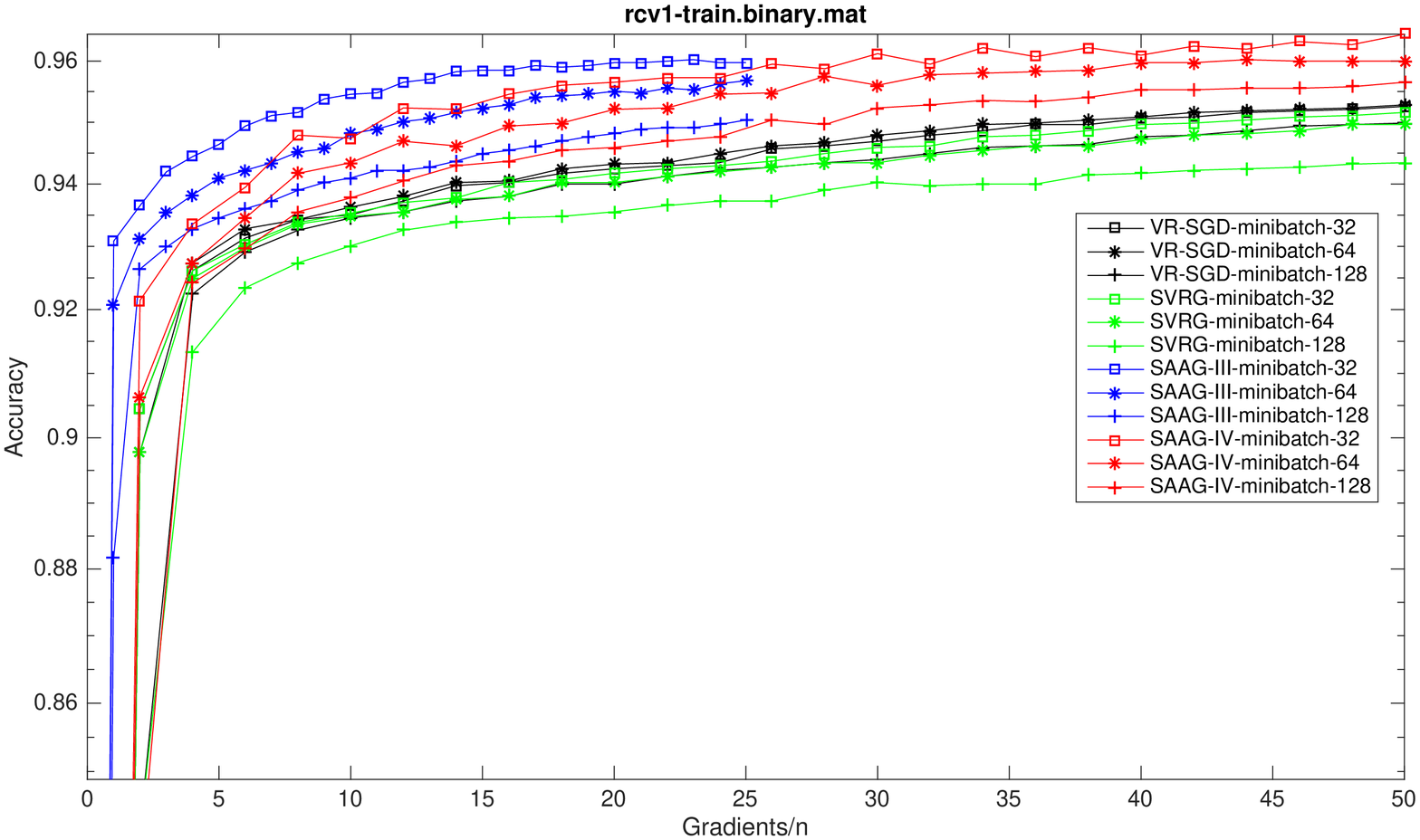}}
	\subfloat{\includegraphics[width=.332\linewidth]{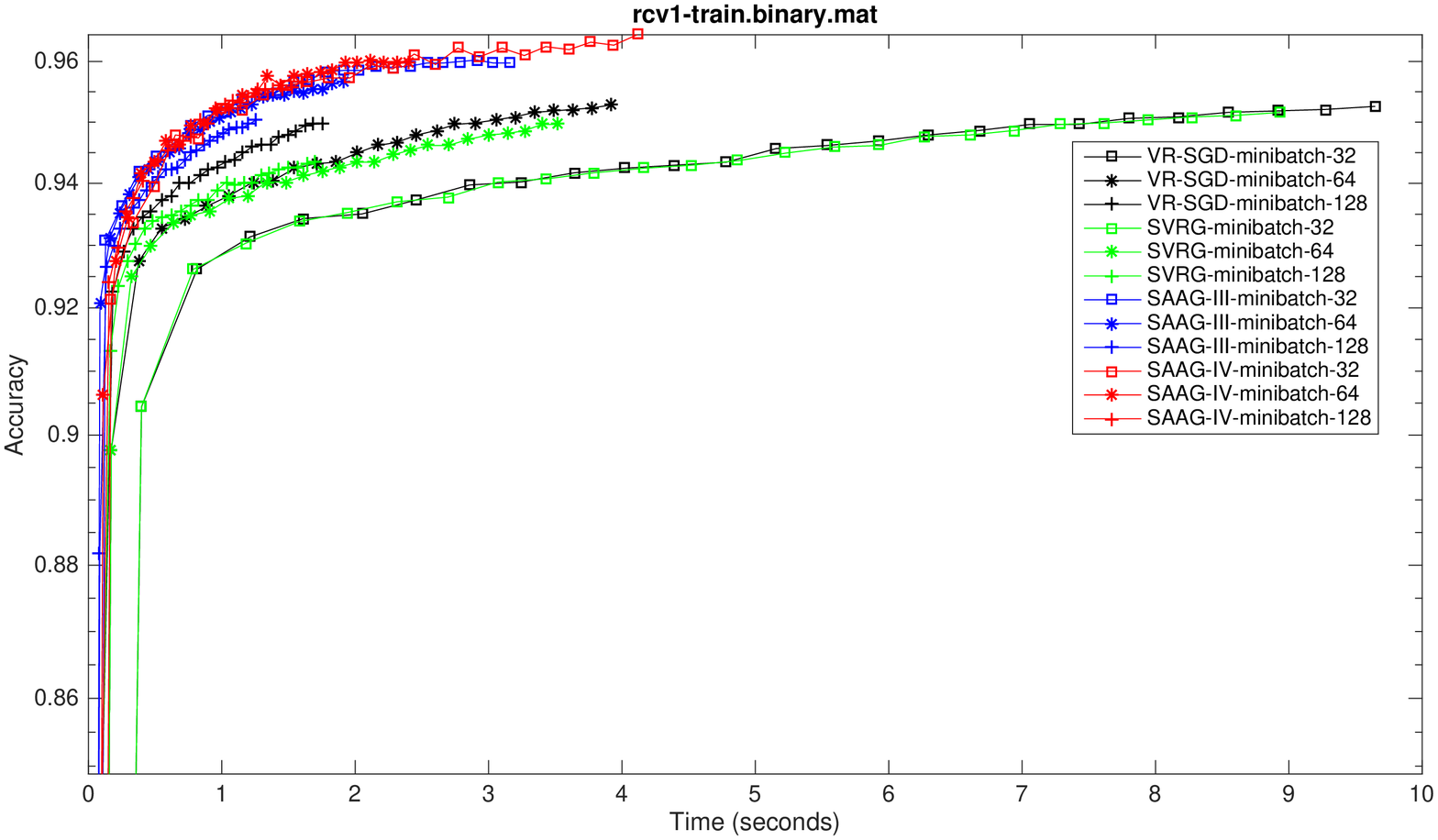}}
	
	\subfloat{\includegraphics[width=.332\linewidth]{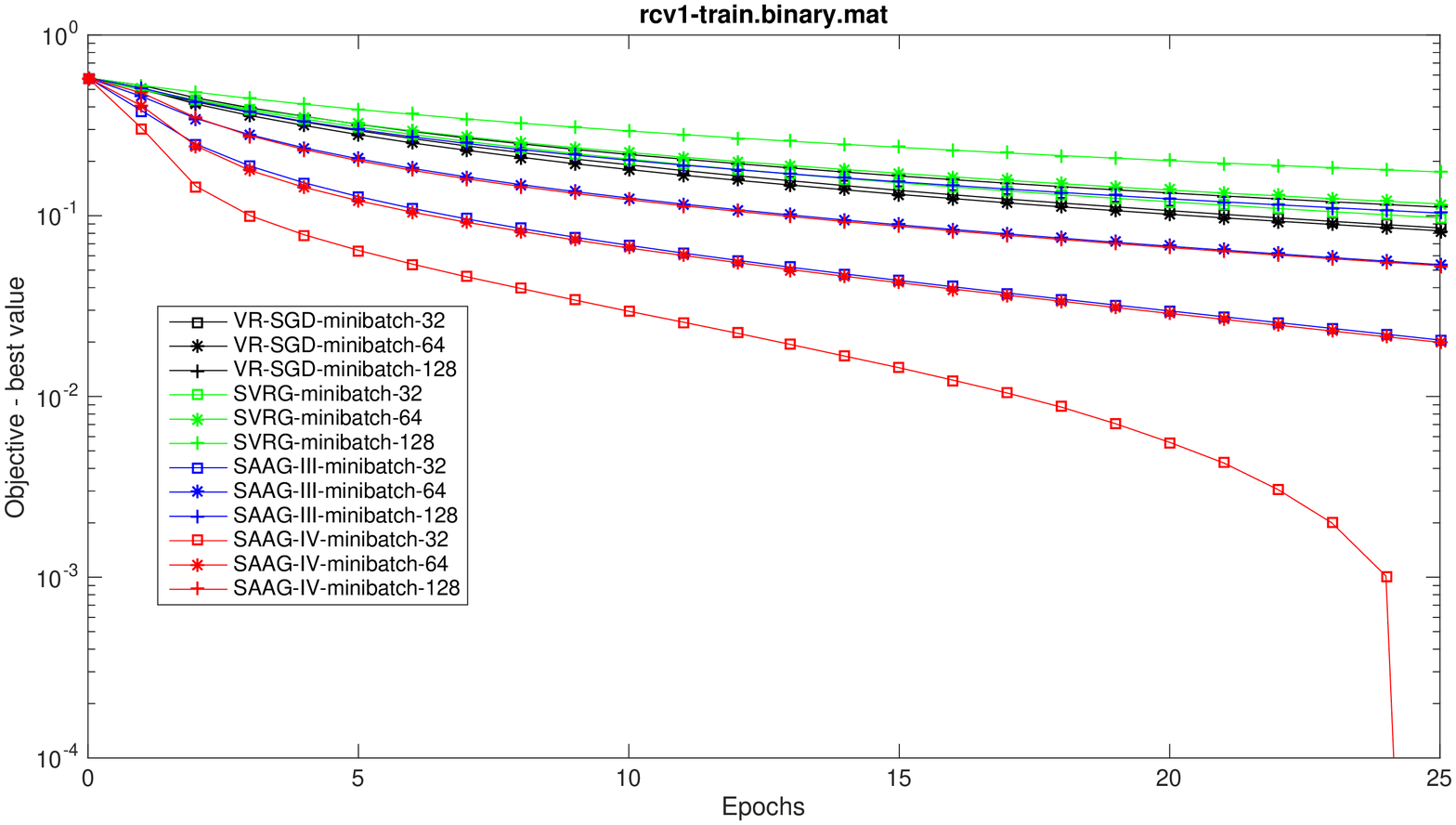}}
	\subfloat{\includegraphics[width=.332\linewidth]{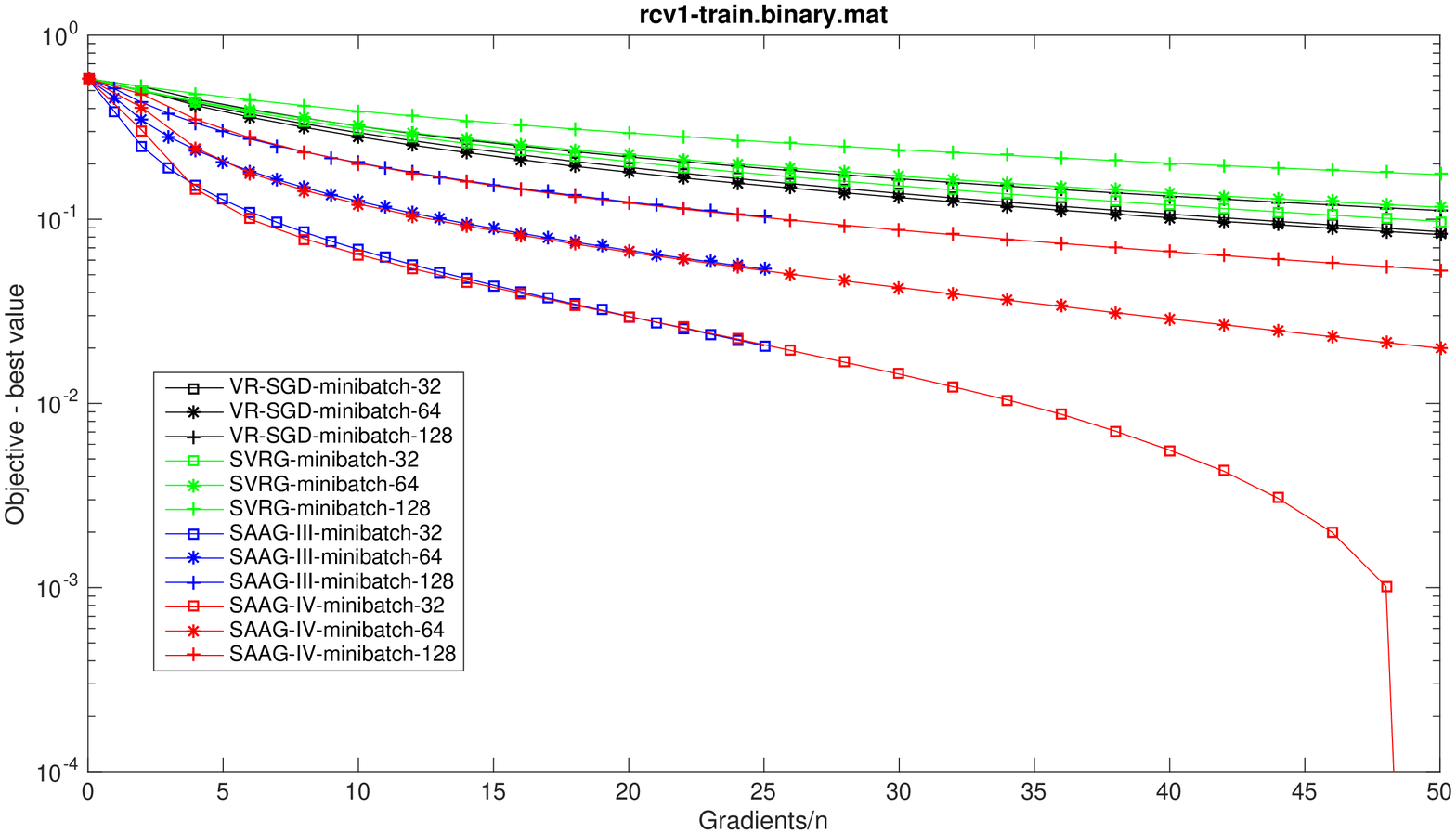}}
	\subfloat{\includegraphics[width=.332\linewidth]{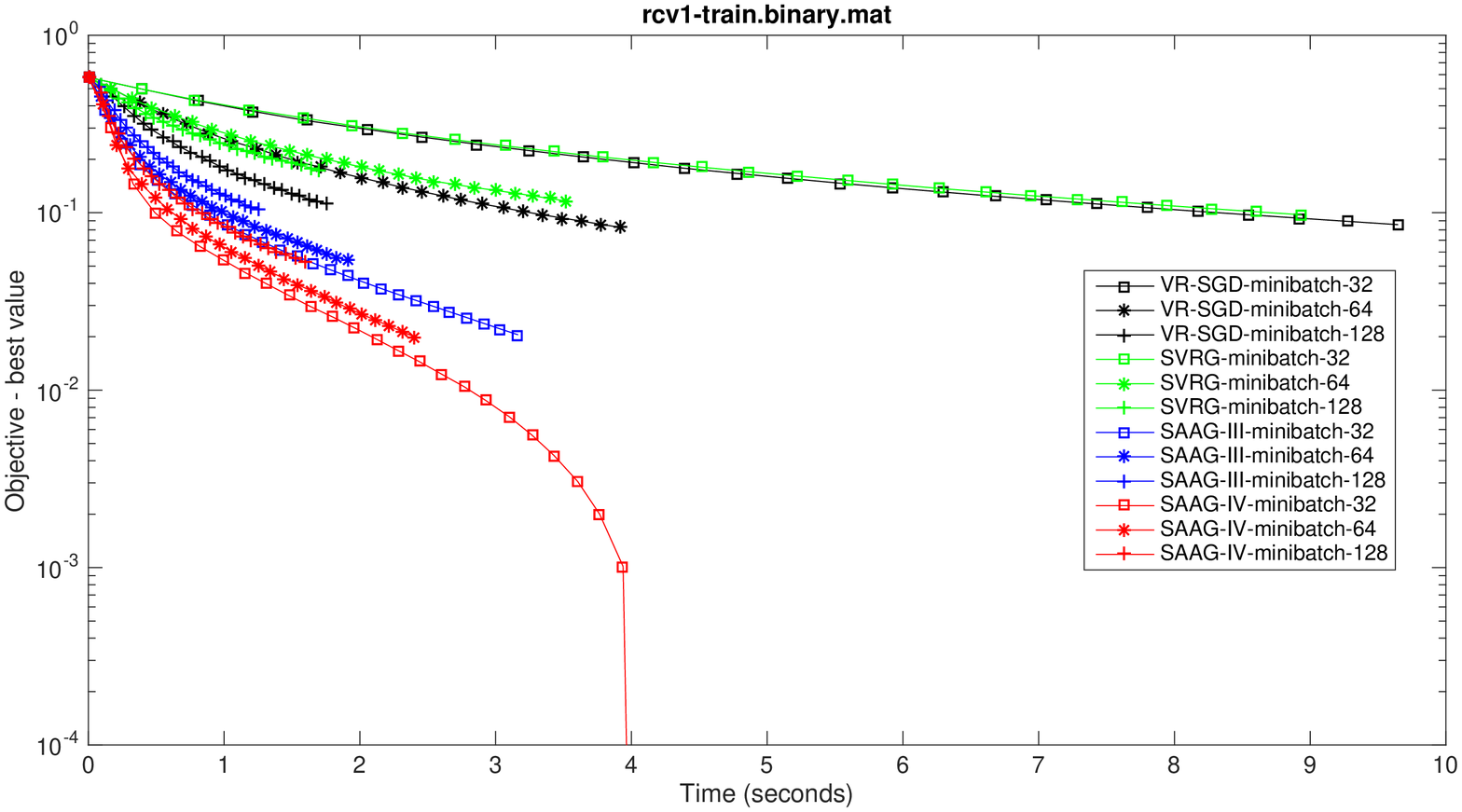}}
	
	\caption{Study of effect of mini-batch size on SAAG-III, IV, SVRG and VR-SGD for smooth problem, using rcv1 dataset with mini-batch sizes of 32, 64 and 128. First row compares accuracy against epochs, gradients/n and time, and second row compares suboptimality against epochs, gradients/n and time.}
	\label{fig_minibatchsize_smooth}
\end{figure}
Figure \ref{fig_minibatchsize_smooth}, studies the effect of mini-batch size on SAAG-III, IV, SVRG and VR-SGD for smooth problem ($l_2$-regularized logistic regression) using rcv1 dataset and considers mini-batch values in \{$32, 64, 128$\} data points. As it is clear from the plots, except for the results with training time, the performance of SAAG-III, IV and SVRG fall with increase in mini-batch size but for VR-SGD performance first improves slightly and then falls slightly. For results with training time the performance of SAAG-III and IV falls with mini-batch size for suboptimality and remains almost same for accuracy. But the performance of VR-SGD and SVRG improves because VR-SGD and SVRG train quickly for large mini-batches.  Similar results are obtained for study of effect of mini-batch size on non-smooth problem so the results are given in the Appendix~\ref{appsub_minibatch_effect_nonsmooth}.

\section{Conclusion}
\label{sec_conclusions}
We have proposed novel variants of SAAG-I and II, called SAAG-III and IV, respectively, by using average of iterates for SAAG-III as a starting point, and average of iterates and last iterate for SAAG-IV as the snap point and starting point, respectively, for new epoch, except the first one. SAAGs (I, II, III and IV), are also extended to solve non-smooth problems by using two different update rules and introducing proximal step for non-smooth problem. Theoretical results proved linear convergence of SAAG-IV for all the four combinations of smoothness and strong-convexity with some initial errors, in expectation. The empirical results proved the efficacy of proposed methods against existing variance reduction methods in terms of, accuracy and suboptimality, against training time, epochs and gradients$/n$.

\begin{acknowledgements}
	First author is thankful to Ministry of Human Resource Development, Government of INDIA, to provide fellowship (University Grants Commission - Senior Research Fellowship) to pursue his PhD.
\end{acknowledgements}

\clearpage
\appendix
\section{More Experiments}
\label{app_more_exps}

\subsection{Results with Support Vector Machine (SVM)}
\label{appsub_svm_chapter_SAAGs}
\begin{figure}[htb]
	\subfloat{\includegraphics[width=.5\linewidth]{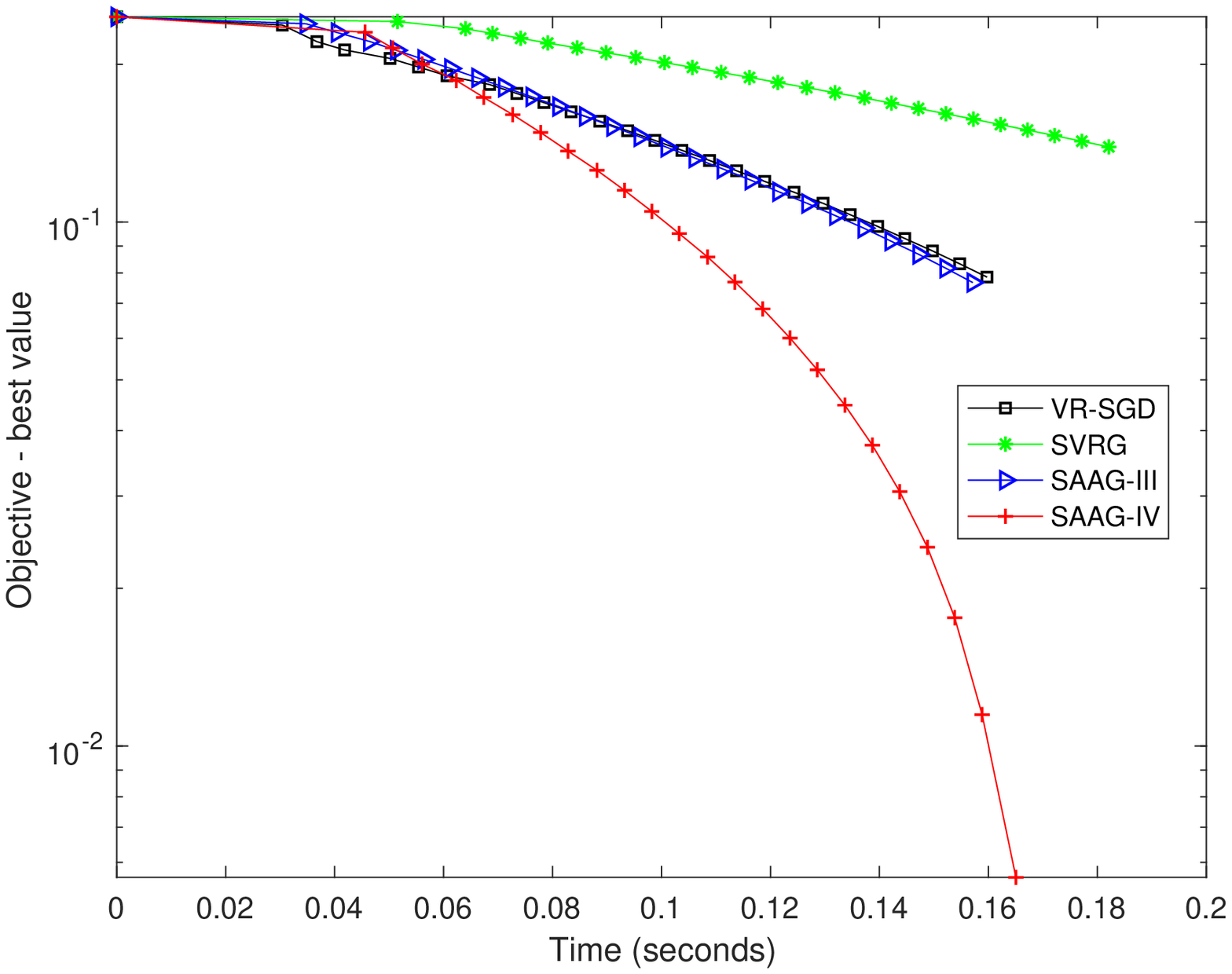}}
	\subfloat{\includegraphics[width=.5\linewidth]{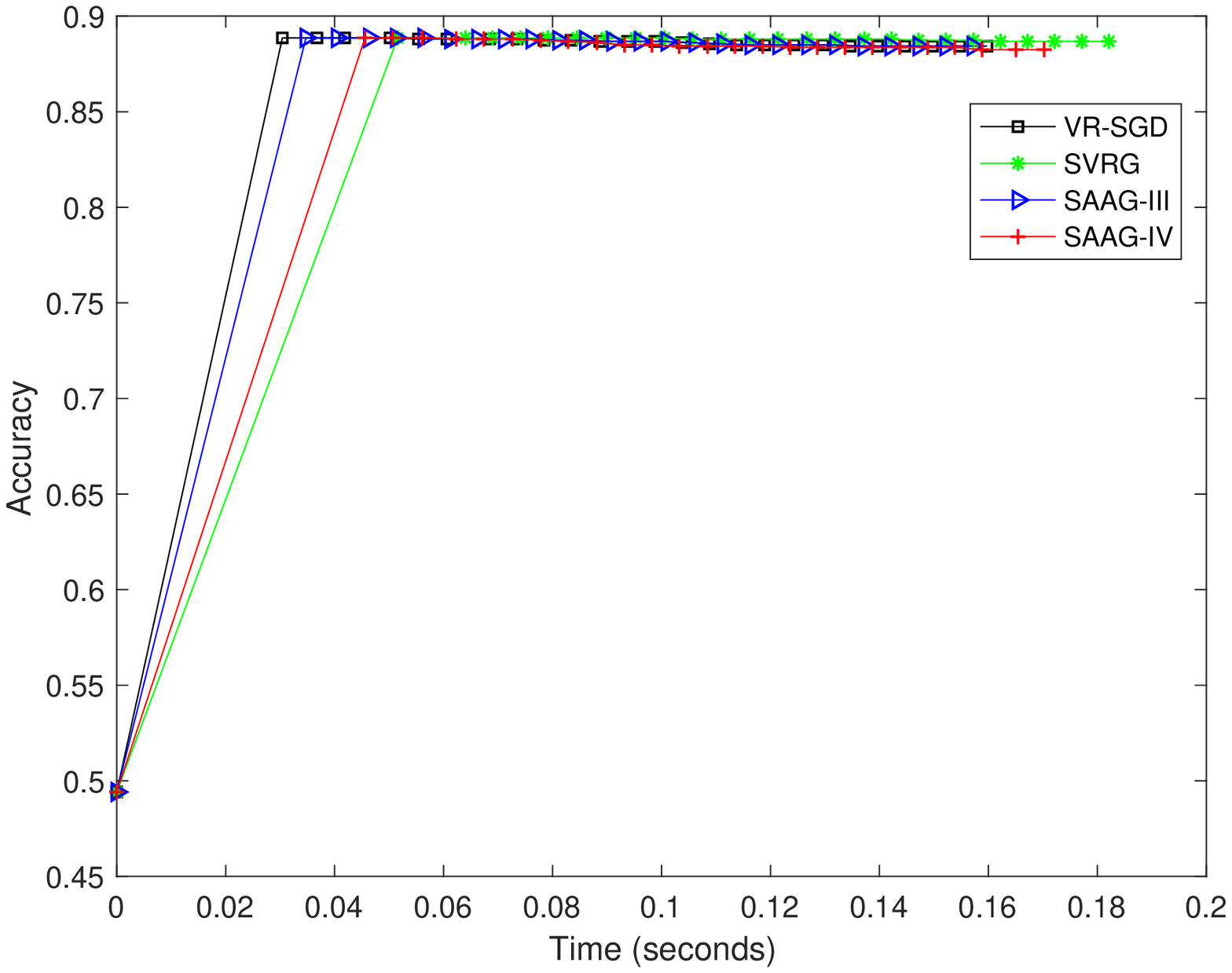}}
	
	\subfloat{\includegraphics[width=.5\linewidth]{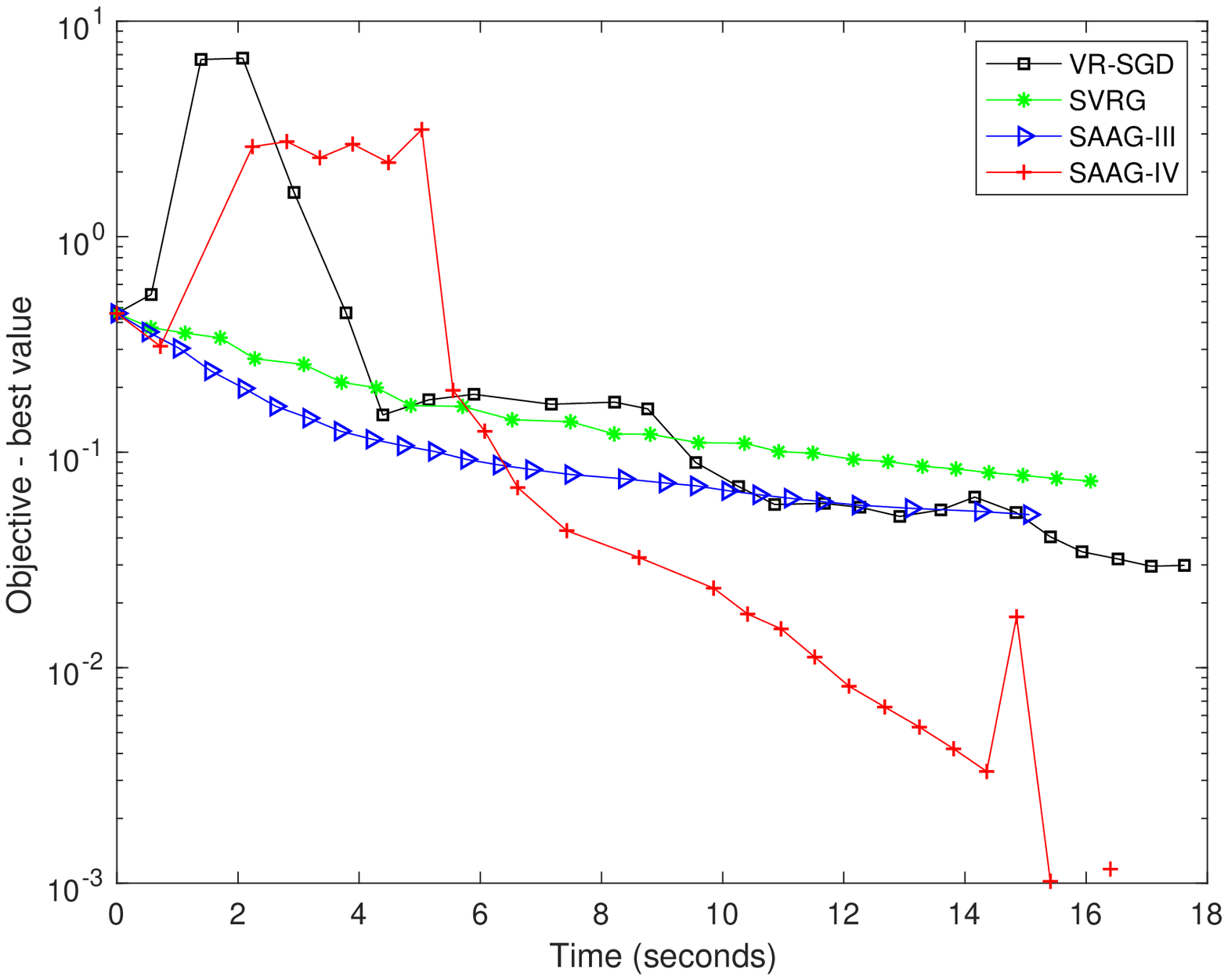}}	
	\subfloat{\includegraphics[width=.5\linewidth]{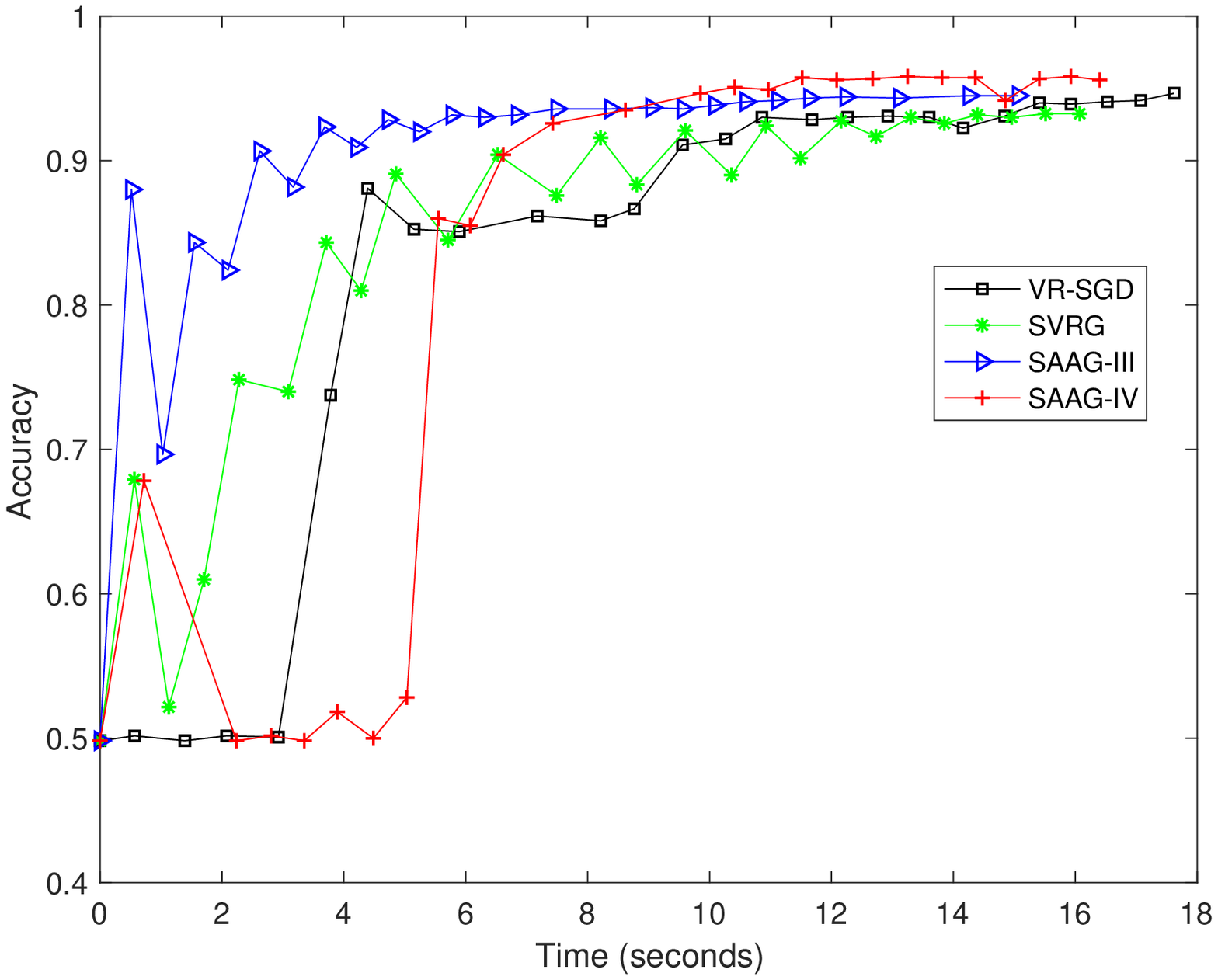}}
	
	\caption{Results with SVM using mini-batch of 1000 data points on mushroom (first row) and gisette (second row) datasets.}
	\label{fig_svm_chapter_SAAGs}
\end{figure}

\noindent \indent This subsection compares SAAGs against SVRG and VR-SGD on SVM problem with mushroom and gisette datasets. Methods use stochastic backtracking line search method to find the step size. Fig.~\ref{fig_svm_chapter_SAAGs} presents the results and compares the suboptimality against the training time (in seconds). Results are similar to experiments with logistic regression but are not that smooth. SAAGs outperform other methods on mushroom dataset (first row) and gisette dataset (second row) for suboptimality against training time and accuracy against time but all methods give almost similar results on accuracy versus training time for mushroom dataset. SAAG-IV outperforms other method and SAAG-III sometimes lags behind VR-SGD method. It is also observed that results with logistic regression are better than the results with the SVM problem. The optimization problem for SVM is given below:
\begin{equation}
\label{eq_l2_svm_chapter_SAAGs}
	\underset{w}{\min} \; F(w) = \dfrac{1}{n} \sum_{i=1}^{n} \max\left(0, 1 - y_i w^T x_i \right)^2 + \dfrac{\lambda}{2} \|w\|^2,
\end{equation}
where $\lambda$ is the regularization coefficient (also penalty parameter) which balances the trade off between margin size and error \cite{Chauhan2018Review}.

\subsection{Comparison of SAAGs (I, II, III and IV) for non-smooth problem}
\label{appsub_saags_comp}
\begin{figure}[htb]
	\subfloat{\includegraphics[width=.332\linewidth]{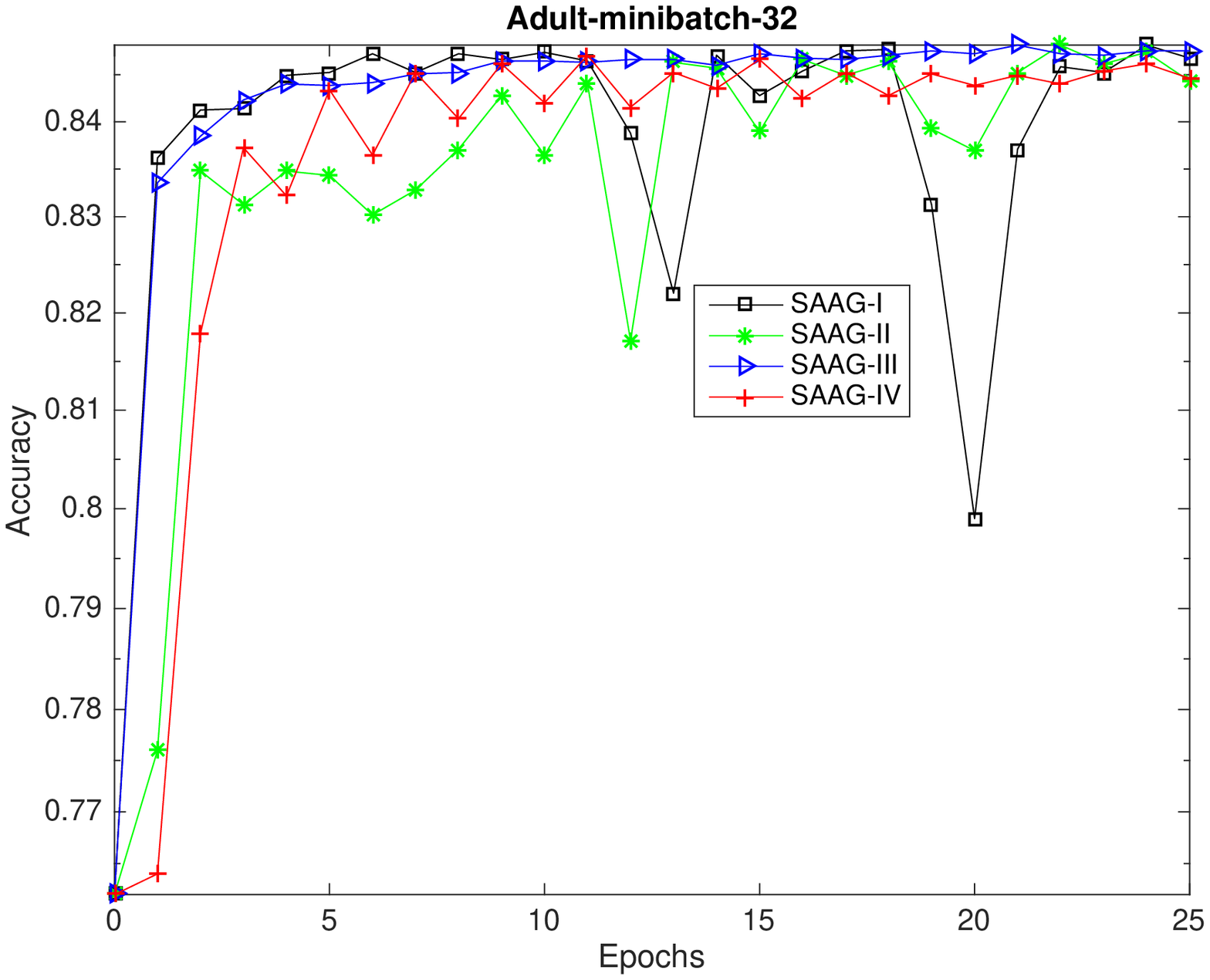}}
	\subfloat{\includegraphics[width=.332\linewidth]{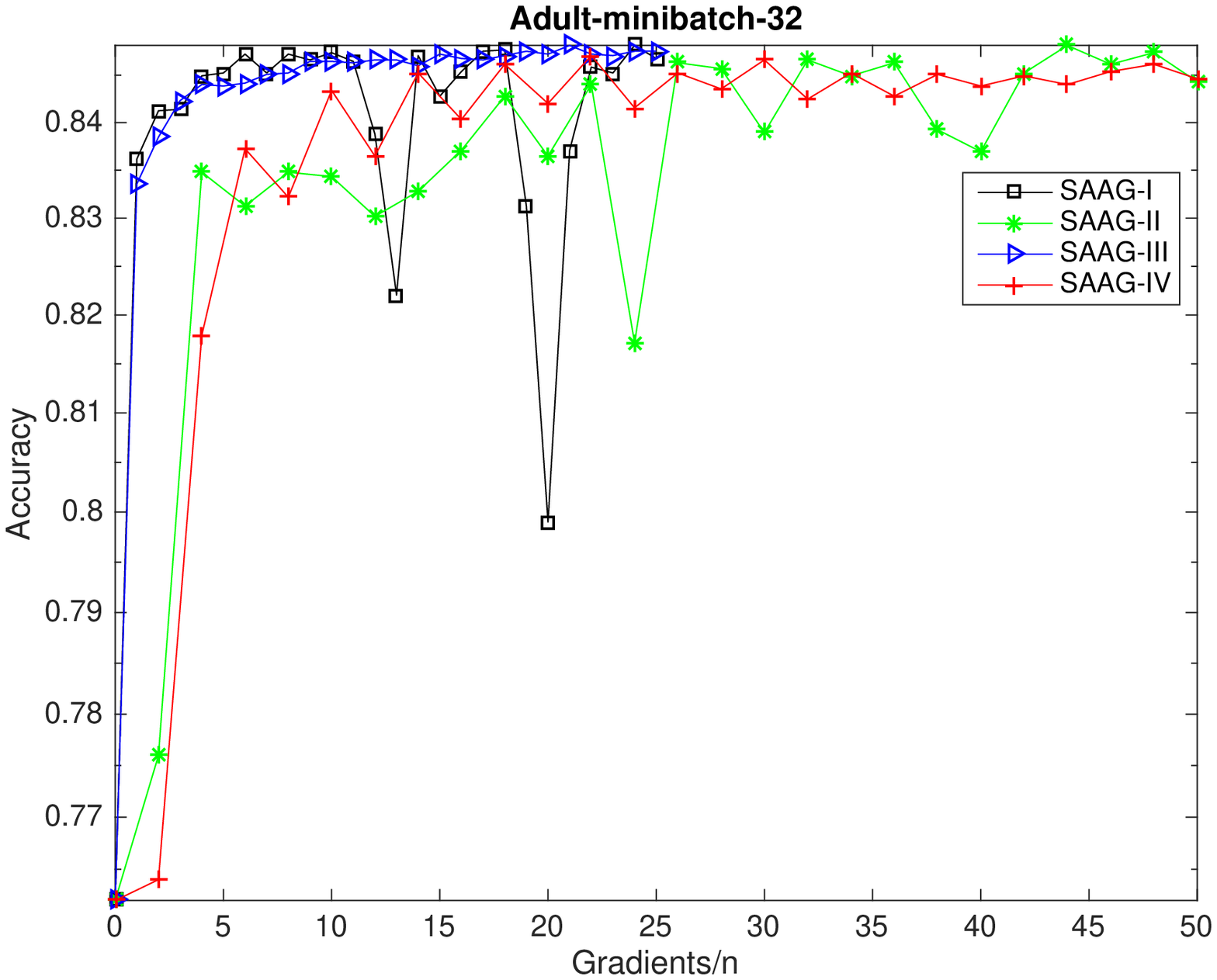}}
	\subfloat{\includegraphics[width=.332\linewidth]{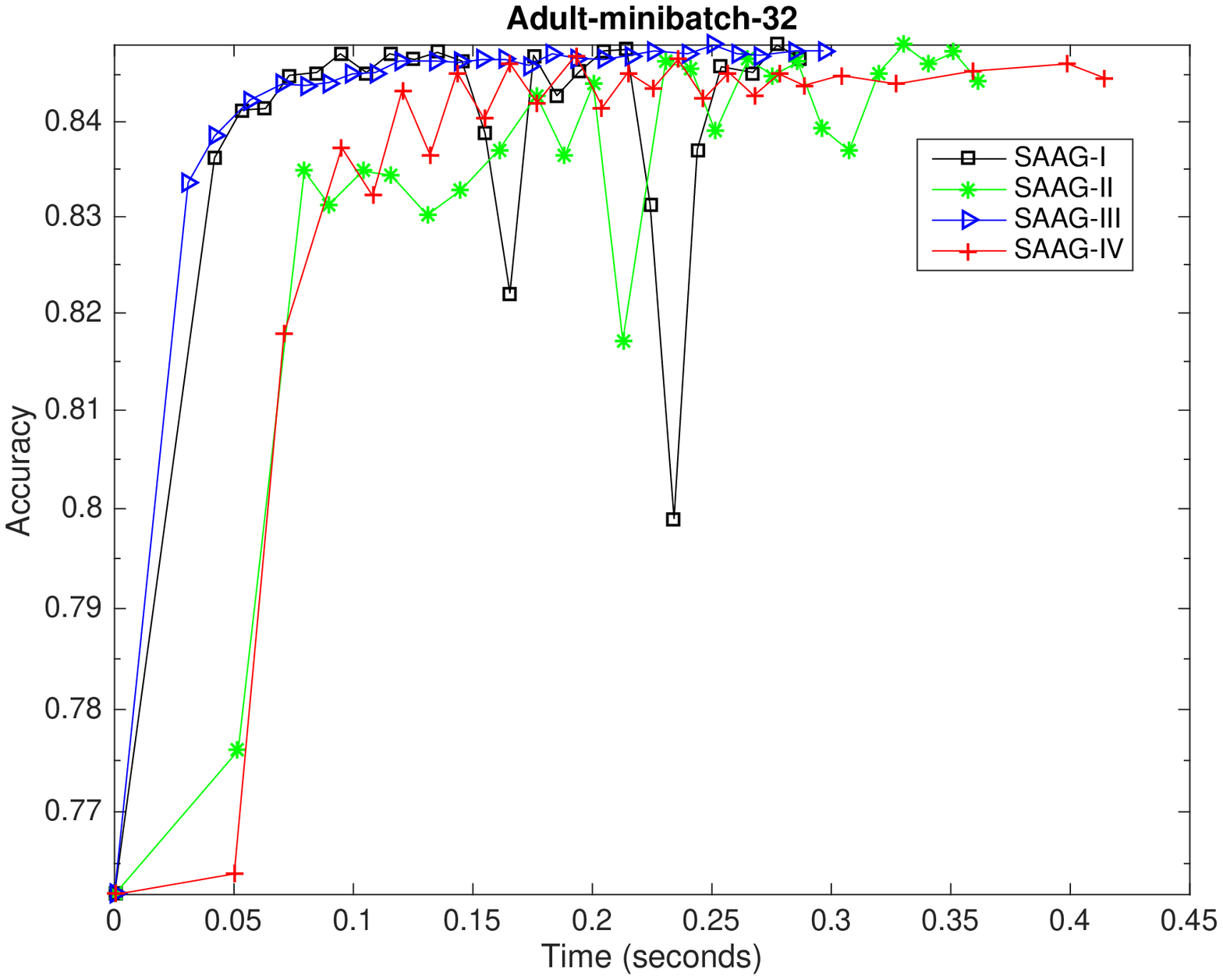}}
	
	\subfloat{\includegraphics[width=.332\linewidth]{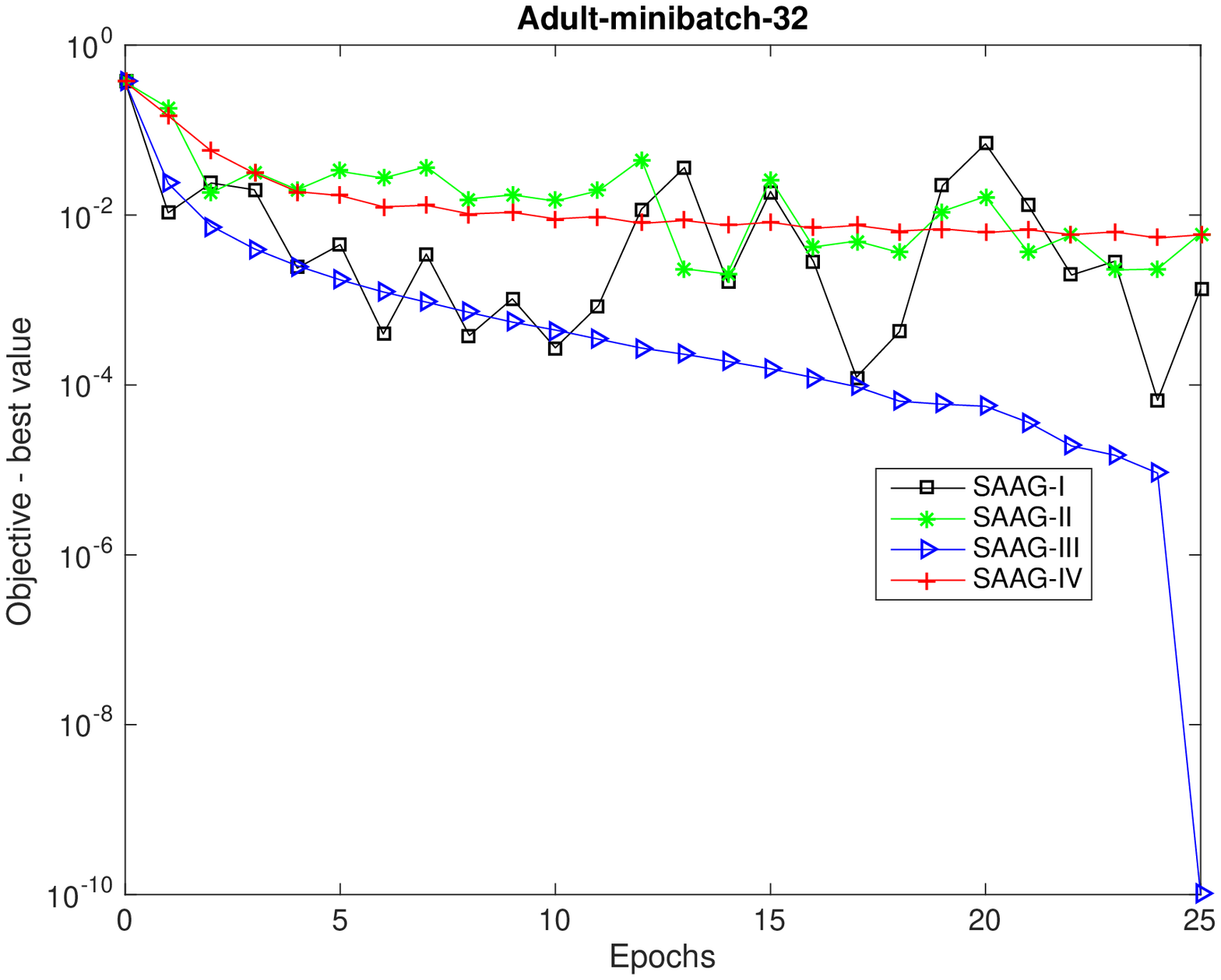}}
	\subfloat{\includegraphics[width=.332\linewidth]{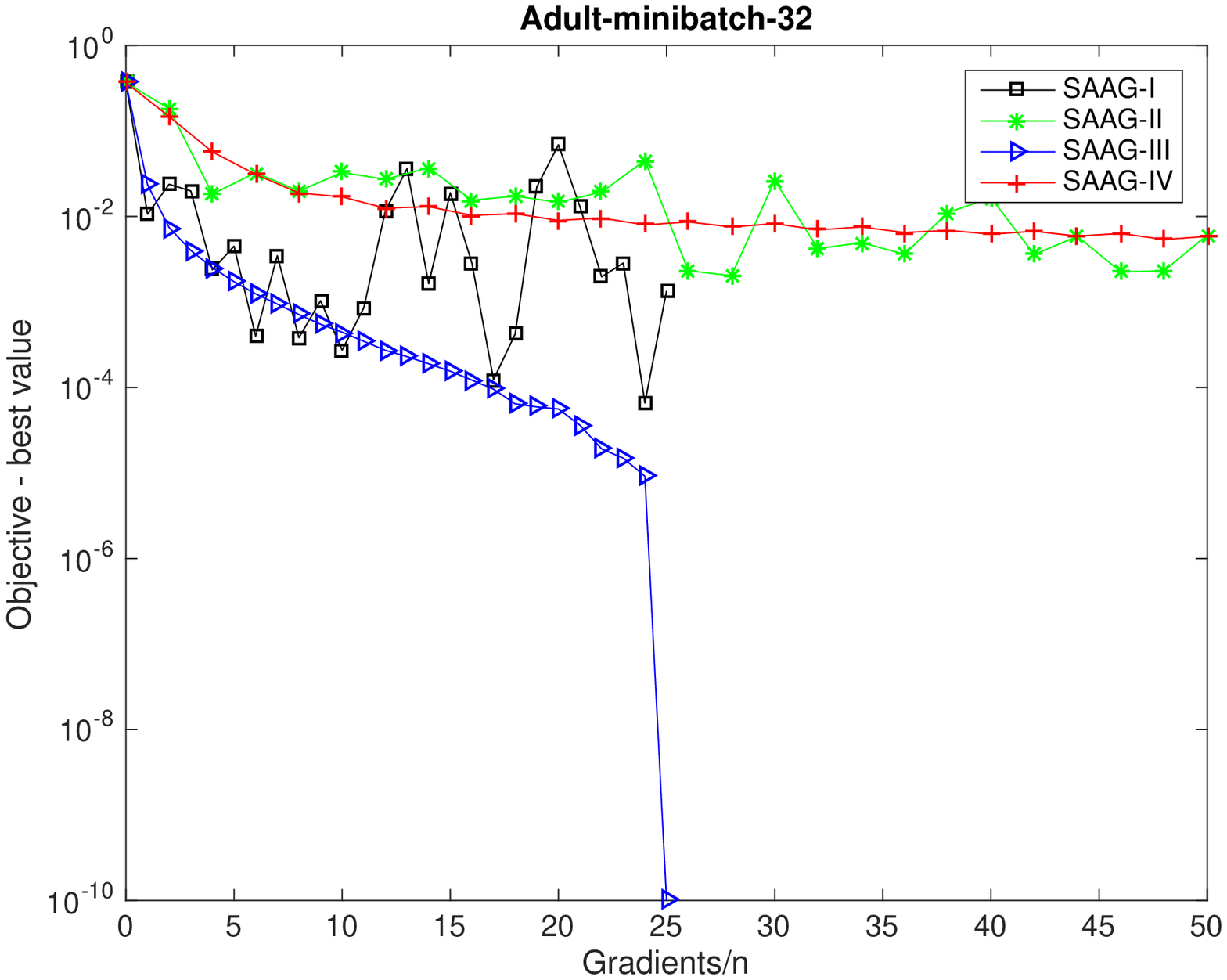}}
	\subfloat{\includegraphics[width=.332\linewidth]{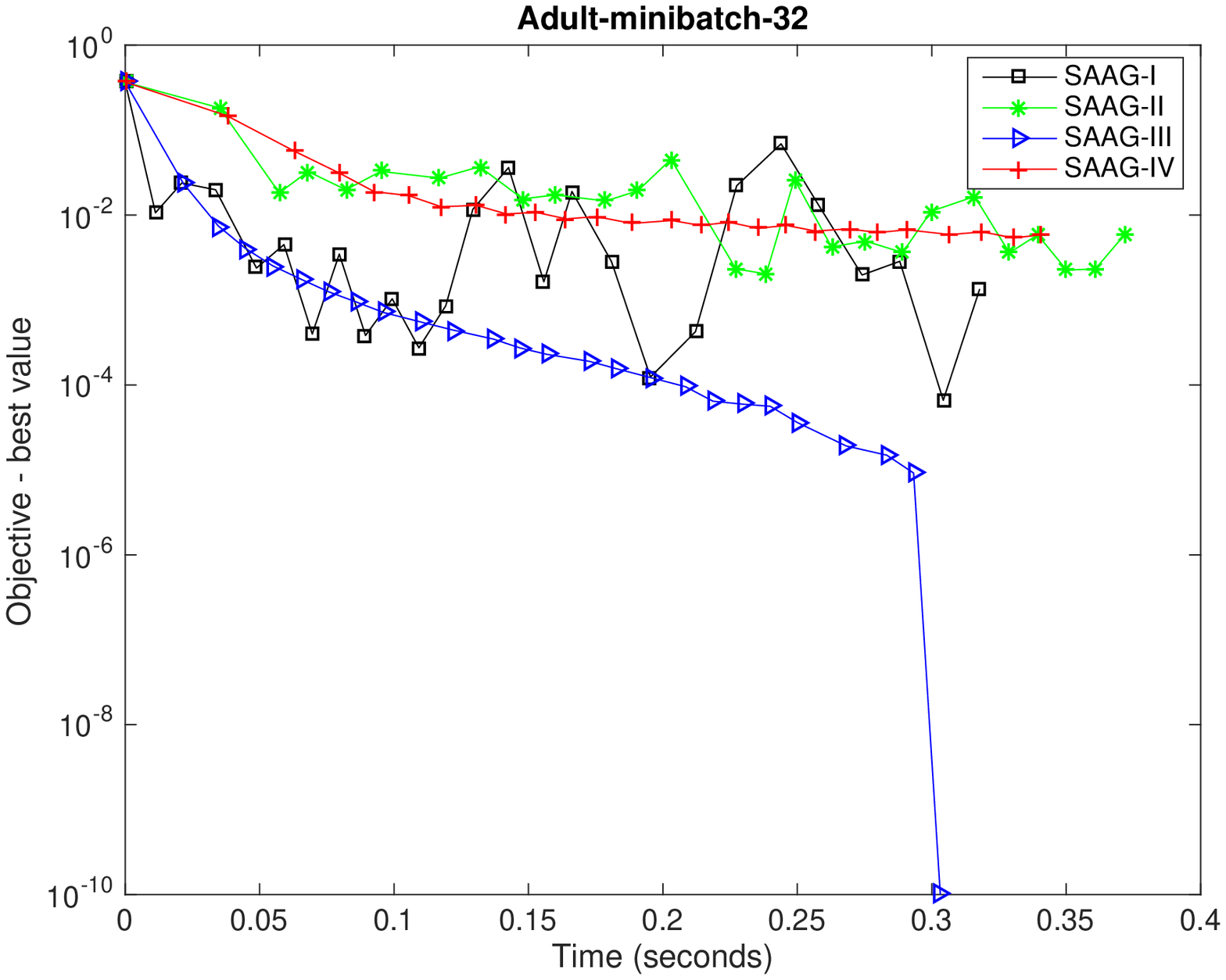}}
	
	\caption{Comparison of SAAG-I, II, III and IV on non-smooth problem (elastic-net-regularized logistic regression) using Adult dataset with mini-batch size of 32 data points. First row compares accuracy against epochs, gradients/n and time, and second row compares suboptimality against epochs, gradients/n and time.}
	\label{fig_saags_nonsmooth}
\end{figure}

Comparison of SAAGs for non-smooth problem is depicted in Figure~\ref{fig_saags_nonsmooth} using Adult dataset with mini-batch of 32 data points. As it is clear from the figure, just like the smooth problem, results with SAAG-III and IV are stable and better or equal to SAAG-I and II.
% Comparison of SAAG-I and SAAG-II with SAAG-III and IV

\subsection{Effect of mini-batch size on SAAG-III, IV, SVRG and VR-SGD for non-smooth problem}
\label{appsub_minibatch_effect_nonsmooth}
\begin{figure}[htb]
	\subfloat{\includegraphics[width=.332\linewidth]{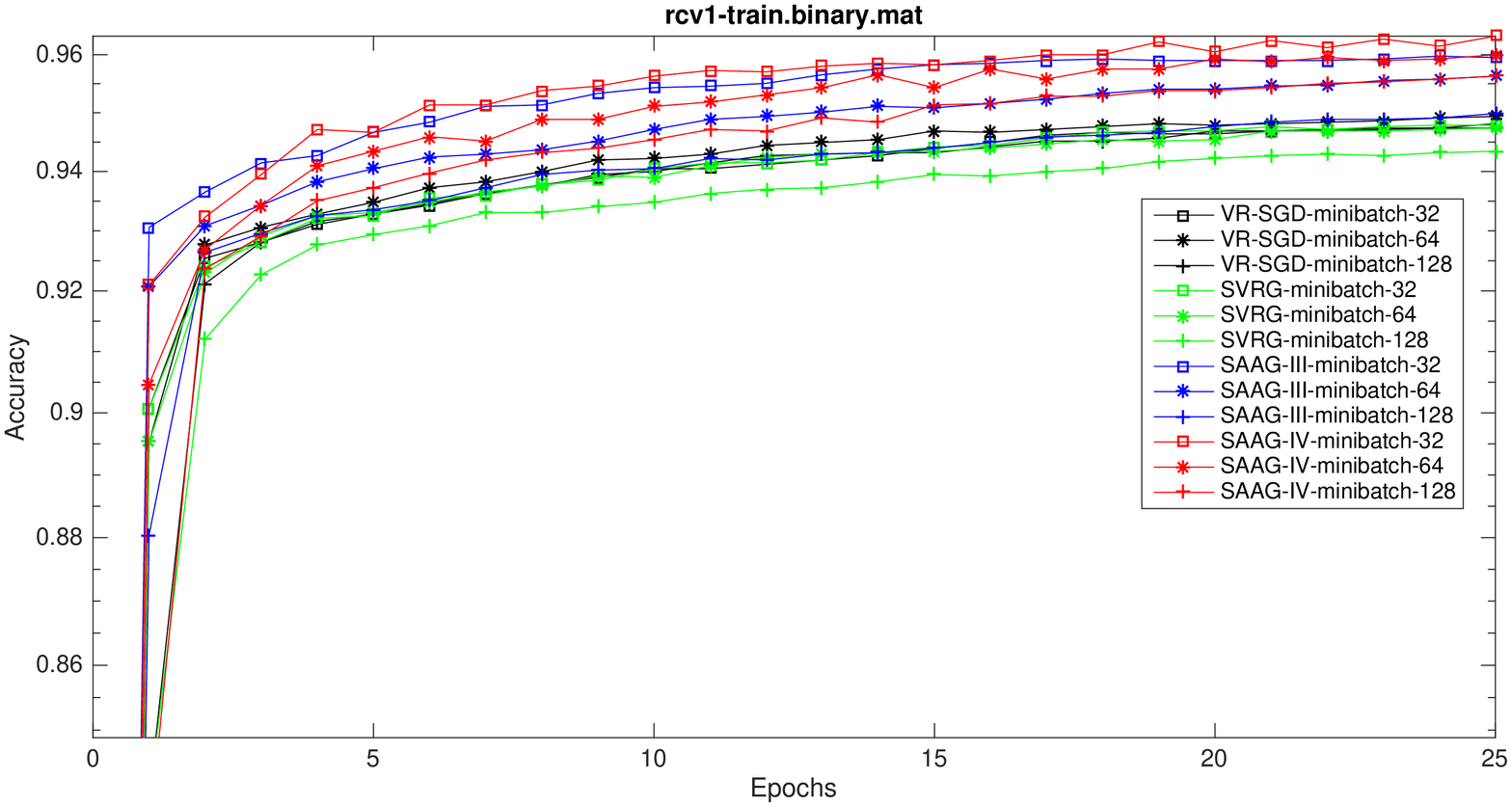}}
	\subfloat{\includegraphics[width=.332\linewidth]{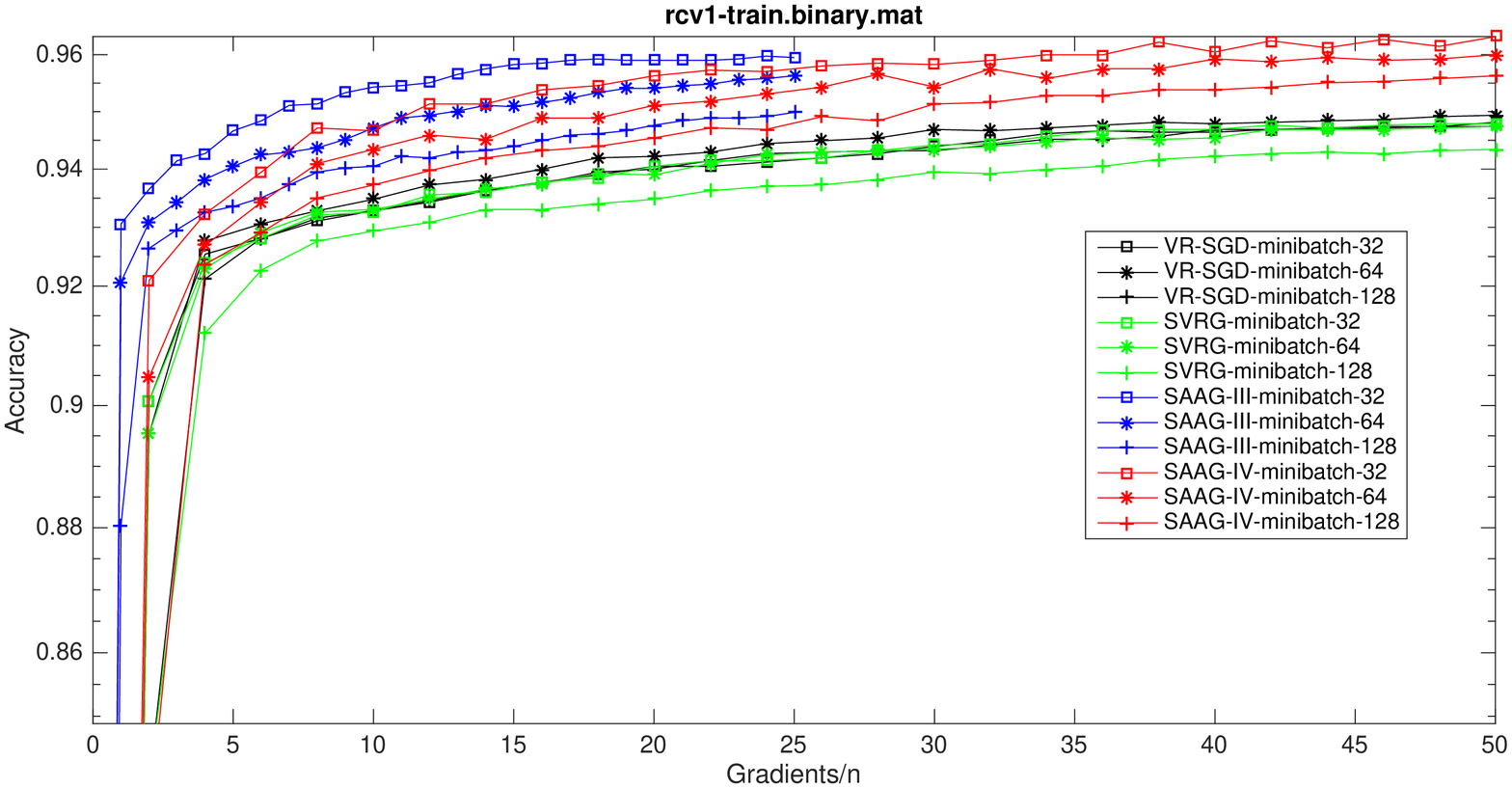}}
	\subfloat{\includegraphics[width=.332\linewidth]{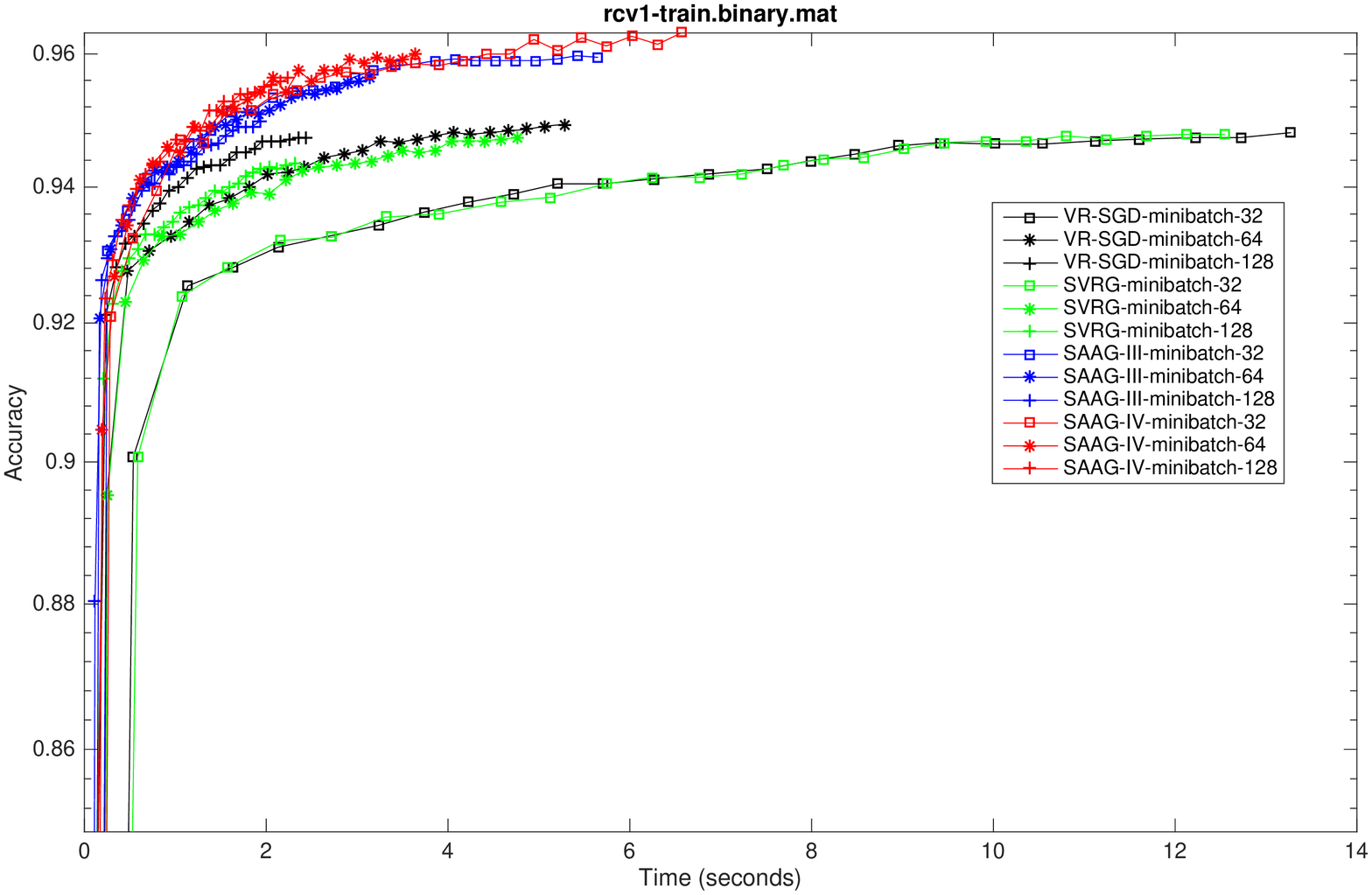}}
	
	\subfloat{\includegraphics[width=.332\linewidth]{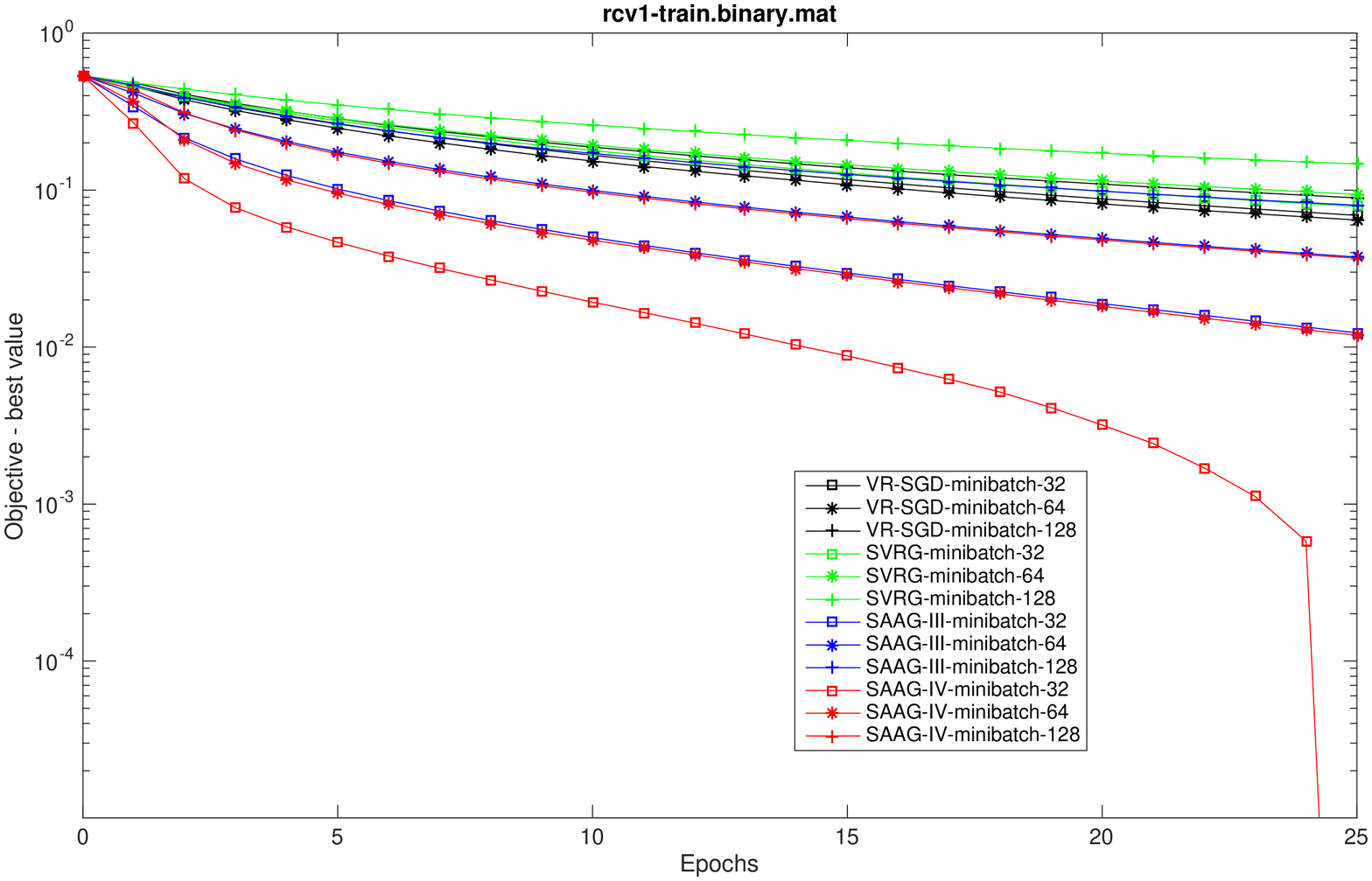}}
	\subfloat{\includegraphics[width=.332\linewidth]{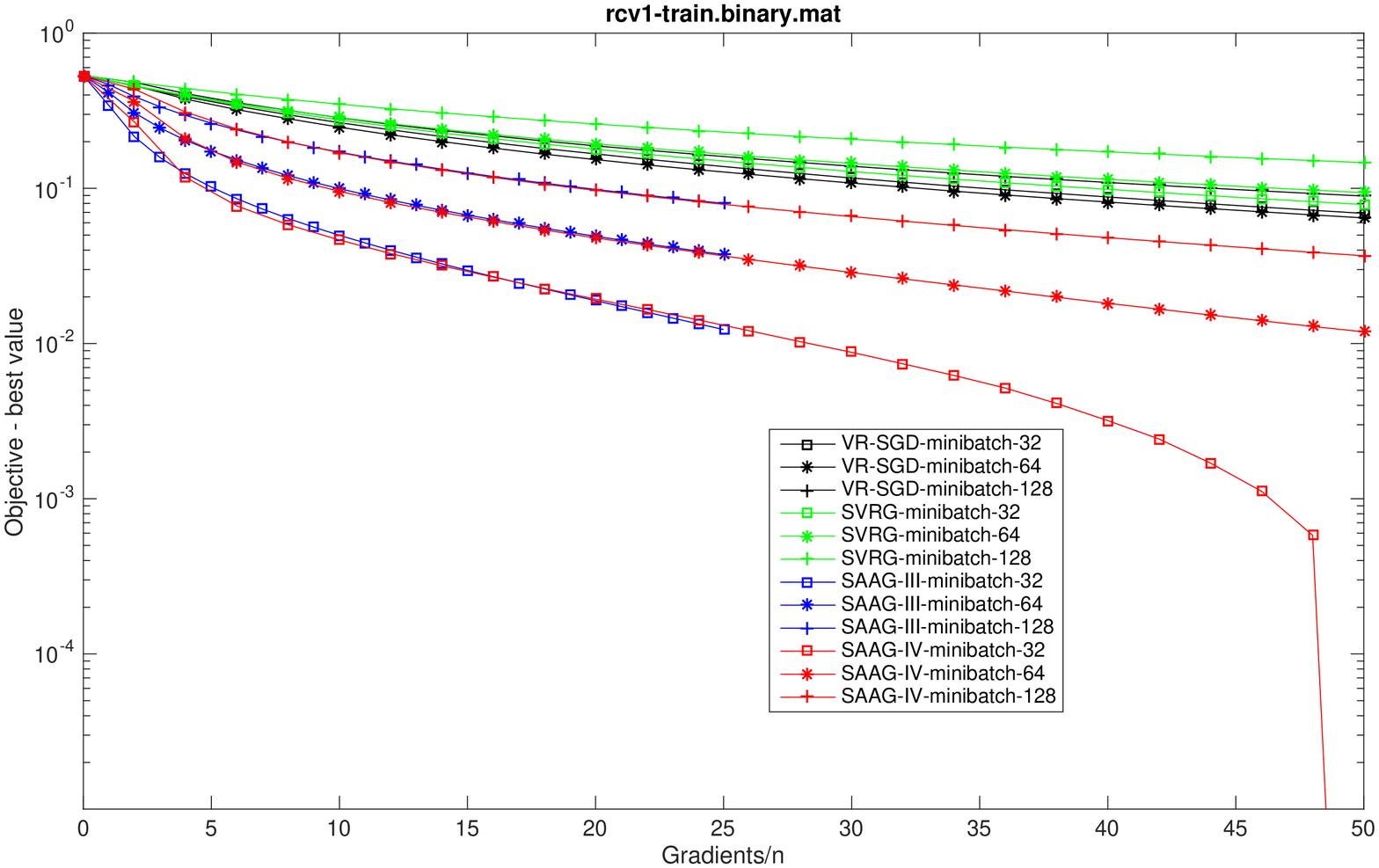}}
	\subfloat{\includegraphics[width=.332\linewidth]{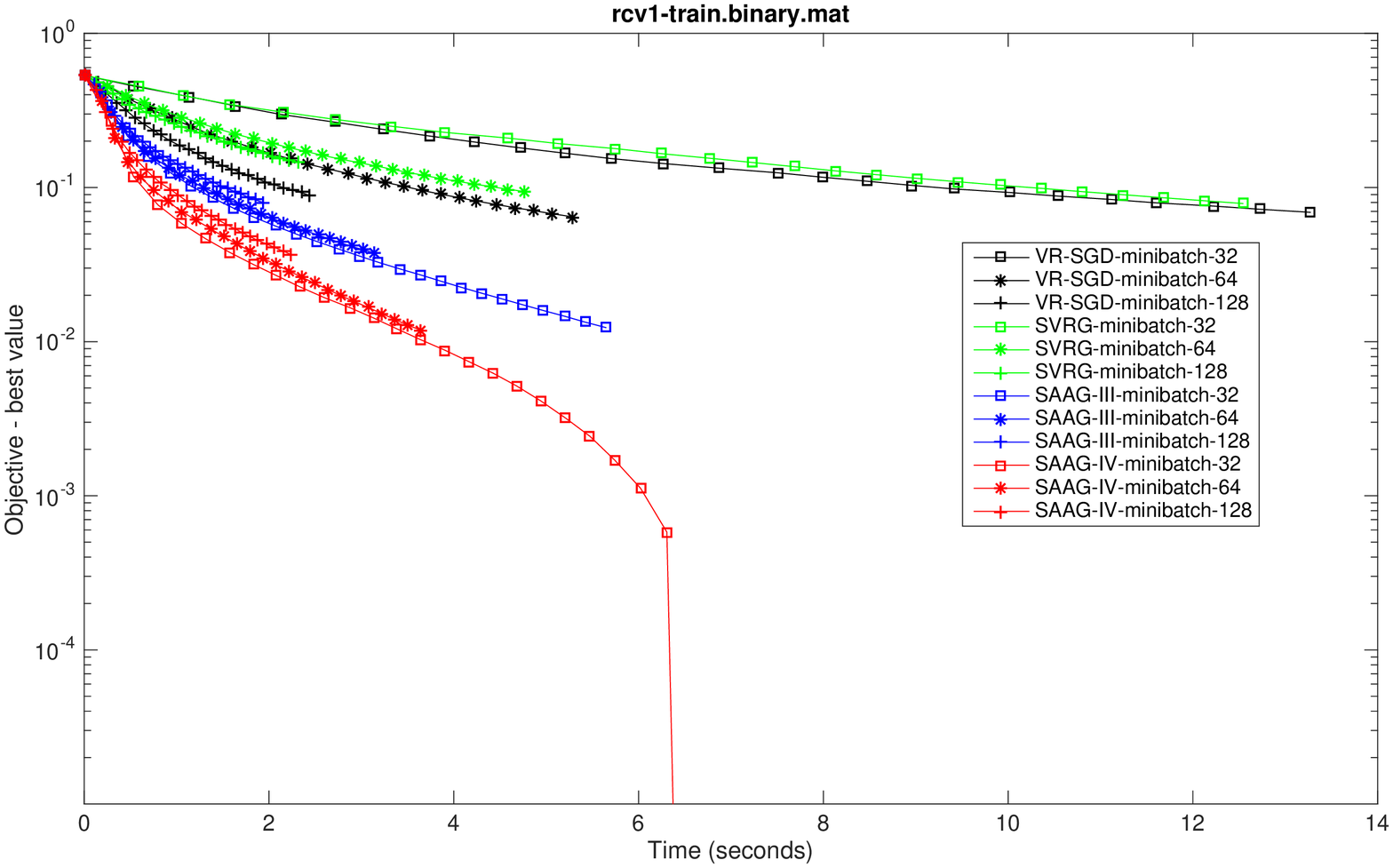}}
	
	\caption{Study of effect of mini-batch size on SAAG-III, IV, SVRG and VR-SGD for non-smooth problem, using rcv1 dataset with mini-batch sizes of 32, 64 and 128. First row compares accuracy against epochs, gradients/n and time, and second row compares suboptimality against epochs, gradients/n and time.}
	\label{fig_minibatchsize_nonsmooth}
\end{figure}

Effect of mini-batch size on SAAG-III, IV, SVRG and VR-SGD for non-smooth problem is depicted in Figure~\ref{fig_minibatchsize_nonsmooth} using rcv1 binary dataset with mini-batch of 32, 64 and 128 data points. Similar to smooth problem, proposed methods outperform SVRG and VR-SGD methods. SAAG-IV gives the best result in terms of time and epochs but in terms of gradients/n, SAAG-III gives best results.
% Comparison of effect of mini-batch size.

\subsection{Effect of mini-batch size on SAAGs (I, II, III, IV) for smooth problem}
\label{appsub_mini-batch_effect_saags_smooth}
\begin{figure}[htb]
	\subfloat{\includegraphics[width=.332\linewidth]{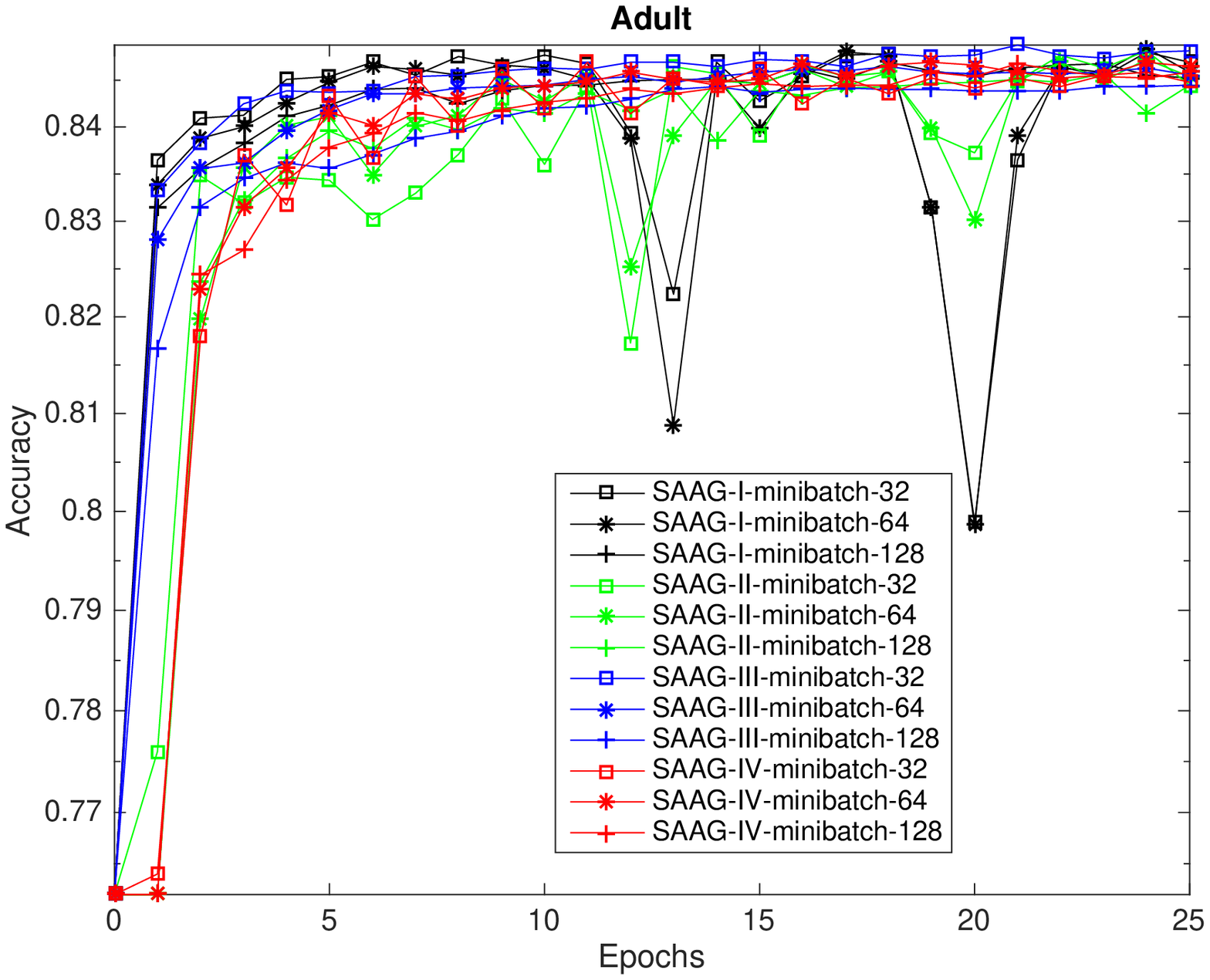}}
	\subfloat{\includegraphics[width=.332\linewidth]{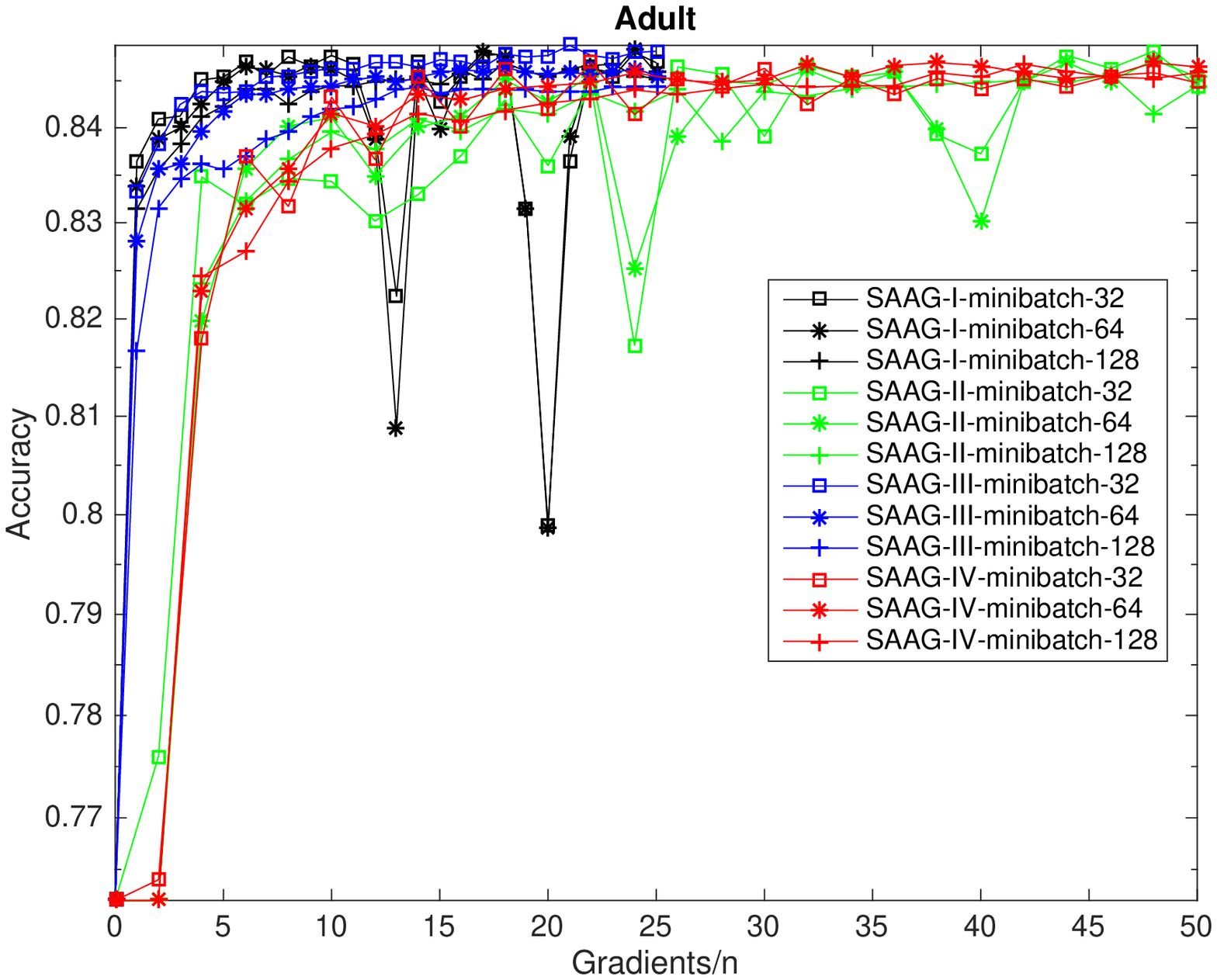}}
	\subfloat{\includegraphics[width=.332\linewidth]{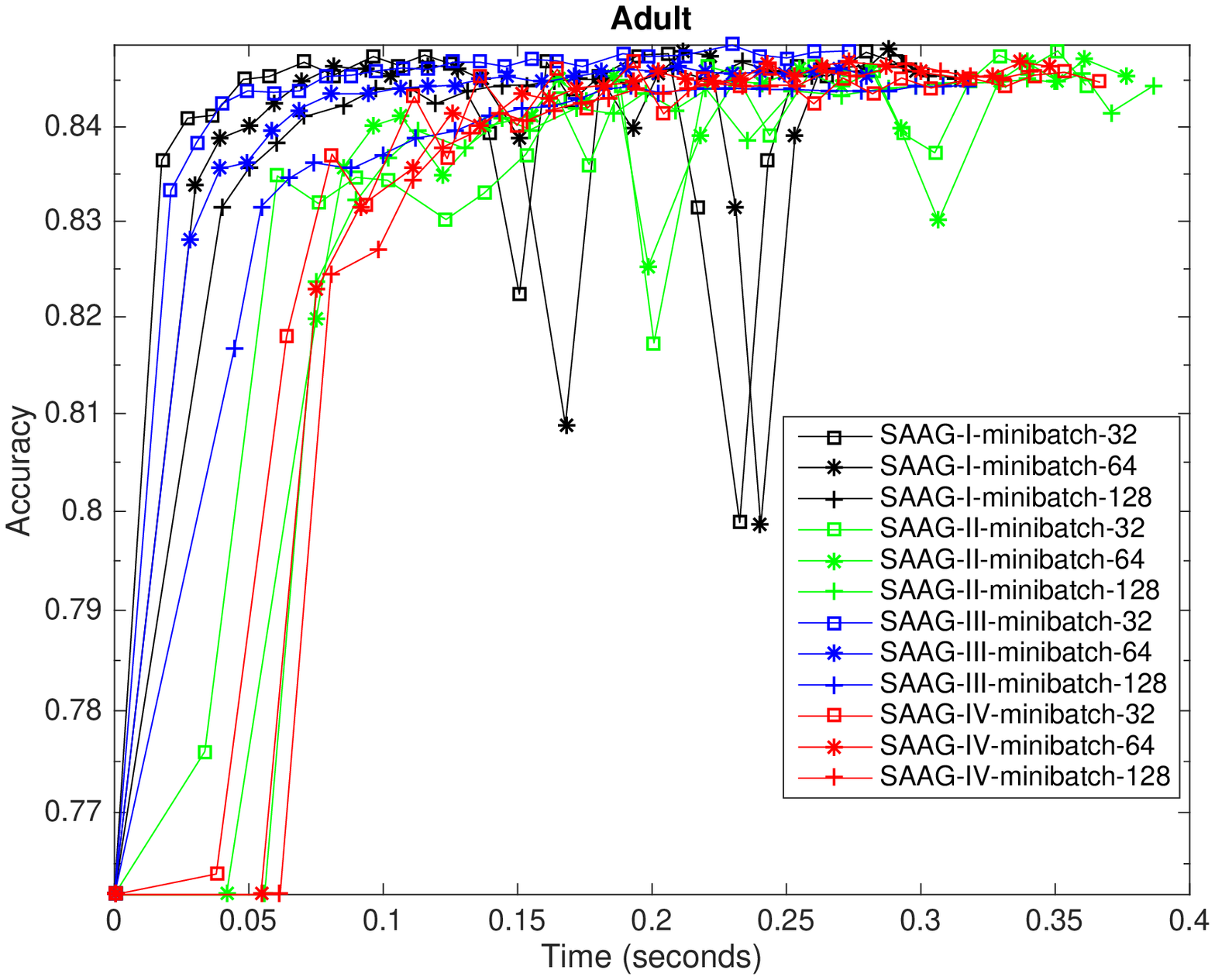}}
	
	\subfloat{\includegraphics[width=.332\linewidth]{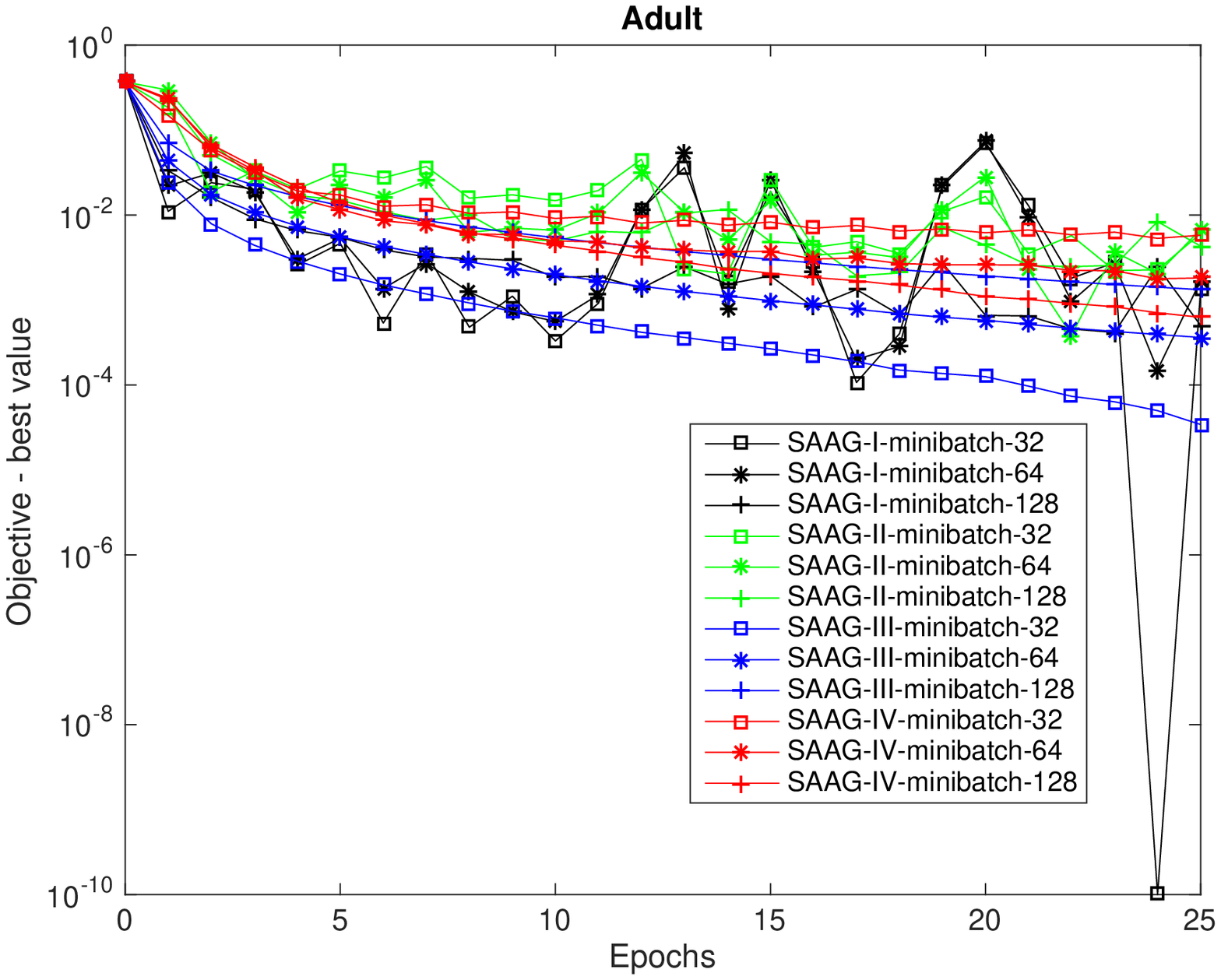}}
	\subfloat{\includegraphics[width=.332\linewidth]{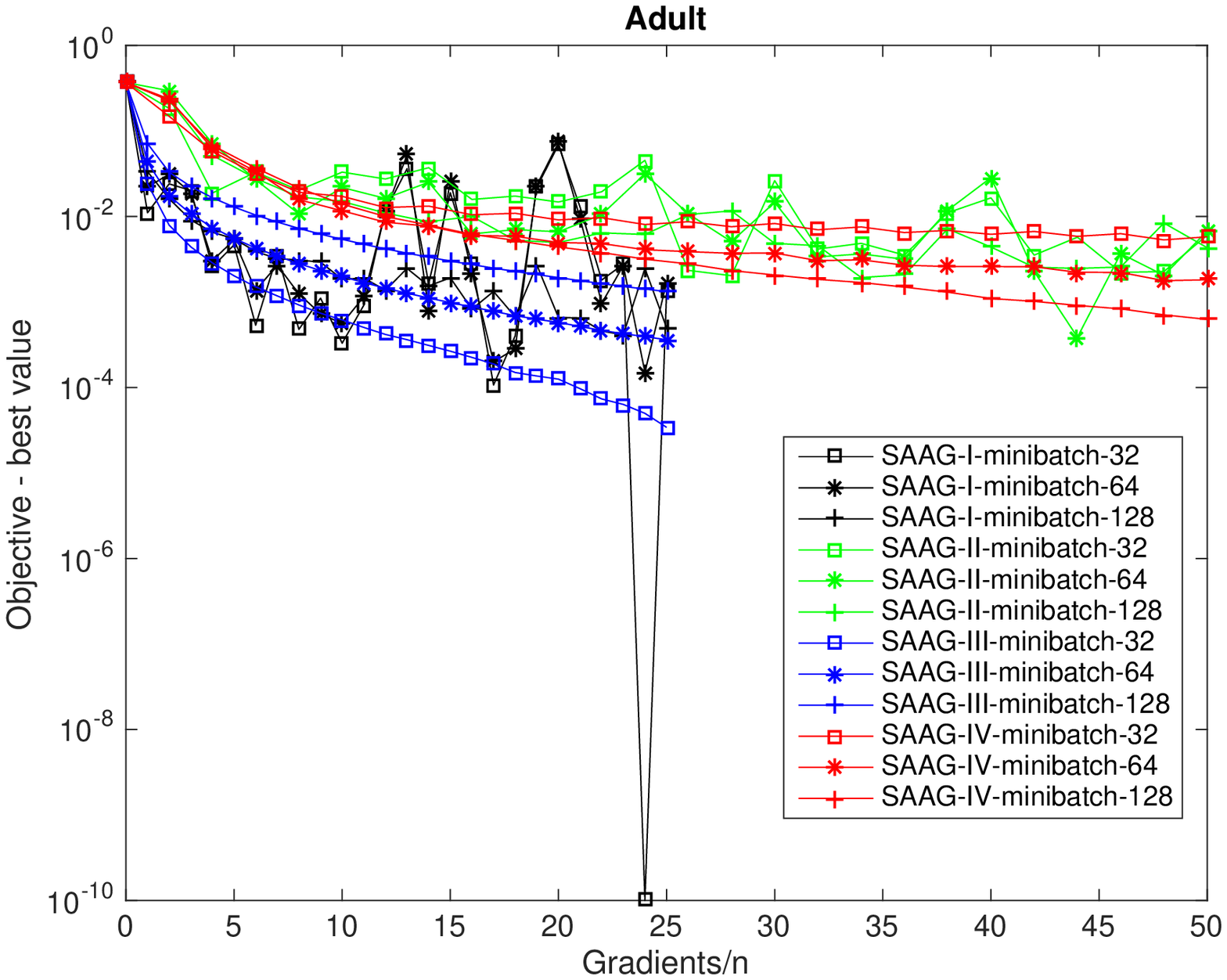}}
	\subfloat{\includegraphics[width=.332\linewidth]{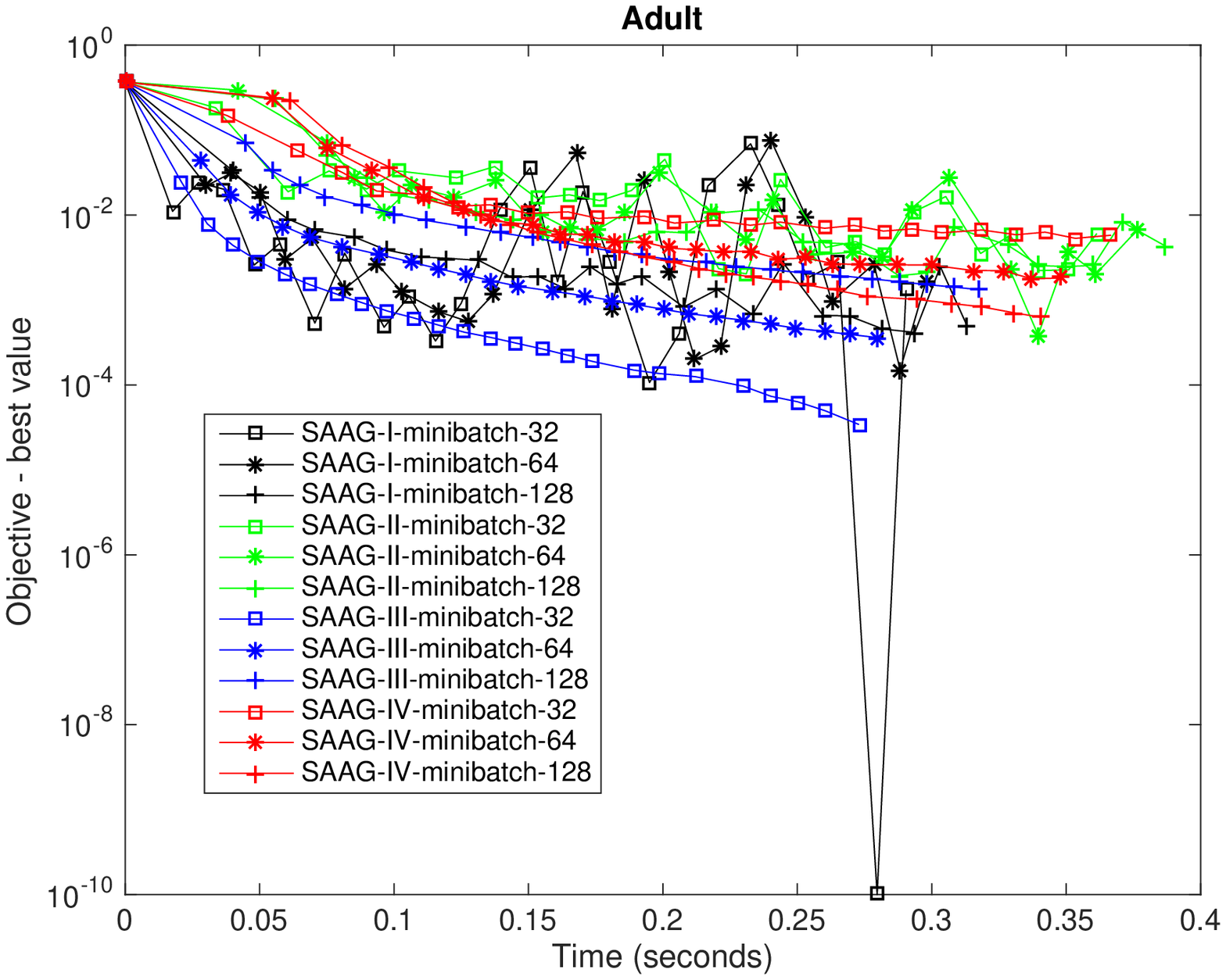}}
	
	\caption{Study of effect of mini-batch size on SAAGs (I, II, III, IV) for smooth problem, using Adult dataset with mini-batch sizes of 32, 64 and 128. First row compares accuracy against epochs, gradients/n and time, and second row compares suboptimality against epochs, gradients/n and time.}
	\label{fig_minibatchsize_saags_smooth}
\end{figure}

Effect of mini-batch size on SAAGs (I, II, III, IV) for smooth problem is depicted in Figure~\ref{fig_minibatchsize_saags_smooth} using Adult dataset with mini-batch sizes of 32, 64 and 128 data points. The results are similar to non-smooth problem.
% Comparison of effect of mini-batch size.

\subsection{Effect of regularization coefficient for non-smooth problem}
\label{appsub_regularization_nonsmooth}
\begin{figure}[htb]
	\subfloat{\includegraphics[width=.332\linewidth]{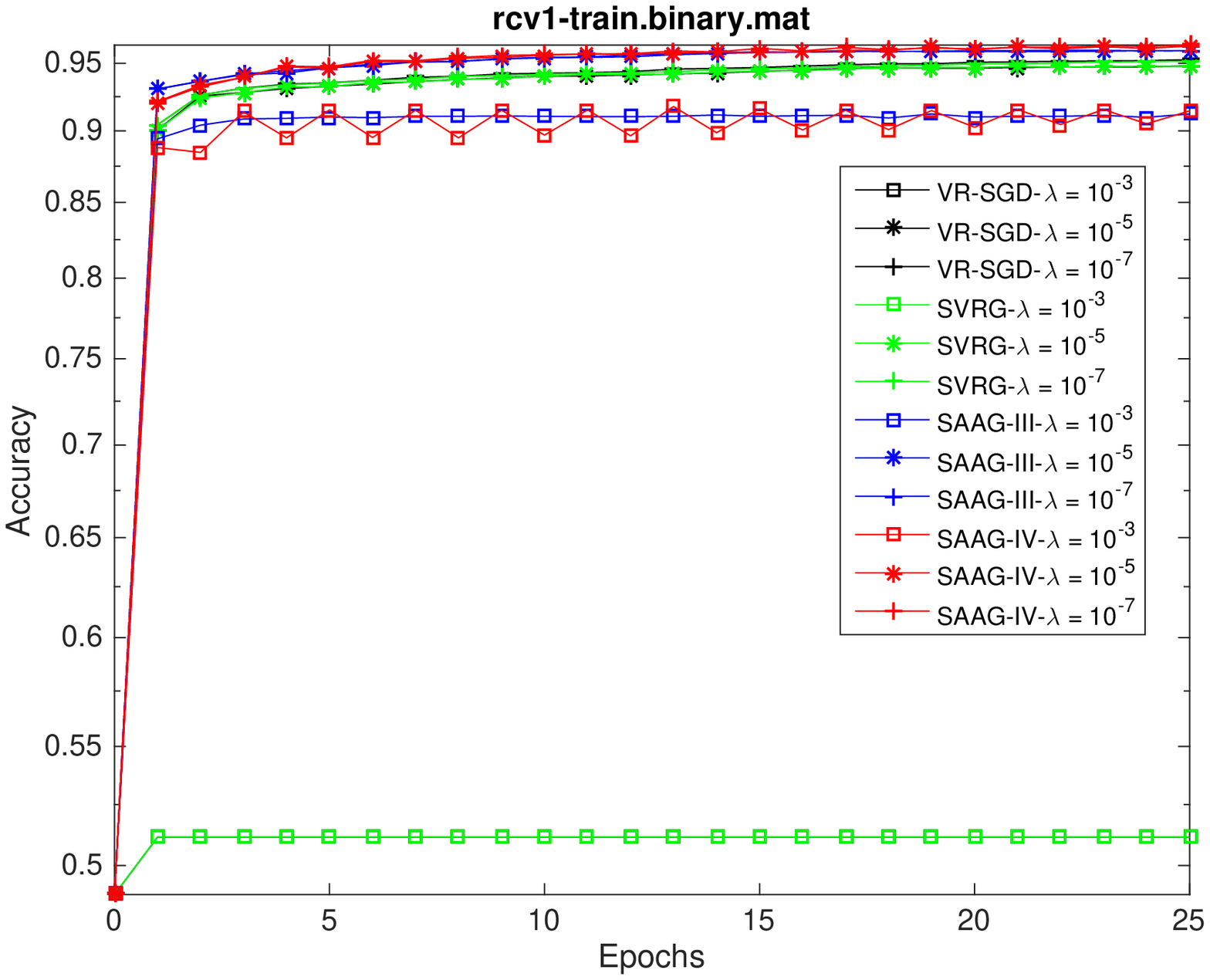}}
	\subfloat{\includegraphics[width=.332\linewidth]{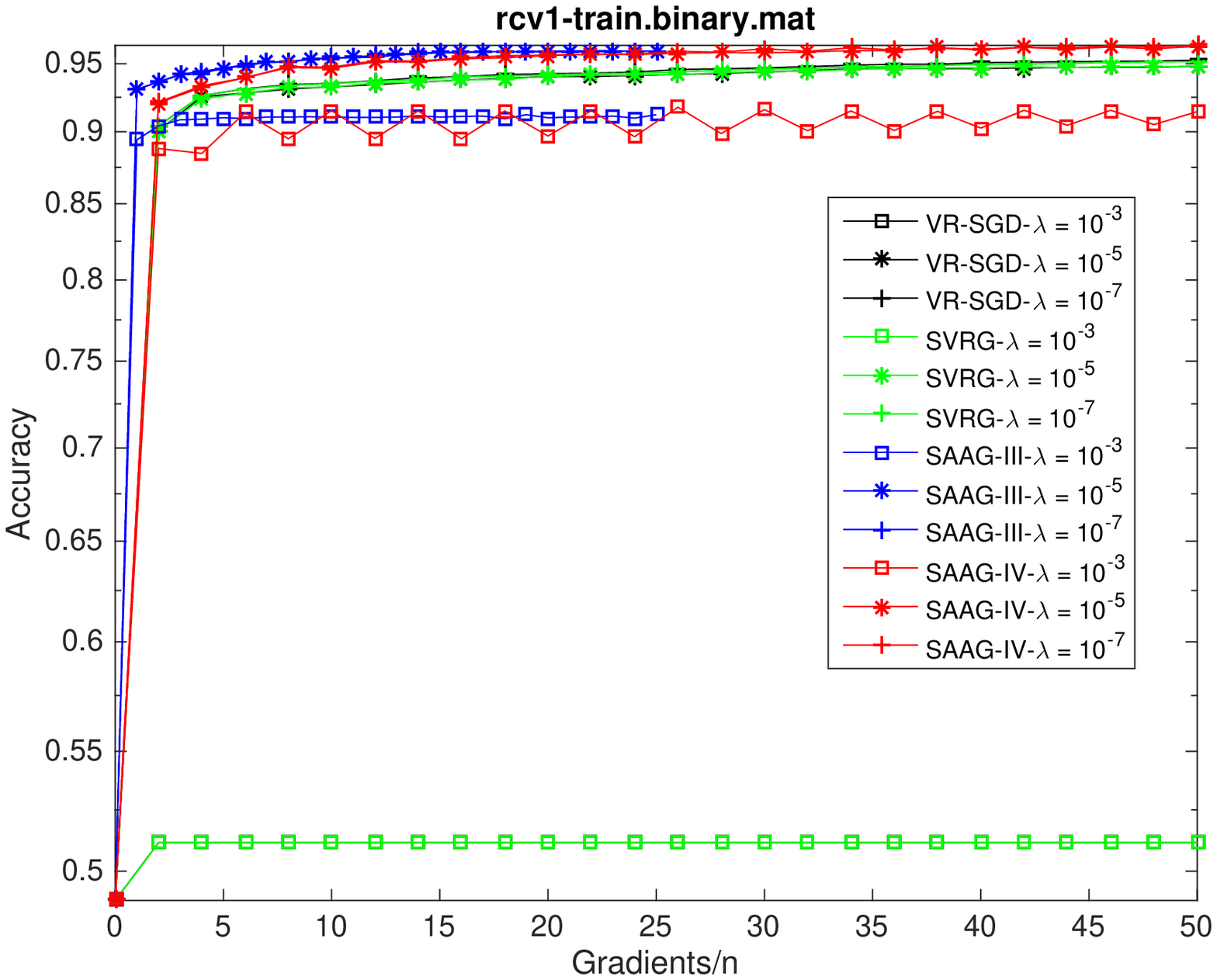}}
	\subfloat{\includegraphics[width=.332\linewidth]{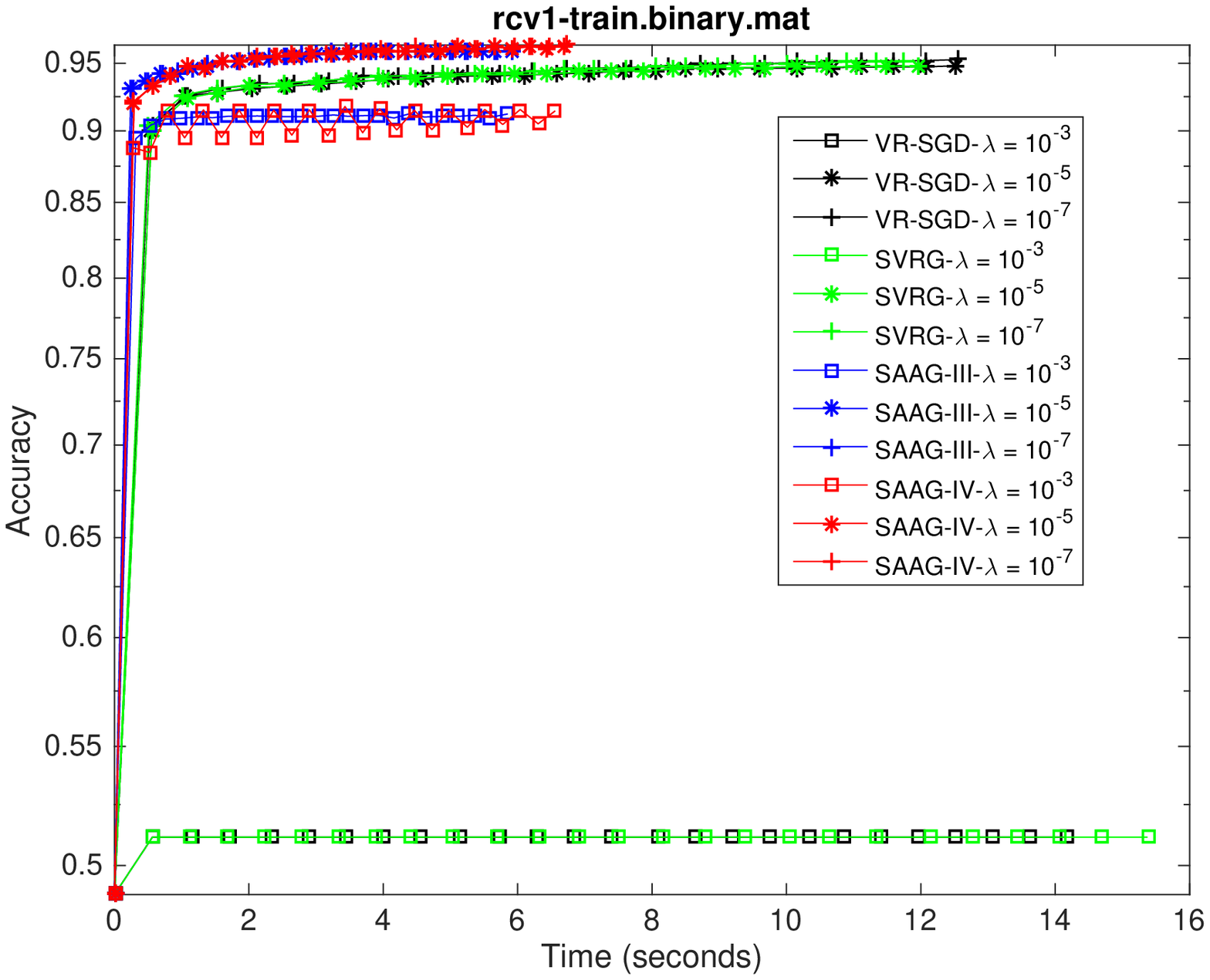}}
	
	\subfloat{\includegraphics[width=.332\linewidth]{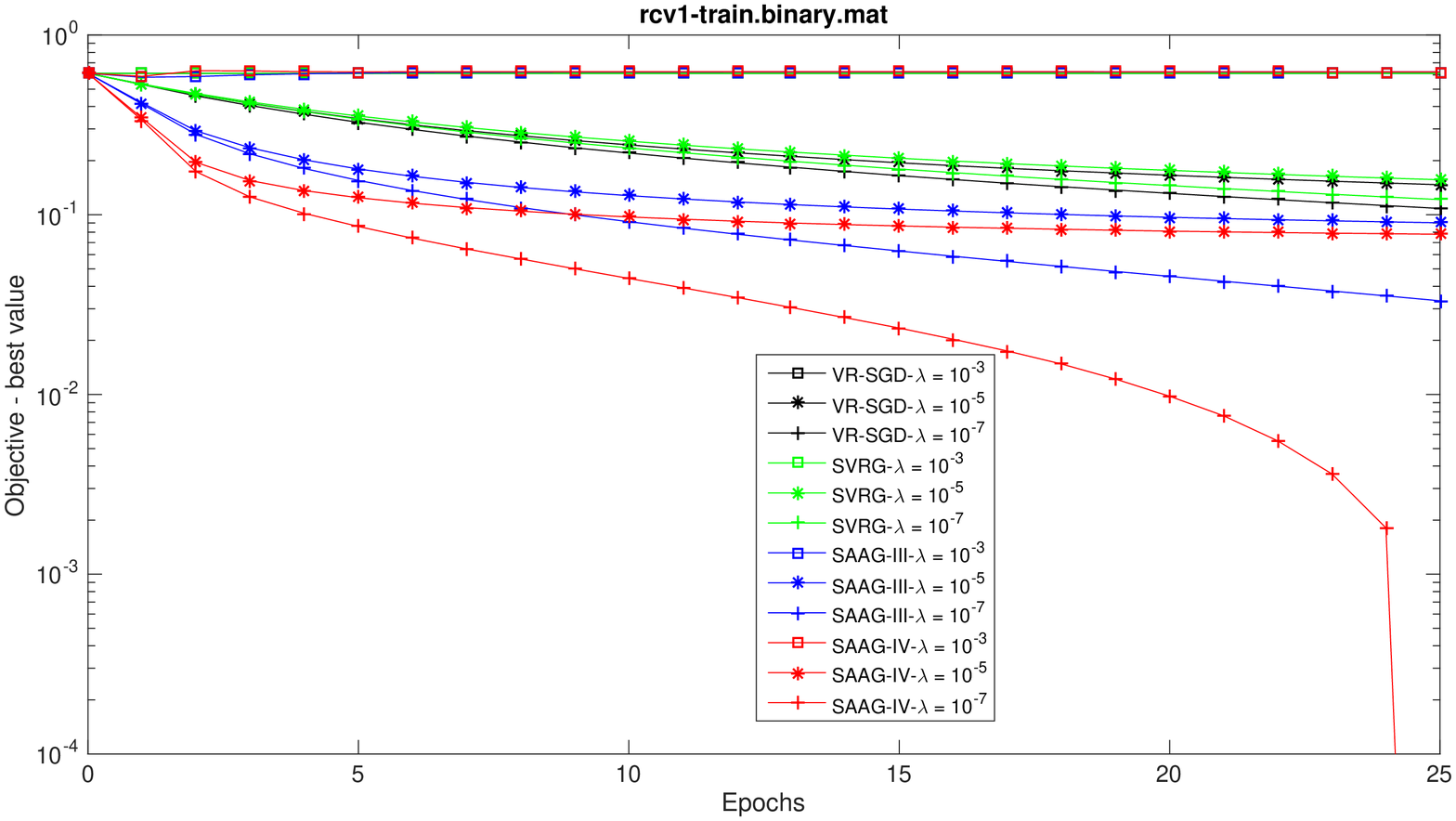}}
	\subfloat{\includegraphics[width=.332\linewidth]{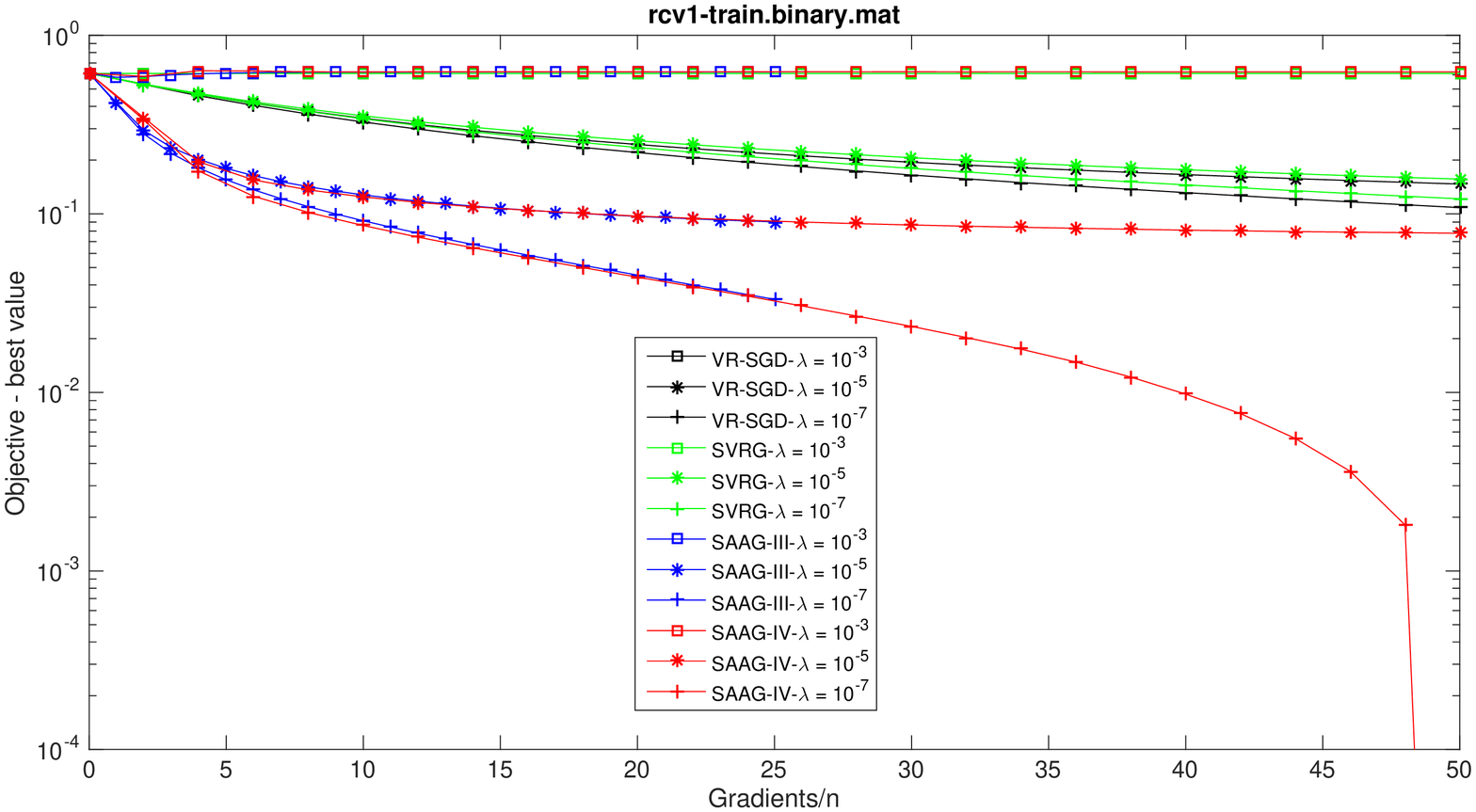}}
	\subfloat{\includegraphics[width=.332\linewidth]{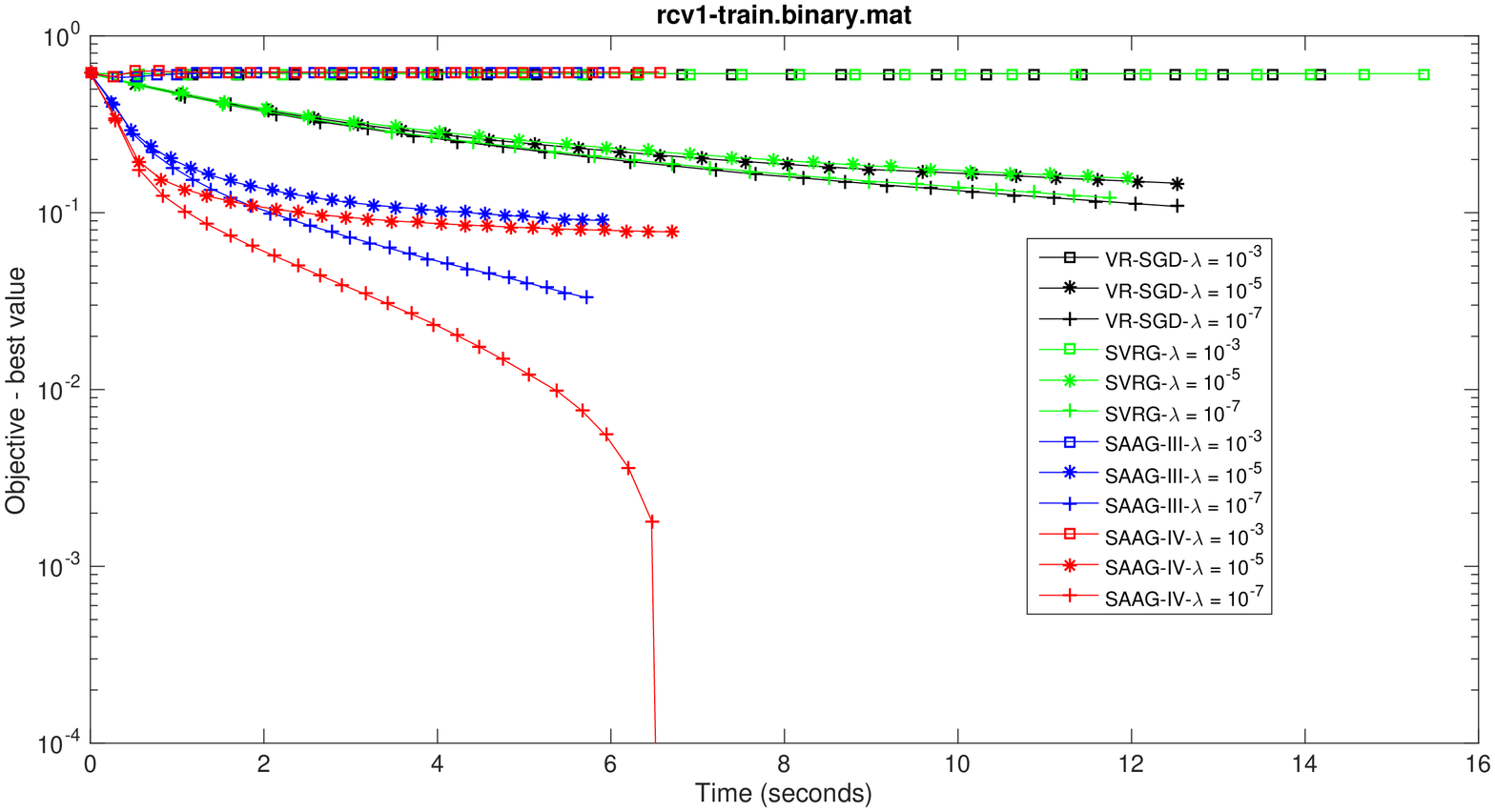}}
	
	\caption{Study of effect of regularization coefficient on SAAG-III, IV, SVRG and VR-SGD for non-smooth problem using rcv1 dataset and taking regularization coefficient values as $10^{-3}$, $10^{-5}$ and $10^{-7}$. First row compares accuracy against epochs, gradients/n and time, and second row compares suboptimality against epochs, gradients/n and time.}
	\label{fig_regularization_nonsmooth}
\end{figure}

Figure~\ref{fig_regularization_nonsmooth} depicts effect of regularization coefficient on SAAG-III, IV, SVRG and VR-SGD for non-smooth problem using rcv1 dataset. It considers regularization coefficient values as $10^{-3}$, $10^{-5}$ and $10^{-7}$. The results are similar to smooth problem. As it is clear from the figure, for larger values, $10^{-3}$, all the methods do not perform well but once the coefficient is sufficiently small, it does not make much difference, and in all the cases our proposed methods outperform SVRG and VR-SGD.
% Comparison of effect of mini-batch size.

\section{Proofs}
%\subsection{Assumptions}
%\label{subsec_assumptions}
Following assumptions are considered in the paper:
\begin{assumption1}[Smoothness]
	\label{assump_lipschitz}
	Suppose function $f_i: \mathbb{R}^n \rightarrow \mathbb{R}$ is convex and differentiable, and that gradient $\nabla f_i, \; \forall i$ is $L$-Lipschitz-continuous, where $L>0$ is Lipschitz constant, then, we have,
	\begin{equation}
	\label{eq_lipschitz1}
	\| \nabla f_i(y) - \nabla f_i(x)\| \le L \|y-x\|,
	\end{equation}
	\begin{equation}
	\label{eq_lipschitz2}
	\begin{array}{ll}
	\text{and},\quad f_i(y) \le f_i(x) +\nabla f_i(x)^{T}(y-x)+ \dfrac{L}{2} \|y-x\|^2.
	\end{array}
	\end{equation}
\end{assumption1}

\begin{assumption1}[Strong Convexity]
	\label{assump_sconvexity}
	Suppose function $F: \mathbb{R}^n \rightarrow \mathbb{R}$ is $\mu$-strongly convex function for $\mu>0$ and $F^{*}$ is the optimal value of $F$, then, we have,
	\begin{equation}
	\label{convexivity_eq1}
	F(y) \ge F(x) +\nabla F(x)^T(y-x) + \dfrac{\mu}{2} \|y-x\|^2,
	\end{equation}
	\begin{equation}
	\label{eq_convex1}
	\begin{array}{ll}
	\text{and},\quad & F(x) - F^{*} \le \dfrac{1}{2\mu}\|\nabla F(x)||^2
	\end{array}
	\end{equation}
\end{assumption1}

\begin{assumption1}[Assumption 3 in \cite{Shang2018}]
	\label{assump_3shang}
	For all $s=1,2,...,S$, the following inequality holds
	\begin{equation}
	\mathbb{E}\left[F(w^s_0) - F(w^{*}) \right] \le c\mathbb{E}\left[F(\tilde{w}^{s-1}) - F(w^{*}) \right]
	\end{equation}
	where $0 < c \ll m$ is a constant.
\end{assumption1}

\label{app_proofs}
We derive our proofs by taking motivation from \cite{Shang2018} and \cite{Xiao2014}. Before providing the proofs, we provide certain lemmas, as given below:
\begin{lemma1}[3-Point Property \cite{Lan2012}]
	\label{lemma_3pt}
	Let $\hat{z}$ be the optimal solution of the following problem:
	$\underset{z\in\mathbb{R}^d}{\min}\quad \dfrac{\tau}{2} \|z-z_0\|^2 + r(z),$
	where $\tau \ge 0$ and $r(z)$ is a convex function (but possibly non-differentiable). Then for any $z\in\mathbb{R}^d$, then the following inequality holds,
	\begin{equation}
	\label{eq_3ptprop}
	\dfrac{\tau}{2} \|\hat{z}-z_0\|^2 + r(\hat{z}) \le r(z) + \dfrac{\tau}{2} \left(\|z-z_0\|^2  - \|z - \hat{z}\|^2 \right)
	\end{equation}
\end{lemma1}

\begin{lemma1}[Theorem 4 in \cite{Konecny2016}] 
	\label{lemma_var_ns}
	For non-smooth problems, taking $\tilde{\nabla}^{'}_{s,k} = \dfrac{1}{b} \sum_{i \in B_k} \nabla f_i (w^s_k) - \dfrac{1}{b} \sum_{i \in B_k} \nabla f_i (\tilde{w}^{s-1}) + \dfrac{1}{n} \sum_{i=1}^{n}f_i(\tilde{w}^{s-1})$, we have $\mathbb{E} \left[\tilde{\nabla}^{'}_{s,k}\right] = \nabla f(w^s_k)$ and the variance satisfies following inequality,
	\begin{equation}
	\label{eq_vb_Konecny}
	\mathbb{E} \left[\|\tilde{\nabla}^{'}_{s,k} - \nabla f(w^s_k)\|^2\right] \le 4L\alpha(b) \left[ F(w^s_k) - F(w^{*}) + F(\tilde{w}^{s-1}) - F(w^{*}) \right],
	\end{equation}
	where $\alpha(b) = (n-b)/(b(n-1))$.
\end{lemma1}
Following the Lemma~\ref{lemma_var_ns} for non-smooth problems, one can easily prove the following results for the smooth problems,
\begin{lemma1}
	\label{lemma_var_s}
	For smooth problems, taking $\tilde{\nabla}^{'}_{s,k} = \dfrac{1}{b} \sum_{i \in B_k} \nabla f_i (w^s_k) - \dfrac{1}{b} \sum_{i \in B_k} \nabla f_i (\tilde{w}^{s-1}) + \dfrac{1}{n} \sum_{i=1}^{n}f_i(\tilde{w}^{s-1})$, we have $\mathbb{E} \left[\tilde{\nabla}^{'}_{s,k}\right] = \nabla f(w^s_k)$ and the variance satisfies following inequality,
	\begin{equation}
	\label{eq_vb_shang}
	\mathbb{E} \left[\|\tilde{\nabla}^{'}_{s,k} - \nabla f(w^s_k)\|^2\right] \le 4L\alpha(b) \left[ f(w^s_k) - f^{*} + f(\tilde{w}^{s-1}) - f^{*}\right],
	\end{equation}
	where $\alpha(b) = (n-b)/(b(n-1))$.
\end{lemma1}

\begin{lemma1}[Extension of Lemma~3.4 in \cite{Xiao2014} to mini-batches]
	\label{lemma_bound_prox_ext1}
	Under Assumption~\ref{assump_lipschitz} for smooth regularizer, we have
	\begin{equation}
	\label{eq_vb_xio1}
	\mathbb{E} \left[\|\nabla_{B_k} f(w^s_k) - \nabla_{B_k} f(w^{*})\|^2\right] \le 2L \left[ f(w^s_k) - f(w^{*}) \right]
	\end{equation}
\end{lemma1}
\begin{proof}
	Given any $k=0,1,...,(m-1)$, consider the function, 
	\begin{equation*}
	\phi_{B_k} (w) = f_{B_k} (w) - f_{B_k}(w^{*}) - \nabla_{B_k} f(w^{*})^T (w-w^{*})
	\end{equation*}
	It is straightforward to check that $\nabla \phi_{B_k} (w^{*}) =0$, hence $\min_{w}\; \phi_{B_k} (w) = \phi_{B_k} (w^{*}) = 0.$ Since $\phi_{B_k} (w)$ is Lipschitz continuous so we have,
	\begin{equation*}
	\begin{aligned}
	\dfrac{1}{2L}\|\nabla\phi_{B_k} (w)\|^2 \le \phi_{B_k} (w) - \min_{w}\; \phi_{B_k} (w) =  \phi_{B_k} (w) - \phi_{B_k} (w^{*}) = \phi_{B_k} (w)\\
	\implies \| \nabla f_{B_k} (w) - \nabla f_{B_k}(w^{*})\|^2 \le 2L \left[ f_{B_k} (w) - f_{B_k}(w^{*}) - \nabla_{B_k} f(w^{*})^T (w-w^{*}) \right]
	\end{aligned}
	\end{equation*}
	Taking expectation, we have
	\begin{equation}
	\label{eq_vb_xio3}
	\begin{aligned}
	\mathbb{E}[\| \nabla f_{B_k} (w) - \nabla f_{B_k}(w^{*})\|^2 ] &\le 2L \left[ f (w) - f(w^{*}) - \nabla f(w^{*})^T (w-w^{*}) \right]
	\end{aligned}
	\end{equation}
	By optimality, $\nabla f(w^{*})=0$, we have
	\begin{equation*}
	\begin{aligned}
	\mathbb{E}[\| \nabla f_{B_k} (w) - \nabla f_{B_k}(w^{*})\|^2 ] \le 2L \left[ f (w) - f(w^{*}) \right]
	\end{aligned}
	\end{equation*}
	This proves the required lemma.
\end{proof}

\begin{lemma1}[Extension of Lemma~3.4 in \cite{Xiao2014} to mini-batches]
	\label{lemma_bound_prox_ext2}
	Under Assumption~\ref{assump_lipschitz} for non-smooth regularizer, we have
	\begin{equation}
	\label{eq_vb_xio2}
	\mathbb{E} \left[\|\nabla_{B_k}f(w^s_k) - \nabla_{B_k} f(w^{*})\|^2\right] \le 2L \left[ F(w^s_k) - F(w^{*}) \right]
	\end{equation}
\end{lemma1}
\begin{proof}
	From inequality~\ref{eq_vb_xio3}, we have
	\begin{equation}
	\begin{aligned}
	\mathbb{E}[\| \nabla f_{B_k} (w) - \nabla f_{B_k}(w^{*})\|^2 ] &\le 2L \left[ f (w) - f(w^{*}) - \nabla f(w^{*})^T (w-w^{*}) \right]
	\end{aligned}
	\end{equation}
	By optimality, there exist $\xi \in \partial g (w^{*})$, such that, $\nabla F(w^{*}) = \nabla f(w^{*}) + \xi=0$, we have
	\begin{equation}
	\begin{aligned}
	\mathbb{E}[\| \nabla f_{B_k} (w) - \nabla f_{B_k}(w^{*})\|^2 ] &\le 2L \left[ f (w) - f(w^{*}) + \xi^T (w-w^{*}) \right]\\
	&\le 2L \left[ f (w) - f(w^{*}) + g(w) - g (w^{*}) \right]\\
	&\le 2L \left[ F(w) - F(w^{*})\right]
	\end{aligned}
	\end{equation}
	second inequality follows from the convexity of $g$. This proves the required lemma.
\end{proof}

\begin{lemma1}[Variance Bound for smooth problem]
	\label{lemma_variance_bound_smooth}
	Under the Assumption~\ref{assump_lipschitz} and taking $\nabla_{B_k} f(w^s_k) = \dfrac{1}{b} \sum_{i \in B_k} \nabla f_i (w^s_k)$, $ \nabla_{B^{'}_k} f(\tilde{w}^{s-1}) = \dfrac{1}{n} \sum_{i \in B_k} \nabla f_i (\tilde{w}^{s-1})$, $\tilde{\mu}^s = \dfrac{1}{n} \sum_{i=1}^{n} \nabla f_i(\tilde{w}^{s-1})$ and the gradient estimator, $\tilde{\nabla}_{s,k} = \nabla_{B_k} f(w^s_k) - \nabla_{B^{'}_k} f(\tilde{w}^{s-1}) +\tilde{\mu}^s $, then the variance satisfies the following inequality\footnote{\label{footnote_1}For the simplification of proof, we take $ f(w) = F(w)$, i.e., $f_i(w) = f_i(w) + g(w)\; \forall i$ and then $g(w) \equiv0$},
	\begin{equation}
	\label{eq_vb}
	\mathbb{E} \left[\| \tilde{\nabla}_{s,k} -\nabla f(w^s_k)\|^2 \right] \le 8L\alpha(b) \left[ f(w^s_k) - f^{*} \right] + \dfrac{8L\left(\alpha(b)m^2+(m-1)^2\right)}{m^2} \left[ f(\tilde{w}^{s-1}) - f^{*} \right] + R^{'}
	\end{equation}
	where $\alpha(b) = (n-b)/(b(n-1))$ and $R^{'}$ is a constant.
\end{lemma1}
\begin{proof}
	First the expectation of estimator is given by
	\begin{equation}
	\label{eq_expectation_estimator}
	\begin{aligned}
	\mathbb{E}\left[\tilde{\nabla}_{s,k} \right] &= \mathbb{E}\left[\nabla_{B_k} f(w^s_k) - \nabla_{B^{'}_k} f(\tilde{w}^{s-1}) +\tilde{\mu}^s \right]\\
	&= \nabla f(w^s_k) - \dfrac{b}{n} \nabla f(\tilde{w}^{s-1}) + \nabla f(\tilde{w}^{s-1})\\
	& = \nabla f(w^s_k) + \dfrac{m-1}{m} \nabla f(\tilde{w}^{s-1}),
	\end{aligned}
	\end{equation}
	second equality follows as $n = mb$. Now the variance bound is calculated as follows,
	\begin{equation}
	\label{eq_vb_1}
	\begin{aligned}
	&\mathbb{E} \left[ \| \tilde{\nabla}_{s,k} - \nabla f(w^s_k) \|^2\right]\\ &= \mathbb{E}\left[\| \nabla_{B_k} f(w^s_k) - \nabla_{B^{'}_k} f(\tilde{w}^{s-1}) +\nabla f (\tilde{w}^{s-1}) - \nabla f (w^s_k)\|^2\right]\\
	&= \mathbb{E}\left[\| \nabla_{B_k} f(w^s_k) - \nabla_{B_k} f(\tilde{w}^{s-1}) +\nabla f (\tilde{w}^{s-1}) - \nabla f (w^s_k) + \dfrac{m-1}{m}\nabla_{B_k} f(\tilde{w}^{s-1}) \|^2\right]\\
	& \le 2 \mathbb{E} \left[ \| \nabla_{B_k} f(w^s_k) - \nabla_{B_k} f(\tilde{w}^{s-1}) +\nabla f (\tilde{w}^{s-1}) - \nabla f (w^s_k)\|^2\right] + \dfrac{2(m-1)^2}{m^2} \mathbb{E} \left[ \| \nabla_{B_k} f(\tilde{w}^{s-1}) \|^2 \right]\\
	& \le 8L\alpha(b) \left[ f(w^s_k) - f^{*} + f(\tilde{w}^{s-1}) - f^{*} \right] + \dfrac{2(m-1)^2}{m^2} \mathbb{E} \left[ \| \nabla_{B_k} f(\tilde{w}^{s-1}) \|^2 \right]
	\end{aligned}
	\end{equation}
	inequality follows from, $\|a+b\|^2 \le 2\left( \|a\|^2 + \|b\|^2 \right)$ for $a,b\in \mathbb{R}^d$ and applying the Lemma~\ref{lemma_var_s}.
	\begin{equation}
	\begin{aligned}
	\text{Now, }&\dfrac{2(m-1)^2}{m^2} \mathbb{E} \left[ \| \nabla_{B_k} f(\tilde{w}^{s-1}) \|^2 \right]\\
	&\le   \dfrac{2(m-1)^2}{m^2} \left[2 \mathbb{E}  \|\nabla_{B_k} f(\tilde{w}^{s-1}) - \nabla_{B_k} f(w^{*}) \|^2 + 2 \mathbb{E}  \|  \nabla_{B_k} f(w^{*}) \|^2 \right]\\
	& \le   \dfrac{8L(m-1)^2}{m^2}\left[ f(\tilde{w}^{s-1}) - f(w^{*})  \right] + R^{'}\\
	\end{aligned}
	\end{equation}
	first inequality follows from, $\|a+b\|^2 \le 2\left( \|a\|^2 + \|b\|^2 \right)$ for $a,b\in \mathbb{R}^d$, second inequality follows from Lemma~\ref{lemma_bound_prox_ext1} and assuming $\mathbb{E}\| \nabla_{B_k} f(w^{*}) \|^2 \le R,\; \forall k \; \text{and}$ where taking $R^{'} = \dfrac{2(m-1)^2}{m^2}* R$. Now, substituting the above inequality in \ref{eq_vb_1}, we have
	\begin{equation}
	\label{eq_vb_2}
	\begin{aligned}
	& \mathbb{E} \left[ \| \tilde{\nabla}_{s,k} - \nabla f(w^s_k) \|^2\right]\\
	& \le 8L\alpha(b) \left[ f(w^s_k) - f^{*} + f(\tilde{w}^{s-1}) - f^{*} \right] +   \dfrac{8L(m-1)^2}{m^2}\left[ f(\tilde{w}^{s-1}) - f(w^{*})  \right] + R^{'}\\
	&= 8L\alpha(b) \left[ f(w^s_k) - f^{*} \right] + \dfrac{8L\left(\alpha(b)m^2+(m-1)^2\right)}{m^2} \left[ f(\tilde{w}^{s-1}) - f^{*} \right] + R^{'},
	\end{aligned}
	\end{equation}
	This proves the required lemma.
\end{proof}

\begin{lemma1}[Variance Bound for non-smooth problem]
	\label{lemma_variance_bound_nonsmooth}
	Under Assumption~\ref{assump_lipschitz} and taking notations as in Lemma~\ref{lemma_variance_bound_smooth}, the variance bound satisfies the following inequality,
	\begin{equation}
	\label{eq_vb_ns}
	\mathbb{E} \left[\| \tilde{\nabla}_{s,k} -\nabla f(w^s_k)\|^2 \right] \le 8L\alpha(b) \left[ F(w^s_k) - F^{*} \right] + \dfrac{8L\left(\alpha(b)m^2+(m-1)^2\right)}{m^2} \left[ F(\tilde{w}^{s-1}) - F^{*} \right] + R^{'},
	\end{equation}
	where $\alpha(b) = (n-b)/(b(n-1))$ and $R^{'}$ is constant.
\end{lemma1}
\begin{proof}
	\begin{equation}
	\label{eq_vb3}
	\begin{aligned}
	&\mathbb{E} \left[ \| \tilde{\nabla}_{s,k} - \nabla f(w^s_k) \|^2\right]\\ &= \mathbb{E}\left[\| \nabla_{B_k} f(w^s_k) - \nabla_{B^{'}_k} f(\tilde{w}^{s-1}) +\nabla f (\tilde{w}^{s-1}) - \nabla f (w^s_k)\|^2\right]\\
	&= \mathbb{E}\left[\| \nabla_{B_k} f(w^s_k) - \nabla_{B_k} f(\tilde{w}^{s-1}) +\nabla f (\tilde{w}^{s-1}) - \nabla f (w^s_k) + \dfrac{m-1}{m}\nabla_{B_k} f(\tilde{w}^{s-1}) \|^2\right]\\
	& \le 2 \mathbb{E} \left[ \| \nabla_{B_k} f(w^s_k) - \nabla_{B_k} f(\tilde{w}^{s-1}) +\nabla f (\tilde{w}^{s-1}) - \nabla f (w^s_k)\|^2\right] + \dfrac{2(m-1)^2}{m^2} \mathbb{E} \left[ \| \nabla_{B_k} f(\tilde{w}^{s-1}) \|^2 \right]\\
	& \le 8L\alpha(b) \left[ F(w^s_k) - F({w^{*}}) + F(\tilde{w}^{s-1}) - F({w^{*}}) \right] + \dfrac{2(m-1)^2}{m^2} \mathbb{E} \left[ \| \nabla_{B_k} f(\tilde{w}^{s-1}) \|^2 \right]
	\end{aligned}
	\end{equation}
	inequality follows from, $\|a+b\|^2 \le 2\left( \|a\|^2 + \|b\|^2 \right)$ for $a,b\in \mathbb{R}^d$ and applying the Lemma~\ref{lemma_var_ns}.
	\begin{equation}
	\begin{aligned}
	\text{Now, }&\dfrac{2(m-1)^2}{m^2} \mathbb{E} \left[ \| \nabla_{B_k} f(\tilde{w}^{s-1}) \|^2 \right]\\
	&\le   \dfrac{2(m-1)^2}{m^2} \left[2 \mathbb{E}  \|\nabla_{B_k} f(\tilde{w}^{s-1}) - \nabla_{B_k} f(w^{*}) \|^2 + 2 \mathbb{E}  \|  \nabla_{B_k} f(w^{*}) \|^2 \right]\\
	& \le   \dfrac{8L(m-1)^2}{m^2}\left[ F(\tilde{w}^{s-1}) - F(w^{*})  \right] + R^{'}\\
	\end{aligned}
	\end{equation}
	first inequality follows from, $\|a+b\|^2 \le 2\left( \|a\|^2 + \|b\|^2 \right)$ for $a,b\in \mathbb{R}^d$, second inequality follows from Lemma~\ref{lemma_bound_prox_ext2} and assuming $\| \nabla_{B_k} f(w^{*}) \|^2 \le R,\; \forall k$ and taking $R^{'} = \dfrac{2(m-1)^2}{m^2}* R$. Now, substituting the above inequality in \ref{eq_vb3}, we have
	\begin{equation}
	\label{eq_vb_4}
	\begin{aligned}
	& \mathbb{E} \left[ \| \tilde{\nabla}_{s,k} - \nabla f(w^s_k) \|^2\right]\\
	& \le 8L\alpha(b) \left[ F(w^s_k) - F^{*} + F(\tilde{w}^{s-1}) - F^{*} \right] +   \dfrac{8L(m-1)^2}{m^2}\left[ F(\tilde{w}^{s-1}) - F(w^{*})  \right] + R^{'}\\
	&= 8L\alpha(b) \left[ F(w^s_k) - F^{*} \right] + \dfrac{8L\left(\alpha(b)m^2+(m-1)^2\right)}{m^2} \left[ F(\tilde{w}^{s-1}) - F^{*} \right] + R^{'},
	\end{aligned}
	\end{equation}
	This proves the required lemma.
\end{proof}

\textbf{Proof of Theorem~1} (Non-strongly convex and smooth problem with SAAG-IV)\\
\begin{proof}
	By smoothness, we have,
	\begin{equation}
	\label{eq_proof_nsc_s1}
	\begin{aligned}
	f(w^s_{k+1}) & \le f(w^s_k) + <\nabla f(w^s_k), w^s_{k+1} - w^s_k> + \dfrac{L}{2} \| w^s_{k+1} - w^s_k \|^2\\
	& = f(w^s_k) + <\nabla f(w^s_k), w^s_{k+1} - w^s_k> + \dfrac{L\beta}{2} \| w^s_{k+1} - w^s_k \|^2 - \dfrac{L(\beta-1)}{2} \| w^s_{k+1} - w^s_k \|^2\\
	& = f(w^s_k) + <\tilde{\nabla}_{s,k}, w^s_{k+1} - w^s_k> + \dfrac{L\beta}{2} \| w^s_{k+1} - w^s_k \|^2 + <\nabla f(w^s_k) - \tilde{\nabla}_{s,k}, w^s_{k+1} - w^s_k>\\& - \dfrac{L(\beta-1}{2} \| w^s_{k+1} - w^s_k \|^2,
	\end{aligned}
	\end{equation}
	where $\beta$ is appropriately chosen positive value. Now, separately simplifying the terms, we have
	\begin{equation}
	\label{eq_proof_nsc_s2}
	\begin{aligned}
	&\mathbb{E}\left[ f(w^s_k) + <\tilde{\nabla}_{s,k}, w^s_{k+1} - w^s_k> + \dfrac{L\beta}{2} \| w^s_{k+1} - w^s_k \|^2 \right]\\
	& = f(w^s_k) + \mathbb{E}\left[<\tilde{\nabla}_{s,k}, w^s_{k+1} - w^s_k>\right] + \dfrac{L\beta}{2} \| w^s_{k+1} - w^s_k \|^2\\
	& = f(w^s_k) + <\nabla f(w^s_k) + \dfrac{m-1}{m}\nabla f(\tilde{w}^{s-1}), w^s_{k+1} - w^s_k> + \dfrac{L\beta}{2} \| w^s_{k+1} - w^s_k \|^2\\
	& = f(w^s_k) + <\nabla f(w^s_k), w^s_{k+1} - w^s_k> + \dfrac{L\beta}{2} \| w^s_{k+1} - w^s_k \|^2 + \dfrac{m-1}{m}<\nabla f(\tilde{w}^{s-1}), w^s_{k+1} - w^s_k>\\
	& \le f(w^s_k) + <\nabla f(w^s_k), w^{*} - w^s_k> + \dfrac{L\beta}{2} \| w^{*} - w^s_k \|^2  - \dfrac{L\beta}{2} \| w^{*} - w^s_{k+1}\|^2\\& + \dfrac{m-1}{m}<\nabla f(\tilde{w}^{s-1}), w^s_{k+1} - w^s_k>\\
	& = f(w^s_k) + <\nabla f(w^s_k), w^{*} - w^s_k> + \dfrac{L\beta}{2} \| w^{*} - w^s_k \|^2  - \dfrac{L\beta}{2} \| w^{*} - w^s_{k+1}\|^2\\ & + \dfrac{m-1}{m}\left[<\nabla f(\tilde{w}^{s-1}), w^s_{k+1} - w^{*}>- <\nabla f(\tilde{w}^{s-1}), w^s_{k} - w^{*}>\right]\\
	& \le f(w^{*}) + \dfrac{L\beta}{2} \left[ \| w^{*} - w^s_k \|^2  -  \| w^{*} - w^s_{k+1}\|^2\right]\\ & + \dfrac{m-1}{m}\left[\dfrac{1}{2\delta}\|\nabla f(\tilde{w}^{s-1})\|^2 + \dfrac{\delta}{2} \|w^s_{k+1} - w^{*}\|^2 - \left[ \dfrac{1}{2\delta}\|\nabla f(\tilde{w}^{s-1})\|^2 +\dfrac{\delta}{2}  \|w^s_{k} - w^{*}\|^2\right]\right]\\
	& = f(w^{*}) + \dfrac{L\beta}{2} \left[ \| w^{*} - w^s_k \|^2  -  \| w^{*} - w^s_{k+1}\|^2\right] + \dfrac{\delta(m-1)}{2m}\left[\|w^s_{k+1} - w^{*}\|^2 - \|w^s_{k} - w^{*}\|^2\right],\\
	& = f(w^{*}) + \left(\dfrac{L\beta}{2} - \dfrac{\delta(m-1)}{2m}\right) \left[ \| w^{*} - w^s_k \|^2  -  \| w^{*} - w^s_{k+1}\|^2 \right],\\
	& = f(w^{*}),
	\end{aligned}
	\end{equation}
	second equality follows from, $\mathbb{E}\left[\tilde{\nabla}_{s,k}\right] = \nabla f(w^s_k) + \dfrac{m-1}{m}\nabla f(\tilde{w}^{s-1})$, first inequality follows from Lemma~\ref{lemma_3pt}, second inequality follows from the convexity, i.e., $ f(w^{*}) \ge f(w^s_k) + <\nabla f(w^s_k), w^{*} - w^s_k>$ and Young's inequality, i.e., $x^Ty \le 1/(2\delta) \|x\|^2 + \delta/2 \|y\|^2$ for $\delta > 0$, and last equality follows by choosing $\delta = \dfrac{mL\beta}{(m-1)}$.
	\begin{equation}
	\label{eq_proof_nsc_s3}
	\begin{aligned}
	& \text{and},\; \mathbb{E} \left[  <\nabla f(w^s_k) - \tilde{\nabla}_{s,k}, w^s_{k+1} - w^s_k> - \dfrac{L(\beta-1)}{2} \| w^s_{k+1} - w^s_k \|^2 \right]\\
	& \le \mathbb{E} \left[ \dfrac{1}{2L(\beta-1)} \|\nabla f(w^s_k) - \tilde{\nabla}_{s,k}\|^2 + \dfrac{L(\beta-1)}{2} \| w^s_{k+1} - w^s_k \|^2 - \dfrac{L(\beta-1)}{2} \| w^s_{k+1} - w^s_k \|^2 \right]\\
	& = \dfrac{1}{2L(\beta-1)} \mathbb{E} \left[\| \tilde{\nabla}_{s,k} - \nabla f(w^s_k)\|^2 \right]\\
	& \le \dfrac{1}{2L(\beta-1)}  \left[8L\alpha(b) \left[ f(w^s_k) - f^{*} \right] + \dfrac{8L\left(\alpha(b)m^2+(m-1)^2\right)}{m^2} \left[ f(\tilde{w}^{s-1}) - f^{*} \right] + R^{'}\right]\\
	& = \dfrac{4\alpha(b)}{(\beta-1)}   \left[ f(w^s_k) - f^{*} \right] + \dfrac{4\left(\alpha(b)m^2+(m-1)^2\right)}{m^2(\beta-1)} \left[ f(\tilde{w}^{s-1}) - f^{*} \right] + R^{''}
	\end{aligned}
	\end{equation}
	first inequality follows from Young's inequality, second inequality follows from Lemma~\ref{lemma_variance_bound_smooth} and $R^{''} = R^{'}/(2L(\beta-1))$.\\
	Now, substituting the values into \ref{eq_proof_nsc_s1} from inequalities \ref{eq_proof_nsc_s2} and \ref{eq_proof_nsc_s3}, and taking expectation w.r.t. mini-batches, we have
	\begin{equation}
	\label{eq_proof_nsc_s4}
	\begin{aligned}
	&\mathbb{E} \left[ f(w^s_{k+1})\right] \le f(w^{*}) + \dfrac{4\alpha(b)}{(\beta-1)}   \left[ f(w^s_k) - f^{*} \right] + \dfrac{4\left(\alpha(b)m^2+(m-1)^2\right)}{m^2(\beta-1)} \left[ f(\tilde{w}^{s-1}) - f^{*} \right] + R^{''}\\
	&\mathbb{E} \left[ f(w^s_{k+1})-f(w^{*})\right] \le \dfrac{4\alpha(b)}{(\beta-1)}   \left[ f(w^s_k) - f^{*} \right] + \dfrac{4\left(\alpha(b)m^2+(m-1)^2\right)}{m^2(\beta-1)} \left[ f(\tilde{w}^{s-1}) - f^{*} \right] + R^{''}
	\end{aligned}
	\end{equation}
	Taking sum over $k=0,1,...,(m-1)$ and dividing by $m$, we have
	\begin{equation}
	\label{eq_proof_nsc_s5}
	\begin{aligned}
	& \dfrac{1}{m}\sum_{k=0}^{m-1}\mathbb{E} \left[ f(w^s_{k+1}) - f^{*}\right] \\& \le \dfrac{1}{m}\sum_{k=0}^{m-1} \left[ \dfrac{4\alpha(b)}{(\beta-1)}   \left[ f(w^s_k) - f^{*} \right] + \dfrac{4\left(\alpha(b)m^2+(m-1)^2\right)}{m^2(\beta-1)} \left[ f(\tilde{w}^{s-1}) - f^{*} \right] + R^{''}\right]\\
	& \dfrac{1}{m}\sum_{k=1}^{m}\mathbb{E} \left[ f(w^s_{k}) - f^{*}\right] \\ & \le \dfrac{4\alpha(b)}{(\beta-1)} \dfrac{1}{m}\sum_{k=1}^{m} \left[ f(w^s_k) - f^{*} \right] + \dfrac{4\alpha(b)}{(\beta-1)} \dfrac{1}{m} \left[ f(w^s_0) - f^{*} - \lbrace f(w^s_m) - f^{*} \rbrace\right]\\& +  \dfrac{4\left(\alpha(b)m^2+(m-1)^2\right)}{m^2(\beta-1)} \left[ f (\tilde{w}^{s-1}) - f^{*} \right]+  R^{''}
	\end{aligned}
	\end{equation}
	Subtracting $\dfrac{4\alpha(b)}{(\beta-1)} \dfrac{1}{m}\sum_{k=1}^{m} \left[ f(w^s_k) - f^{*} \right] $ from both sides, we have
	\begin{equation}
	\label{eq_proof_nsc_s6}
	\begin{aligned}
	& \left(1-\dfrac{4\alpha(b)}{(\beta-1)}\right)\dfrac{1}{m}\sum_{k=1}^{m}\mathbb{E} \left[ f(w^s_{k}) - f^{*}\right] \\ & \le  \dfrac{4\alpha(b)}{(\beta-1)} \dfrac{1}{m} \left[ f(w^s_0) - f^{*} - \lbrace f(w^s_m) - f^{*} \rbrace\right] +  \dfrac{4\left(\alpha(b)m^2+(m-1)^2\right)}{m^2(\beta-1)} \left[ f (\tilde{w}^{s-1}) - f^{*} \right]+  R^{''}
	\end{aligned}
	\end{equation}
	
	Since $f(w^s_m) - f^{*} \ge 0$ so dropping this term and using Assumption~\ref{assump_3shang}, we have
	\begin{equation}
	\label{eq_proof_nsc_s7}
	\begin{aligned}
	& \left(1-\dfrac{4\alpha(b)}{(\beta-1)}\right)\dfrac{1}{m}\sum_{k=1}^{m}\mathbb{E} \left[ f(w^s_{k}) - f^{*}\right] \\ & \le  \dfrac{4\alpha(b)}{(\beta-1)} \dfrac{1}{m} \left[ f(w^s_0) - f^{*} \right] +  \dfrac{4\left(\alpha(b)m^2+(m-1)^2\right)}{m^2(\beta-1)} \left[ f (\tilde{w}^{s-1}) - f^{*} \right]+  R^{''}\\
	& \le  \dfrac{4\alpha(b)}{(\beta-1)} \dfrac{1}{m} \left[ c\left(f (\tilde{w}^{s-1}) - f^{*}\right)  \right] +  \dfrac{4\left(\alpha(b)m^2+(m-1)^2\right)}{m^2(\beta-1)} \left[ f (\tilde{w}^{s-1}) - f^{*} \right]+  R^{''}\\
	& =  \left[ \dfrac{4\alpha(b)}{(\beta-1)}\dfrac{c}{m} +\dfrac{4\left(\alpha(b)m^2+(m-1)^2\right)}{m^2(\beta-1)} \right] \left[ \left(f (\tilde{w}^{s-1}) - f^{*}\right)  \right]+  R^{''},\\
	\end{aligned}
	\end{equation}
	Dividing both sides by $\left(1-\dfrac{4\alpha(b)}{(\beta-1)}\right)$, and since $\tilde{w}^{s} = 1/m \sum_{k=1}^{m}w^s_k$ so by convexity, $f(\tilde{w}^{s}) \le 1/m \sum_{k=1}^{m} f(w^s_k)$, we have
	\begin{equation}
	\label{eq_proof_nsc_s8}
	\begin{aligned}
	\mathbb{E} \left[ f(\tilde{w}^{s}) - f^{*}\right] \le  \left[ \dfrac{4\alpha(b)}{(\beta-1-4\alpha(b))}\dfrac{c}{m} +\dfrac{4\left(\alpha(b)m^2+(m-1)^2\right)}{m^2(\beta-1-4\alpha(b))} \right] \left[ f (\tilde{w}^{s-1}) - f^{*}  \right]+  R^{'''},\\
	\end{aligned}
	\end{equation}
	where $R^{'''} = R^{''} (\beta-1)/\left(\beta-1-4\alpha(b)\right)$. Now, applying this inequality recursively, we have
	\begin{equation}
	\label{eq_proof_nsc_s9}
	\begin{aligned}
	\mathbb{E} \left[ f(\tilde{w}^{s}) - f^{*}\right] \le  C^s \left[ f (\tilde{w}^{0}) - f^{*}  \right] + R^{''''},\\
	\end{aligned}
	\end{equation}
	inequality follows for $R^{''''}= R^{'''}/(1-C)$, since $\sum_{i=0}^{k} r^i \le \sum_{i=0}^{\infty} r^i = \dfrac{1}{1-r}, \quad \|r\|<1$ and $C = \left[ \dfrac{4\alpha(b)}{(\beta-1-4\alpha(b))}\dfrac{c}{m} +\dfrac{4\left(\alpha(b)m^2+(m-1)^2\right)}{m^2(\beta-1-4\alpha(b))} \right]$. For certain choice of $\beta$, one can easily prove that $C<1$. This proves linear convergence with some initial error.
	
\end{proof}

\textbf{Proof of Theorem~2} (Strongly convex and smooth problem with SAAG-IV)\\
\begin{proof}
	By smoothness, we have,
	\begin{equation}
	\label{eq_proof_sc_s1}
	\begin{aligned}
	f(w^s_{k+1}) & \le f(w^s_k) + <\nabla f(w^s_k), w^s_{k+1} - w^s_k> + \dfrac{L}{2} \| w^s_{k+1} - w^s_k \|^2\\
	& = f(w^s_k) + <\nabla f(w^s_k), w^s_{k+1} - w^s_k> + \dfrac{L\beta}{2} \| w^s_{k+1} - w^s_k \|^2 - \dfrac{L(\beta-1}{2} \| w^s_{k+1} - w^s_k \|^2\\
	& = f(w^s_k) + <\tilde{\nabla}_{s,k}, w^s_{k+1} - w^s_k> + \dfrac{L\beta}{2} \| w^s_{k+1} - w^s_k \|^2 + <\nabla f(w^s_k) - \tilde{\nabla}_{s,k}, w^s_{k+1} - w^s_k>\\& - \dfrac{L(\beta-1}{2} \| w^s_{k+1} - w^s_k \|^2
	\end{aligned}
	\end{equation}
	where $\beta$ is appropriately chosen positive value. Now, separately simplifying the terms, we have
	\begin{equation}
	\label{eq_proof_sc_s2}
	\begin{aligned}
	&\mathbb{E}\left[ f(w^s_k) + <\tilde{\nabla}_{s,k}, w^s_{k+1} - w^s_k> + \dfrac{L\beta}{2} \| w^s_{k+1} - w^s_k \|^2 \right]\\
	& = f(w^s_k) + \mathbb{E}\left[<\tilde{\nabla}_{s,k}, w^s_{k+1} - w^s_k>\right] + \dfrac{L\beta}{2} \| w^s_{k+1} - w^s_k \|^2\\
	& = f(w^s_k) + <\nabla f(w^s_k) + \dfrac{m-1}{m}\nabla f(\tilde{w}^{s-1}), w^s_{k+1} - w^s_k> + \dfrac{L\beta}{2} \| w^s_{k+1} - w^s_k \|^2\\
	& = f(w^s_k) + <\nabla f(w^s_k), w^s_{k+1} - w^s_k> + \dfrac{L\beta}{2} \| w^s_{k+1} - w^s_k \|^2 + \dfrac{m-1}{m}<\nabla f(\tilde{w}^{s-1}), w^s_{k+1} - w^s_k>\\
	& \le f(w^s_k) + <\nabla f(w^s_k), w^{*} - w^s_k> + \dfrac{L\beta}{2} \| w^{*} - w^s_k \|^2  - \dfrac{L\beta}{2} \| w^{*} - w^s_{k+1}\|^2\\& + \dfrac{m-1}{m}<\nabla f(\tilde{w}^{s-1}), w^s_{k+1} - w^s_k>\\
	& = f(w^s_k) + <\nabla f(w^s_k), w^{*} - w^s_k> + \dfrac{L\beta}{2} \| w^{*} - w^s_k \|^2  - \dfrac{L\beta}{2} \| w^{*} - w^s_{k+1}\|^2\\ & + \dfrac{m-1}{m}\left[<\nabla f(\tilde{w}^{s-1}), w^s_{k+1} - \tilde{w}^{s-1}>- <\nabla f(\tilde{w}^{s-1}), w^s_{k} - \tilde{w}^{s-1}>\right]\\
	& \le f(w^{*}) + \dfrac{L\beta}{2} \left[ \| w^{*} - w^s_k \|^2  -  \| w^{*} - w^s_{k+1}\|^2\right]\\ & + \dfrac{m-1}{m}\left[ f(w^s_{k+1}) - f(\tilde{w}^{s-1}) - \left( f(w^s_{k}) - f(\tilde{w}^{s-1})\right)\right]\\
	& = f(w^{*}) + \dfrac{L\beta}{2} \left[ \| w^{*} - w^s_k \|^2  -  \| w^{*} - w^s_{k+1}\|^2\right] + \dfrac{m-1}{m}\left[ f(w^s_{k+1}) - f(w^s_{k})\right],\\
	\end{aligned}
	\end{equation}
	second equality follows from, $\mathbb{E}\left[\tilde{\nabla}_{s,k}\right] = \nabla f(w^s_k) + \dfrac{m-1}{m}\nabla f(\tilde{w}^{s-1})$, first inequality follows from Lemma~\ref{lemma_3pt} and second inequality follows from the convexity, i.e., $ f(x) \ge f(y) + <\nabla f(y), x - y>$.
	\begin{equation}
	\label{eq_proof_sc_s3}
	\begin{aligned}
	& \text{and},\; \mathbb{E} \left[  <\nabla f(w^s_k) - \tilde{\nabla}_{s,k}, w^s_{k+1} - w^s_k> - \dfrac{L(\beta-1)}{2} \| w^s_{k+1} - w^s_k \|^2 \right]\\
	& \le \mathbb{E} \left[ \dfrac{1}{2L(\beta-1)} \|\nabla f(w^s_k) - \tilde{\nabla}_{s,k}\|^2 + \dfrac{L(\beta-1)}{2} \| w^s_{k+1} - w^s_k \|^2 - \dfrac{L(\beta-1)}{2} \| w^s_{k+1} - w^s_k \|^2 \right]\\
	& \le \dfrac{1}{2L(\beta-1)}  \left[8L\alpha(b) \left[ f(w^s_k) - f^{*} \right] + \dfrac{8L\left(\alpha(b)m^2+(m-1)^2\right)}{m^2} \left[ f(\tilde{w}^{s-1}) - f^{*} \right] + R^{'}\right]\\
	& = \dfrac{4\alpha(b)}{(\beta-1)}   \left[ f(w^s_k) - f^{*} \right] + \dfrac{4\left(\alpha(b)m^2+(m-1)^2\right)}{m^2(\beta-1)} \left[ f(\tilde{w}^{s-1}) - f^{*} \right] + R^{''}
	\end{aligned}
	\end{equation}
	first inequality follows from Young's inequality and second inequality follows from Lemma~\ref{lemma_variance_bound_smooth} and $R^{''} = R^{'}/(2L(\beta-1))$.\\
	Now, substituting the values into \ref{eq_proof_sc_s1} from inequalities \ref{eq_proof_sc_s2} and \ref{eq_proof_sc_s3}, and taking expectation w.r.t. mini-batches, we have
	\begin{equation}
	\label{eq_proof_sc_s4}
	\begin{aligned}
	&\mathbb{E} \left[ f(w^s_{k+1})\right] \\& \le f(w^{*}) + \dfrac{L\beta}{2} \left[ \| w^{*} - w^s_k \|^2  -  \| w^{*} - w^s_{k+1}\|^2\right] + \dfrac{m-1}{m}\left[ f(w^s_{k+1}) - f(w^s_{k})\right]\\ &+ \dfrac{4\alpha(b)}{(\beta-1)}   \left[ f(w^s_k) - f^{*} \right] + \dfrac{4\left(\alpha(b)m^2+(m-1)^2\right)}{m^2(\beta-1)} \left[ f(\tilde{w}^{s-1}) - f^{*} \right] + R^{''},\\
	&\mathbb{E} \left[ f(w^s_{k+1})-f(w^{*})\right] \\& = \dfrac{L\beta}{2} \left[ \| w^{*} - w^s_k \|^2  -  \| w^{*} - w^s_{k+1}\|^2\right] + \dfrac{m-1}{m}\left[ f(w^s_{k+1}) - f(w^s_{k})\right]\\ &+ \dfrac{4\alpha(b)}{(\beta-1)}   \left[ f(w^s_k) - f^{*} \right] + \dfrac{4\left(\alpha(b)m^2+(m-1)^2\right)}{m^2(\beta-1)} \left[ f(\tilde{w}^{s-1}) - f^{*} \right] + R^{''}
	\end{aligned}
	\end{equation}
	Taking sum over $k=0,1,...,(m-1)$ and dividing by $m$, we have
	\begin{equation}
	\label{eq_proof_sc_s5}
	\begin{aligned}
	&\dfrac{1}{m}\sum_{k=0}^{m-1}\mathbb{E} \left[ f(w^s_{k+1})-f(w^{*})\right] \\& \le \dfrac{1}{m}\sum_{k=0}^{m-1}\left\lbrace \dfrac{L\beta}{2} \left[ \| w^{*} - w^s_k \|^2  -  \| w^{*} - w^s_{k+1}\|^2\right] + \dfrac{m-1}{m}\left[ f(w^s_{k+1}) - f(w^s_{k})\right]\right\rbrace\\ &+ \dfrac{1}{m}\sum_{k=0}^{m-1}\left\lbrace\dfrac{4\alpha(b)}{(\beta-1)}   \left[ f(w^s_k) - f^{*} \right] + \dfrac{4\left(\alpha(b)m^2+(m-1)^2\right)}{m^2(\beta-1)} \left[ f(\tilde{w}^{s-1}) - f^{*} \right] + R^{''}\right\rbrace\\
	&\dfrac{1}{m}\sum_{k=1}^{m}\mathbb{E} \left[ f(w^s_{k})-f(w^{*})\right] \\& \le \dfrac{L\beta}{2m} \left[ \| w^{*} - w^s_0 \|^2  -  \| w^{*} - w^s_{m}\|^2\right] + \dfrac{m-1}{m^2}\left[ f(w^s_{m}) - f(w^s_{0})\right]\\ &+ \dfrac{4\alpha(b)}{(\beta-1)}\dfrac{1}{m}\left\lbrace \sum_{k=1}^{m}\left[ f(w^s_k) - f^{*} \right] + f(w^s_0) - f^{*}  - \left(f(w^s_m) - f^{*} \right)\right\rbrace\\& + \dfrac{4\left(\alpha(b)m^2+(m-1)^2\right)}{m^2(\beta-1)} \left[ f(\tilde{w}^{s-1}) - f^{*} \right] + R^{''}\\
	\end{aligned}
	\end{equation}
	Subtracting $\dfrac{4\alpha(b)}{(\beta-1)}\dfrac{1}{m}\sum_{k=1}^{m}\left[ f(w^s_k) - f^{*} \right]$ from both sides, we have
	\begin{equation}
	\label{eq_proof_sc_s6}
	\begin{aligned}
	& \left(1-\dfrac{4\alpha(b)}{(\beta-1)}\right)\dfrac{1}{m}\sum_{k=1}^{m}\mathbb{E} \left[ f(w^s_{k}) - f^{*}\right] \\ & \le  \dfrac{L\beta}{2m} \left[ \| w^{*} - w^s_0 \|^2  -  \| w^{*} - w^s_{m}\|^2\right] - \left( \dfrac{m-1}{m^2} - \dfrac{4\alpha(b)}{(\beta-1)} \dfrac{1}{m}\right) \left[ f(w^s_0) -f(w^s_m) \right] \\& + \dfrac{4\left(\alpha(b)m^2+(m-1)^2\right)}{m^2(\beta-1)} \left[ f(\tilde{w}^{s-1}) - f^{*} \right] + R^{''}\\
	& \le  \dfrac{L\beta}{2m} \| w^{*} - w^s_0 \|^2  - \left( \dfrac{m-1}{m^2} - \dfrac{4\alpha(b)}{(\beta-1)} \dfrac{1}{m}\right) \left[ f(w^s_0) -f^{*} - \lbrace f(w^s_m) - f^{*}\rbrace \right] \\& + \dfrac{4\left(\alpha(b)m^2+(m-1)^2\right)}{m^2(\beta-1)} \left[ f(\tilde{w}^{s-1}) - f^{*} \right] + R^{''}\\
	& \le  \dfrac{L\beta}{2m} \dfrac{2}{\mu}\left( f(w^s_0)- f^{*} \right)   - \left( \dfrac{m-1}{m^2} - \dfrac{4\alpha(b)}{(\beta-1)} \dfrac{1}{m}\right) \left[ c\left[ f(\tilde{w}^{s-1}) - f^{*} \right] - c\left[ f(\tilde{w}^{s}) - f^{*} \right] \right] \\& + \dfrac{4\left(\alpha(b)m^2+(m-1)^2\right)}{m^2(\beta-1)} \left[ f(\tilde{w}^{s-1}) - f^{*} \right] + R^{''}\\
	& \le  \dfrac{L\beta}{2m} \dfrac{2}{\mu}c\left[ f(\tilde{w}^{s-1}) - f^{*} \right]   - \left( \dfrac{m-1}{m^2} - \dfrac{4\alpha(b)}{(\beta-1)} \dfrac{1}{m}\right) \left[ c\left[ f(\tilde{w}^{s-1}) - f^{*} \right] - c\left[ f(\tilde{w}^{s}) - f^{*} \right] \right] \\& + \dfrac{4\left(\alpha(b)m^2+(m-1)^2\right)}{m^2(\beta-1)} \left[ f(\tilde{w}^{s-1}) - f^{*} \right] + R^{''}
	\end{aligned}
	\end{equation}
	second inequality follows by dropping, $\| w^{*} - w^s_{m}\|^2 > 0$, third inequality follows from the strong convexity, i.e., $\|  w^s_0 - w^{*}\|^2 \le 2/ \mu\left( f(w^s_0)- f^{*} \right) $ and application of Assumption~\ref{assump_3shang} twice, and fourth inequality follows from Assumption~\ref{assump_3shang}.\\
	Since $\tilde{w}^{s} = 1/m \sum_{k=1}^{m}w^s_k$ so by convexity using, $f(\tilde{w}^{s}) \le 1/m \sum_{k=1}^{m} f(w^s_k)$, we have
	\begin{equation}
	\label{eq_proof_sc_s7}
	\begin{aligned}
	& \left(1-\dfrac{4\alpha(b)}{(\beta-1)}\right)\mathbb{E} \left[ f(\tilde{w}^{s}) - f^{*}\right]\\ &\le  \dfrac{L\beta}{2m} \dfrac{2}{\mu}c\left[ f(\tilde{w}^{s-1}) - f^{*} \right]   - \left( \dfrac{m-1}{m^2} - \dfrac{4\alpha(b)}{(\beta-1)} \dfrac{1}{m}\right) \left[ c\left[ f(\tilde{w}^{s-1}) - f^{*} \right] - c\left[ f(\tilde{w}^{s}) - f^{*} \right] \right] \\& + \dfrac{4\left(\alpha(b)m^2+(m-1)^2\right)}{m^2(\beta-1)} \left[ f(\tilde{w}^{s-1}) - f^{*} \right] + R^{''}
	\end{aligned}
	\end{equation}
	Subtracting, $ c\left( \dfrac{m-1}{m^2} - \dfrac{4\alpha(b)}{(\beta-1)} \dfrac{1}{m}\right) \mathbb{E} \left[ f(\tilde{w}^{s}) - f^{*}\right]$ both sides, we have
	\begin{equation}
	\label{eq_proof_sc_s8}
	\begin{aligned}
	&\left(1-\dfrac{4\alpha(b)}{(\beta-1)} - c\left( \dfrac{m-1}{m^2} - \dfrac{4\alpha(b)}{(\beta-1)} \dfrac{1}{m}\right)\right) \mathbb{E} \left[ f(\tilde{w}^{s}) - f^{*}\right]\\ & \le \left[\dfrac{cL\beta}{m\mu} + \dfrac{4\left(\alpha(b)m^2+(m-1)^2\right)}{m^2(\beta-1)}   -\dfrac{c(m-1)}{m^2} + \dfrac{4\alpha(b)}{(\beta-1)}\right] \left[ f(\tilde{w}^{s-1}) -f^{*} \right]+  R^{''}\\
	\end{aligned}
	\end{equation}
	Dividing both sides by $ \left(1-\dfrac{4\alpha(b)}{(\beta-1)} - c\left( \dfrac{m-1}{m^2} - \dfrac{4\alpha(b)}{(\beta-1)} \dfrac{1}{m}\right)\right)$, we have
	\begin{equation}
	\label{eq_proof_sc_s9}
	\begin{aligned}
	\mathbb{E} \left[ f(\tilde{w}^{s}) - f^{*}\right]  \le C\left[ f(\tilde{w}^{s-1}) -f^{*} \right]+  R^{'''}\\
	\end{aligned}
	\end{equation}
	where\\ $C = \left[\dfrac{cL\beta}{m\mu} + \dfrac{4\left(\alpha(b)m^2+(m-1)^2\right)}{m^2(\beta-1)}   -\dfrac{c(m-1)}{m^2} + \dfrac{4\alpha(b)}{(\beta-1)}\right] \left(1-\dfrac{4\alpha(b)}{(\beta-1)} - c\left( \dfrac{m-1}{m^2} - \dfrac{4\alpha(b)}{(\beta-1)} \dfrac{1}{m}\right)\right)^{-1}$ and $R^{'''}= R^{''}\left(1-\dfrac{4\alpha(b)}{(\beta-1)} - c\left( \dfrac{m-1}{m^2} - \dfrac{4\alpha(b)}{(\beta-1)} \dfrac{1}{m}\right)\right)^{-1}$.\\
	Now, recursively applying the inequality, we have
	\begin{equation}
	\label{eq_proof_sc_s10}
	\begin{aligned}
	\mathbb{E} \left[ f(\tilde{w}^{s}) - f^{*}\right]  \le C^{s}\left[ f(\tilde{w}^{0}) -f^{*} \right]+  R^{''''},
	\end{aligned}
	\end{equation}
	inequality follows for $R^{''''}= R^{'''}/(1-C)$, since $\sum_{i=0}^{k} r^i \le \sum_{i=0}^{\infty} r^i = \dfrac{1}{1-r}, \quad \|r\|<1$. For certain choice of $\beta$, one can easily prove that $C<1$. This proves linear convergence with some initial error.
\end{proof}

\textbf{Proof of Theorem 3} (Non-strongly convex and non-smooth problem with SAAG-IV)\\
\begin{proof}
	By smoothness, we have,
	\begin{equation}
	\begin{aligned}
	f(w^s_{k+1}) & \le f(w^s_k) + <\nabla f(w^s_k), w^s_{k+1} - w^s_k> + \dfrac{L}{2} \| w^s_{k+1} - w^s_k \|^2
	\end{aligned}
	\end{equation}
	\begin{equation}
	\label{eq_proof_nsc_ns1}
	\begin{aligned}
	\text{Now, } & F(w^s_{k+1}) = f(w^s_{k+1}) + g(w^s_{k+1})\\
	& \le f(w^s_k) + g(w^s_{k+1}) + <\nabla f(w^s_k), w^s_{k+1} - w^s_k> + \dfrac{L}{2} \| w^s_{k+1} - w^s_k \|^2\\
	& = f(w^s_k) + g(w^s_{k+1}) + <\nabla f(w^s_k), w^s_{k+1} - w^s_k> + \dfrac{L\beta}{2} \| w^s_{k+1} - w^s_k \|^2 - \dfrac{L(\beta-1}{2} \| w^s_{k+1} - w^s_k \|^2\\
	& = f(w^s_k) + g(w^s_{k+1}) + <\tilde{\nabla}_{s,k}, w^s_{k+1} - w^s_k> + \dfrac{L\beta}{2} \| w^s_{k+1} - w^s_k \|^2 + <\nabla f(w^s_k) - \tilde{\nabla}_{s,k}, w^s_{k+1} - w^s_k>\\& - \dfrac{L(\beta-1}{2} \| w^s_{k+1} - w^s_k \|^2
	\end{aligned}
	\end{equation}
	where $\beta$ is appropriately chosen positive value. Now, separately simplifying the terms, we have
	\begin{equation}
	\label{eq_proof_nsc_ns2}
	\begin{aligned}
	&\mathbb{E}\left[ f(w^s_k) + g(w^s_{k+1}) + <\tilde{\nabla}_{s,k}, w^s_{k+1} - w^s_k> + \dfrac{L\beta}{2} \| w^s_{k+1} - w^s_k \|^2 \right]\\
	& = f(w^s_k) + g(w^s_{k+1}) + \mathbb{E}\left[<\tilde{\nabla}_{s,k}, w^s_{k+1} - w^s_k>\right] + \dfrac{L\beta}{2} \| w^s_{k+1} - w^s_k \|^2\\
	& = f(w^s_k) + g(w^s_{k+1}) + <\nabla f(w^s_k) + \dfrac{m-1}{m}\nabla f(\tilde{w}^{s-1}), w^s_{k+1} - w^s_k> + \dfrac{L\beta}{2} \| w^s_{k+1} - w^s_k \|^2\\
	& = f(w^s_k) + g(w^s_{k+1}) + <\nabla f(w^s_k), w^s_{k+1} - w^s_k> + \dfrac{L\beta}{2} \| w^s_{k+1} - w^s_k \|^2 + \dfrac{m-1}{m}<\nabla f(\tilde{w}^{s-1}), w^s_{k+1} - w^s_k>\\
	& \le f(w^s_k) + g(w^{*}) + <\nabla f(w^s_k), w^{*} - w^s_k> + \dfrac{L\beta}{2} \| w^{*} - w^s_k \|^2  - \dfrac{L\beta}{2} \| w^{*} - w^s_{k+1}\|^2\\& + \dfrac{m-1}{m}<\nabla f(\tilde{w}^{s-1}), w^s_{k+1} - w^s_k>\\
	& = f(w^s_k) + g(w^{*}) + <\nabla f(w^s_k), w^{*} - w^s_k> + \dfrac{L\beta}{2} \| w^{*} - w^s_k \|^2  - \dfrac{L\beta}{2} \| w^{*} - w^s_{k+1}\|^2\\ & + \dfrac{m-1}{m}\left[<\nabla f(\tilde{w}^{s-1}), w^s_{k+1} - w^{*}>- <\nabla f(\tilde{w}^{s-1}), w^s_{k} - w^{*}>\right]\\
	& \le f(w^{*}) + g(w^{*}) + \dfrac{L\beta}{2} \left[ \| w^{*} - w^s_k \|^2  -  \| w^{*} - w^s_{k+1}\|^2\right]\\ & + \dfrac{m-1}{m}\left[\dfrac{1}{2\delta}\|\nabla f(\tilde{w}^{s-1})\|^2 + \dfrac{\delta}{2} \|w^s_{k+1} - w^{*}\|^2 - \left[ \dfrac{1}{2\delta}\|\nabla f(\tilde{w}^{s-1})\|^2 +\dfrac{\delta}{2}  \|w^s_{k} - w^{*}\|^2\right]\right]\\
	& = F(w^{*}) + \dfrac{L\beta}{2} \left[ \| w^{*} - w^s_k \|^2  -  \| w^{*} - w^s_{k+1}\|^2\right] + \dfrac{\delta(m-1)}{2m}\left[\|w^s_{k+1} - w^{*}\|^2 - \|w^s_{k} - w^{*}\|^2\right],\\
	& = F(w^{*}) + \left(\dfrac{L\beta}{2} - \dfrac{\delta(m-1)}{2m}\right) \left[ \| w^{*} - w^s_k \|^2  -  \| w^{*} - w^s_{k+1}\|^2 \right],\\
	& =F(w^{*}),
	\end{aligned}
	\end{equation}
	second equality follows from, $\mathbb{E}\left[\tilde{\nabla}_{s,k}\right] = \nabla f(w^s_k) + \dfrac{m-1}{m}\nabla f(\tilde{w}^{s-1})$, first inequality follows from Lemma~\ref{lemma_3pt}, second inequality follows from the convexity, i.e., $ f(w^{*}) \ge f(w^s_k) + <\nabla f(w^s_k), w^{*} - w^s_k>$ and Young's inequality, i.e., $x^Ty \le 1/(2\delta) \|x\|^2 + \delta/2 \|y\|^2$ for $\delta > 0$, and last equality follows by choosing $\delta = \dfrac{mL\beta}{(m-1)}$.
	\begin{equation}
	\label{eq_proof_nsc_ns3}
	\begin{aligned}
	& \text{And},\; \mathbb{E} \left[  <\nabla f(w^s_k) - \tilde{\nabla}_{s,k}, w^s_{k+1} - w^s_k> - \dfrac{L(\beta-1)}{2} \| w^s_{k+1} - w^s_k \|^2 \right]\\
	& \le \mathbb{E} \left[ \dfrac{1}{2L(\beta-1)} \|\nabla f(w^s_k) - \tilde{\nabla}_{s,k}\|^2 + \dfrac{L(\beta-1)}{2} \| w^s_{k+1} - w^s_k \|^2 - \dfrac{L(\beta-1)}{2} \| w^s_{k+1} - w^s_k \|^2 \right]\\
	& = \dfrac{1}{2L(\beta-1)} \mathbb{E} \left[\| \tilde{\nabla}_{s,k} - \nabla f(w^s_k)\|^2 \right]\\
	& \le \dfrac{1}{2L(\beta-1)}\left[8L\alpha(b) \left[ F(w^s_k) - F^{*} \right] + \dfrac{8L\left(\alpha(b)m^2+(m-1)^2\right)}{m^2} \left[ F(\tilde{w}^{s-1}) - F^{*} \right] + R^{'}\right]\\
	& = \dfrac{4\alpha(b) }{(\beta-1)}\left[ F(w^s_k) - F^{*} \right] + \dfrac{4\left(\alpha(b)m^2+(m-1)^2\right)}{m^2(\beta-1)} \left[ F(\tilde{w}^{s-1}) - F^{*} \right] + R^{''}
	\end{aligned}
	\end{equation}
	first inequality follows from Young's inequality and second inequality follows from Lemma~\ref{lemma_variance_bound_nonsmooth} and $R^{''} = R^{'}/(2L(\beta-1))$. Now, substituting the values into \ref{eq_proof_nsc_ns1} from inequalities \ref{eq_proof_nsc_ns2} and \ref{eq_proof_nsc_ns3}, and taking expectation w.r.t. mini-batches, we have
	\begin{equation}
	\label{eq_proof_nsc_ns4}
	\begin{aligned}
	\mathbb{E} \left[ F(w^s_{k+1})\right] \le F(w^{*}) + \dfrac{4\alpha(b) }{(\beta-1)}\left[ F(w^s_k) - F^{*} \right] + \dfrac{4\left(\alpha(b)m^2+(m-1)^2\right)}{m^2(\beta-1)} \left[ F(\tilde{w}^{s-1}) - F^{*} \right] + R^{''}\\
	\mathbb{E} \left[ F(w^s_{k+1})-F(w^{*})\right] \le  \dfrac{4\alpha(b) }{(\beta-1)}\left[ F(w^s_k) - F^{*} \right] + \dfrac{4\left(\alpha(b)m^2+(m-1)^2\right)}{m^2(\beta-1)} \left[ F(\tilde{w}^{s-1}) - F^{*} \right] + R^{''}
	\end{aligned}
	\end{equation}
	Taking sum over $k=0,1,...,(m-1)$ and dividing by $m$, we have
	\begin{equation}
	\label{eq_proof_nsc_ns5}
	\begin{aligned}
	& \dfrac{1}{m}\sum_{k=0}^{m-1}\mathbb{E} \left[ F(w^s_{k+1})-F(w^{*})\right] \\& \le \dfrac{4\alpha(b)}{(\beta-1)} \dfrac{1}{m}\sum_{k=0}^{m-1} \left[ F(w^s_k) - F(w^{*}) \right] + \dfrac{4\left(\alpha(b)m^2+(m-1)^2\right)}{m^2(\beta-1)} \left[ F(\tilde{w}^{s-1}) - F(w^{*}) \right]  + R^{''}\\
	& \dfrac{1}{m}\sum_{k=1}^{m}\mathbb{E} \left[ F(w^s_{k}) - F(w^{*})\right] \\ & \le \dfrac{4\alpha(b)}{(\beta-1)} \dfrac{1}{m} \left\lbrace \sum_{k=1}^{m} \left[ F(w^s_k) - F(w^{*})  \right] + F(w^s_0) - F(w^{*}) - \lbrace F(w^s_m) - F(w^{*}) \rbrace \right\rbrace\\ & + \dfrac{4\left(\alpha(b)m^2+(m-1)^2\right)}{m^2(\beta-1)} \left[ F(\tilde{w}^{s-1}) - F(w^{*}) \right]  + R^{''}
	\end{aligned}
	\end{equation}
	Subtracting $\dfrac{4\alpha(b)}{(\beta-1)} \dfrac{1}{m}\sum_{k=1}^{m} \left[ F(w^s_{k}) - F(w^{*})\right] $ from both sides, we have
	\begin{equation}
	\label{eq_proof_nsc_ns6}
	\begin{aligned}
	& \left(1-\dfrac{4\alpha(b)}{(\beta-1)}\right)\dfrac{1}{m}\sum_{k=1}^{m}\mathbb{E} \left[ F(w^s_{k}) - F(w^{*})\right] \\ & \le \dfrac{4\alpha(b)}{(\beta-1)} \dfrac{1}{m} \left[ F(w^s_0) - F(w^{*}) - \lbrace F(w^s_m) - F(w^{*}) \rbrace\right]\\& + \dfrac{4\left(\alpha(b)m^2+(m-1)^2\right)}{m^2(\beta-1)} \left[ F(\tilde{w}^{s-1}) - F(w^{*}) \right]  + R^{''}
	\end{aligned}
	\end{equation}
	Since $F(w^s_m) - F(w^{*}) \ge 0$ so dropping this term and using Assumption~\ref{assump_3shang}, we have
	\begin{equation}
	\label{eq_proof_nsc_ns7}
	\begin{aligned}
	& \left(1-\dfrac{4\alpha(b)}{(\beta-1)}\right)\dfrac{1}{m}\sum_{k=1}^{m}\mathbb{E} \left[ F(w^s_{k}) - F(w^{*})\right] \\ & \le \dfrac{4\alpha(b)}{(\beta-1)} \dfrac{1}{m} \left[ F(w^s_0) - F(w^{*}) \right] + \dfrac{4\left(\alpha(b)m^2+(m-1)^2\right)}{m^2(\beta-1)} \left[ F(\tilde{w}^{s-1}) - F(w^{*}) \right]  + R^{''}\\
	& \le  \dfrac{4\alpha(b)}{(\beta-1)} \dfrac{1}{m} \left[ c\left(F(\tilde{w}^{s-1}) - F(w^{*})\right)  \right] + \dfrac{4\left(\alpha(b)m^2+(m-1)^2\right)}{m^2(\beta-1)} \left[ F(\tilde{w}^{s-1}) - F(w^{*}) \right]  + R^{''}\\
	& =  \left( \dfrac{4\alpha(b)}{(\beta-1)} \dfrac{c}{m}+ \dfrac{4\left(\alpha(b)m^2+(m-1)^2\right)}{m^2(\beta-1)} \right) \left[ \left(F(\tilde{w}^{s-1}) - F(w^{*})\right)  \right] + R^{''},\\
	\end{aligned}
	\end{equation}
	Dividing both sides by $\left(1-\dfrac{4\alpha(b)}{(\beta-1)}\right)$, and since $\tilde{w}^{s} = 1/m \sum_{k=1}^{m}w^s_k$ so by convexity, $F(\tilde{w}^{s}) \le 1/m \sum_{k=1}^{m} F(w^s_k)$, we have
	\begin{equation}
	\label{eq_proof_nsc_ns8}
	\begin{aligned}
	&\mathbb{E} \left[ F(\tilde{w}^{s}) - F^{*}\right] \le  \left( \dfrac{4\alpha(b)}{(\beta-1-4\alpha(b))} \dfrac{c}{m}+ \dfrac{4\left(\alpha(b)m^2+(m-1)^2\right)}{m^2(\beta-1-4\alpha(b))}\right)  \left[ F (\tilde{w}^{s-1}) - F(w^{*})  \right] + R^{'''},
	\end{aligned}
	\end{equation}
	where $ R^{'''}= R^{''} \left(1-\dfrac{4\alpha(b)}{(\beta-1)}\right)^{-1}$. Now, applying above inequality recursively, we have
	\begin{equation}
	\label{eq_proof_nsc_ns9}
	\begin{aligned}
	\mathbb{E} \left[ F(\tilde{w}^{s}) - F(w^{*})\right] \le  C^s \left[ F(\tilde{w}^{0}) - F(w^{*})  \right] + R^{''''},\\
	\end{aligned}
	\end{equation}
	inequality follows for $R^{''''}= R^{'''}/(1-C)$, since $\sum_{i=0}^{k} r^i \le \sum_{i=0}^{\infty} r^i = \dfrac{1}{1-r}, \quad \|r\|<1$ and $C = \left( \dfrac{4\alpha(b)}{(\beta-1-4\alpha(b))} \dfrac{c}{m}+ \dfrac{4\left(\alpha(b)m^2+(m-1)^2\right)}{m^2(\beta-1-4\alpha(b))}\right) $. For certain choice of $\beta$, one can easily prove that $C<1$. This proves linear convergence with some initial error.
\end{proof}

\textbf{Proof of Theorem 4} (Strongly convex and non-smooth problem with SAAG-IV)\\
\begin{proof}
	By smoothness, we have,
	\begin{equation*}
	\begin{aligned}
	f(w^s_{k+1}) & \le f(w^s_k) + <\nabla f(w^s_k), w^s_{k+1} - w^s_k> + \dfrac{L}{2} \| w^s_{k+1} - w^s_k \|^2\\
	\end{aligned}
	\end{equation*}
	Now,
	\begin{equation}
	\label{eq_proof_sc_ns1}
	\begin{aligned}
	& F(w^s_{k+1}) = f(w^s_{k+1}) + g(w^s_{k+1})\\& \le f(w^s_k) + g(w^s_{k+1}) + <\nabla f(w^s_k), w^s_{k+1} - w^s_k> + \dfrac{L}{2} \| w^s_{k+1} - w^s_k \|^2\\
	& = f(w^s_k) + g(w^s_{k+1}) + <\nabla f(w^s_k), w^s_{k+1} - w^s_k> + \dfrac{L\beta}{2} \| w^s_{k+1} - w^s_k \|^2 - \dfrac{L(\beta-1}{2} \| w^s_{k+1} - w^s_k \|^2\\
	& = f(w^s_k) + g(w^s_{k+1}) + <\tilde{\nabla}_{s,k}, w^s_{k+1} - w^s_k> + \dfrac{L\beta}{2} \| w^s_{k+1} - w^s_k \|^2 + <\nabla f(w^s_k) - \tilde{\nabla}_{s,k}, w^s_{k+1} - w^s_k>\\& - \dfrac{L(\beta-1}{2} \| w^s_{k+1} - w^s_k \|^2
	\end{aligned}
	\end{equation}
	where $\beta$ is appropriately chosen positive value. Now, separately simplifying the terms, we have
	\begin{equation}
	\label{eq_proof_sc_ns2}
	\begin{aligned}
	&\mathbb{E}\left[ f(w^s_k) + g(w^s_{k+1}) + <\tilde{\nabla}_{s,k}, w^s_{k+1} - w^s_k> + \dfrac{L\beta}{2} \| w^s_{k+1} - w^s_k \|^2 \right]\\
	& = f(w^s_k) + g(w^s_{k+1}) + \mathbb{E}\left[<\tilde{\nabla}_{s,k}, w^s_{k+1} - w^s_k>\right] + \dfrac{L\beta}{2} \| w^s_{k+1} - w^s_k \|^2\\
	& = f(w^s_k) + g(w^s_{k+1}) + <\nabla f(w^s_k) + \dfrac{m-1}{m}\nabla f(\tilde{w}^{s-1}), w^s_{k+1} - w^s_k> + \dfrac{L\beta}{2} \| w^s_{k+1} - w^s_k \|^2\\
	& = f(w^s_k) + g(w^s_{k+1}) + <\nabla f(w^s_k), w^s_{k+1} - w^s_k> + \dfrac{L\beta}{2} \| w^s_{k+1} - w^s_k \|^2 + \dfrac{m-1}{m}<\nabla f(\tilde{w}^{s-1}), w^s_{k+1} - w^s_k>\\
	& \le f(w^s_k) + g(w^{*}) + <\nabla f(w^s_k), w^{*} - w^s_k> + \dfrac{L\beta}{2} \| w^{*} - w^s_k \|^2  - \dfrac{L\beta}{2} \| w^{*} - w^s_{k+1}\|^2\\& + \dfrac{m-1}{m}<\nabla f(\tilde{w}^{s-1}), w^s_{k+1} - w^s_k>\\
	& = f(w^s_k) + g(w^{*}) + <\nabla f(w^s_k), w^{*} - w^s_k> + \dfrac{L\beta}{2} \| w^{*} - w^s_k \|^2  - \dfrac{L\beta}{2} \| w^{*} - w^s_{k+1}\|^2\\ & + \dfrac{m-1}{m}\left[<\nabla f(\tilde{w}^{s-1}), w^s_{k+1} - \tilde{w}^{s-1}>- <\nabla f(\tilde{w}^{s-1}), w^s_{k} - \tilde{w}^{s-1}>\right]\\
	& \le f(w^{*}) + g(w^{*}) + \dfrac{L\beta}{2} \left[ \| w^{*} - w^s_k \|^2  -  \| w^{*} - w^s_{k+1}\|^2\right]\\ & + \dfrac{m-1}{m}\left[ f(w^s_{k+1}) - f(\tilde{w}^{s-1}) - \left( f(w^s_{k}) - f(\tilde{w}^{s-1})\right)\right]\\
	& = f(w^{*}) + g(w^{*})  + \dfrac{L\beta}{2} \left[ \| w^{*} - w^s_k \|^2  -  \| w^{*} - w^s_{k+1}\|^2\right] + \dfrac{m-1}{m}\left[ f(w^s_{k+1}) - f(w^s_{k})\right],\\
	& = F(w^{*}) + \dfrac{L\beta}{2} \left[ \| w^{*} - w^s_k \|^2  -  \| w^{*} - w^s_{k+1}\|^2\right] + \dfrac{m-1}{m}\left[ f(w^s_{k+1}) - f(w^s_{k})\right],
	\end{aligned}
	\end{equation}
	second equality follows from, $\mathbb{E}\left[\tilde{\nabla}_{s,k}\right] = \nabla f(w^s_k) + \dfrac{m-1}{m}\nabla f(\tilde{w}^{s-1})$, first inequality follows from Lemma~\ref{lemma_3pt} and second inequality follows from the convexity, i.e., $ f(x) \ge f(y) + <\nabla f(y), x - y>$.
	
	\begin{equation}
	\label{eq_proof_sc_ns3}
	\begin{aligned}
	& \text{And},\; \mathbb{E} \left[  <\nabla f(w^s_k) - \tilde{\nabla}_{s,k}, w^s_{k+1} - w^s_k> - \dfrac{L(\beta-1)}{2} \| w^s_{k+1} - w^s_k \|^2 \right]\\
	& \le \mathbb{E} \left[ \dfrac{1}{2L(\beta-1)} \|\nabla f(w^s_k) - \tilde{\nabla}_{s,k}\|^2 + \dfrac{L(\beta-1)}{2} \| w^s_{k+1} - w^s_k \|^2 - \dfrac{L(\beta-1)}{2} \| w^s_{k+1} - w^s_k \|^2 \right]\\
	& = \dfrac{1}{2L(\beta-1)} \mathbb{E} \left[\| \tilde{\nabla}_{s,k} - \nabla f(w^s_k)\|^2 \right]\\
	& \le \dfrac{1}{2L(\beta-1)}\left[8L\alpha(b) \left[ F(w^s_k) - F^{*} \right] + \dfrac{8L\left(\alpha(b)m^2+(m-1)^2\right)}{m^2} \left[ F(\tilde{w}^{s-1}) - F^{*} \right] + R^{'}\right]\\
	& = \dfrac{4\alpha(b) }{(\beta-1)}\left[ F(w^s_k) - F^{*} \right] + \dfrac{4\left(\alpha(b)m^2+(m-1)^2\right)}{m^2(\beta-1)} \left[ F(\tilde{w}^{s-1}) - F^{*} \right] + R^{''},
	\end{aligned}
	\end{equation}
	first inequality follows from Young's inequality and second inequality follows from Lemma~\ref{lemma_variance_bound_nonsmooth} and $R^{''} = R^{'}/(2L(\beta-1))$. Now, substituting the values into \ref{eq_proof_sc_ns1} from inequalities \ref{eq_proof_sc_ns2} and \ref{eq_proof_sc_ns3}, and taking expectation w.r.t. mini-batches, we have
	\begin{equation}
	\label{eq_proof_sc_ns4}
	\begin{aligned}
	&\mathbb{E} \left[ F(w^s_{k+1})\right] \\& \le F(w^{*}) + \dfrac{L\beta}{2} \left[ \| w^{*} - w^s_k \|^2  -  \| w^{*} - w^s_{k+1}\|^2\right] + \dfrac{m-1}{m}\left[ f(w^s_{k+1}) - f(w^s_{k})\right]\\ &+ \dfrac{4\alpha(b) }{(\beta-1)}\left[ F(w^s_k) - F^{*} \right] + \dfrac{4\left(\alpha(b)m^2+(m-1)^2\right)}{m^2(\beta-1)} \left[ F(\tilde{w}^{s-1}) - F^{*} \right] + R^{''}\\
	&\mathbb{E} \left[ F(w^s_{k+1})-F(w^{*})\right] \\& \le \dfrac{L\beta}{2} \left[ \| w^{*} - w^s_k \|^2  -  \| w^{*} - w^s_{k+1}\|^2\right] + \dfrac{m-1}{m}\left[ f(w^s_{k+1}) - f(w^s_{k})\right]\\ &+ \dfrac{4\alpha(b) }{(\beta-1)}\left[ F(w^s_k) - F^{*} \right] + \dfrac{4\left(\alpha(b)m^2+(m-1)^2\right)}{m^2(\beta-1)} \left[ F(\tilde{w}^{s-1}) - F^{*} \right] + R^{''}
	\end{aligned}
	\end{equation}
	Taking sum over $k=0,1,...,(m-1)$ and dividing by $m$, we have
	\begin{equation}
	\label{eq_proof_sc_ns5}
	\begin{aligned}
	&\dfrac{1}{m}\sum_{k=0}^{m-1}\mathbb{E} \left[ F(w^s_{k+1})-F(w^{*})\right] \\& \le \dfrac{1}{m}\sum_{k=0}^{m-1}\left\lbrace \dfrac{L\beta}{2} \left[ \| w^{*} - w^s_k \|^2  -  \| w^{*} - w^s_{k+1}\|^2\right] + \dfrac{m-1}{m}\left[ f(w^s_{k+1}) - f(w^s_{k})\right] \right\rbrace\\ &+ \dfrac{1}{m}\sum_{k=0}^{m-1}\left\lbrace\dfrac{4\alpha(b) }{(\beta-1)}\left[ F(w^s_k) - F^{*} \right] + \dfrac{4\left(\alpha(b)m^2+(m-1)^2\right)}{m^2(\beta-1)} \left[ F(\tilde{w}^{s-1}) - F^{*} \right] + R^{''} \right\rbrace\\
	&\dfrac{1}{m}\sum_{k=1}^{m}\mathbb{E} \left[ F(w^s_{k})-F(w^{*})\right] \\& \le \dfrac{L\beta}{2m} \left[ \| w^{*} - w^s_0 \|^2  -  \| w^{*} - w^s_{m}\|^2\right] + \dfrac{m-1}{m^2}\left[ f(w^s_{m}) - f(w^s_{0})\right]\\ &+ \dfrac{4\alpha(b)}{(\beta-1)}\dfrac{1}{m}\left\lbrace \sum_{k=1}^{m}\left[ F(w^s_k) - F(w^{*}) \right] + F(w^s_0) - F(w^{*})  - \left(F(w^s_m) - F(w^{*}) \right)\right\rbrace\\& + \dfrac{4\left(\alpha(b)m^2+(m-1)^2\right)}{m^2(\beta-1)} \left[ F(\tilde{w}^{s-1}) - F(w^{*}) \right]  + R^{''}
	\end{aligned}
	\end{equation}
	Subtracting, $\dfrac{4\alpha(b)}{(\beta-1)}\dfrac{1}{m}\sum_{k=1}^{m}\left[ F(w^s_k) - F(w^{*}) \right]$ from both sides, we have
	\begin{equation}
	\label{eq_proof_sc_ns6}
	\begin{aligned}
	& \left(1-\dfrac{4\alpha(b)}{(\beta-1)}\right)\dfrac{1}{m}\sum_{k=1}^{m}\mathbb{E} \left[ F(w^s_k) - F(w^{*}) \right] \\ & \le \dfrac{L\beta}{2m} \left[ \| w^{*} - w^s_0 \|^2  -  \| w^{*} - w^s_{m}\|^2\right] + \dfrac{m-1}{m^2}\left[ f(w^s_{m}) - f(w^s_{0})\right]\\ &+ \dfrac{4\alpha(b)}{(\beta-1)}\dfrac{1}{m}\left\lbrace F(w^s_0) - F(w^{*})  - \left(F(w^s_m) - F(w^{*}) \right)\right\rbrace\\& + \dfrac{4\left(\alpha(b)m^2+(m-1)^2\right)}{m^2(\beta-1)} \left[ F(\tilde{w}^{s-1}) - F(w^{*}) \right]  + R^{''}\\
	& \le \dfrac{L\beta}{2m} \left[ \| w^{*} - w^s_0 \|^2 \right] + \dfrac{m-1}{m^2}\left[ F(w^s_{m}) - F(w^s_{0})\right] + \dfrac{4\alpha(b)}{(\beta-1)}\dfrac{1}{m}\left\lbrace F(w^s_0) - F(w^{*})  - \left(F(w^s_m) - F(w^{*}) \right)\right\rbrace\\& + \dfrac{4\left(\alpha(b)m^2+(m-1)^2\right)}{m^2(\beta-1)} \left[ F(\tilde{w}^{s-1}) - F(w^{*}) \right]  + R^{'''}\\
	& \le \dfrac{L\beta}{2m} \dfrac{2}{\mu} \left[ F(w^s_0) - F(w^{*}) \right] + \dfrac{m-1}{m^2}\left[ F(w^s_{m}) - F(w^s_{0})\right] + \dfrac{4\alpha(b)}{(\beta-1)}\dfrac{1}{m}\left\lbrace F(w^s_0) - F(w^{*})  - \left(F(w^s_m) - F(w^{*}) \right)\right\rbrace\\& + \dfrac{4\left(\alpha(b)m^2+(m-1)^2\right)}{m^2(\beta-1)} \left[ F(\tilde{w}^{s-1}) - F(w^{*}) \right]  + R^{'''}\\
	& = \left( \dfrac{L\beta}{m\mu} + \dfrac{4\alpha(b)}{m(\beta-1)} - \dfrac{m-1}{m^2}\right) \left[ F(w^s_0) - F(w^{*}) \right] + \left(\dfrac{m-1}{m^2} - \dfrac{4\alpha(b)}{m(\beta-1)} \right) \left[F(w^s_{m}) - F(w^{*})\right]\\& + \dfrac{4\left(\alpha(b)m^2+(m-1)^2\right)}{m^2(\beta-1)} \left[ F(\tilde{w}^{s-1}) - F(w^{*}) \right]  + R^{'''}\\
	& \le \left( \dfrac{L\beta}{m\mu} + \dfrac{4\alpha(b)}{m(\beta-1)} - \dfrac{m-1}{m^2}\right) c \left[ F(\tilde{w}^{s-1}) - F(w^{*}) \right] + \left(\dfrac{m-1}{m^2} - \dfrac{4\alpha(b)}{m(\beta-1)} \right) c \left[F(\tilde{w}^{s}) - F(w^{*})\right]\\& + \dfrac{4\left(\alpha(b)m^2+(m-1)^2\right)}{m^2(\beta-1)} \left[ F(\tilde{w}^{s-1}) - F(w^{*}) \right]  + R^{'''}\\
	& \le \left(\dfrac{Lc\beta}{m\mu} + \dfrac{4c\alpha(b)}{m(\beta-1)} - \dfrac{c(m-1)}{m^2}  + \dfrac{4\left(\alpha(b)m^2+(m-1)^2\right)}{m^2(\beta-1)}\right) \left[ F(\tilde{w}^{s-1}) - F(w^{*}) \right] \\&+ \left(\dfrac{m-1}{m^2} - \dfrac{4\alpha(b)}{m(\beta-1)}\right)c \left[F(\tilde{w}^{s}) - F(w^{*})\right]  + R^{'''}
	\end{aligned}
	\end{equation}
	second inequality follows from dropping, $\| w^{*} - w^s_{m}\|^2 \ge 0$ and converting, $f(w^s_{m}) - f(w^s_{0})$ to $F(w^s_{m}) - F(w^s_{0})$ by introducing some constant, third inequality follows from the strong convexity, i.e., $\|  w^s_0 - w^{*}\|^2 \le 2/ \mu\left( f(w^s_0)- f^{*} \right) $, fourth inequality follows from Assumption~\ref{assump_3shang} and $R^{'''} = R^{''} + (m-1)g(w^s_0)/m^2$.\\
	Since $\tilde{w}^{s} = 1/m \sum_{k=1}^{m}w^s_k$ so by convexity using, $f(\tilde{w}^{s}) \le 1/m \sum_{k=1}^{m} f(w^s_k)$, and subtracting $\left(\dfrac{m-1}{m^2} - \dfrac{4\alpha(b)}{m(\beta-1)} \right) c \left[F(\tilde{w}^{s}) - F(w^{*})\right]$ from both sides, we have
	\begin{equation}
	\label{eq_proof_sc_ns6_1}
	\begin{aligned}
	&\left(1-\dfrac{4\alpha(b)}{(\beta-1)} - \dfrac{c(m-1)}{m^2} + \dfrac{4c\alpha(b)}{m(\beta-1)}\right)\mathbb{E} \left[ F(\tilde{w}^{s}) - F(w^{*}) \right] \\ & \le \left(\dfrac{Lc\beta}{m\mu} + \dfrac{4c\alpha(b)}{m(\beta-1)} - \dfrac{c(m-1)}{m^2}  + \dfrac{4\left(\alpha(b)m^2+(m-1)^2\right)}{m^2(\beta-1)}\right) \left[ F(\tilde{w}^{s-1}) - F(w^{*}) \right]  + R^{'''}
	\end{aligned}
	\end{equation}
	Dividing both sides by $ \left(1-\dfrac{4\alpha(b)}{(\beta-1)} - \dfrac{c(m-1)}{m^2} + \dfrac{4c\alpha(b)}{m(\beta-1)}\right)$, we have
	\begin{equation}
	\label{eq_proof_sc_ns7}
	\begin{aligned}
	&\mathbb{E} \left[ F(\tilde{w}^{s}) - F(w^{*}) \right] \le C \left[ F(\tilde{w}^{s-1}) - F(w^{*}) \right] + R^{''''}
	\end{aligned}
	\end{equation}
	where\\ $C = \left(\dfrac{Lc\beta}{m\mu} + \dfrac{4c\alpha(b)}{m(\beta-1)} - \dfrac{c(m-1)}{m^2}  + \dfrac{4\left(\alpha(b)m^2+(m-1)^2\right)}{m^2(\beta-1)}\right) \left(1-\dfrac{4\alpha(b)}{(\beta-1)} - \dfrac{c(m-1)}{m^2} + \dfrac{4c\alpha(b)}{m(\beta-1)}\right)^{-1} $ and $R^{''''} = R^{'''}\left(1-\dfrac{4\alpha(b)}{(\beta-1)} - \dfrac{c(m-1)}{m^2} + \dfrac{4c\alpha(b)}{m(\beta-1)}\right)^{-1} $ . Now, applying this inequality recursively, we have
	\begin{equation}
	\label{eq_proof_sc_ns8}
	\begin{aligned}
	\mathbb{E} \left[ F(\tilde{w}^{s}) - F(w^{*}) \right] \le C^s \left[ F(\tilde{w}^{0}) - F(w^{*}) \right] + R^{'''''},
	\end{aligned}
	\end{equation}
	inequality follows for $R^{'''''}= R^{''''}/(1-C)$, since $\sum_{i=0}^{k} r^i \le \sum_{i=0}^{\infty} r^i = \dfrac{1}{1-r}, \quad \|r\|<1$. For certain choice of $\beta$, one can easily prove that $C<1$. This proves linear convergence with some initial error.	
\end{proof}

\end{document}